\documentclass[AMA,Times1COL]{WileyNJDv5} %STIX1COL,STIX2COL,STIXSMALL

\articletype{RESEARCH ARTICLE}%

\received{Date Month Year}
\revised{Date Month Year}
\accepted{Date Month Year}
\journal{Journal}
\volume{00}
\copyyear{2023}
\startpage{1}

\usepackage[T1]{fontenc}
\usepackage{geometry}
\usepackage{amsmath}

\DeclareMathOperator*{\argmin}{arg\,min}
\usepackage{amsthm}
\usepackage{stmaryrd}
\usepackage{graphicx}
\usepackage{booktabs}
\usepackage{multirow}
\usepackage{makecell}
\usepackage{xurl}
\usepackage{esint}
\usepackage{algorithm}
\usepackage{algpseudocode}
\makeatletter

\providecommand{\proofname}{Proof}
\fi

\journal{submitted to IJNME}

\usepackage{hyperref}
\hypersetup{colorlinks = true, allcolors = blue}

\usepackage[nameinlink]{cleveref}

\crefname{figure}{Fig.}{Figs.}
\crefformat{equation}{Eq.~#2(#1)#3}
\crefformat{section}{Section~#2#1#3}
\AtBeginDocument{%
	\let\cite\cite
}

\usepackage[labelfont=bf]{caption}
\@ifundefined{showcaptionsetup}{}{%
	\PassOptionsToPackage{caption=false}{subfig}}
\usepackage{subfig}
\makeatother

\usepackage{babel}

\begin{document}

\title{DCEM: A deep complementary energy method  for linear elasticity}

\author[rvt,rvt3]{Yizheng Wang}

\author[rvt]{Jia Sun}

\author[rvt3]{Timon Rabczuk}

\author[rvt]{Yinghua Liu}

\address[rvt]{\orgdiv{Department of Engineering Mechanics}, \orgname{Tsinghua University}, \orgaddress{\state{Beijing}, \country{China}}}

\address[rvt3]{\orgdiv{Institute of Structural Mechanics}, \orgname{Bauhaus-Universit\"{a}t Weimar, Marienstr. 15}, \orgaddress{\state{Weimar}, \country{Germany}}}

\corres{Yinghua Liu, Department of Engineering Mechanics, Tsinghua University, Beijing 100084,
	China. \email{\url{yhliu@mail.tsinghua.edu.cn}}}

%\presentaddress{This is sample for present address text this is sample for present address text.}

\fundingInfo{Key Project of the National Natural Science Foundation of China (12332005)}
%\JELinfo{ejlje}h

\abstract[Abstract]{In recent years, the rapid advancement of deep learning has significantly impacted various fields, particularly in solving partial differential equations (PDEs) in the realm of solid mechanics, benefiting greatly from the remarkable approximation capabilities of neural networks.
	In solving PDEs, Physics-Informed Neural Networks (PINNs) and the Deep Energy Method (DEM) have garnered substantial attention.
	The principle of minimum potential energy and complementary energy are two important variational principles in solid mechanics. However, the well-known Deep Energy Method (DEM) is based on the principle of minimum potential energy, but it lacks the important form of minimum complementary energy.  To bridge this gap, we propose the deep complementary energy method (DCEM) based on the principle of minimum complementary energy. The output function of DCEM is the stress function, which inherently satisfies the equilibrium equation. 
	We present numerical results of classical linear elasticity using the Prandtl and Airy stress functions, and compare DCEM with existing PINNs and DEM algorithms when modeling representative mechanical problems. The results demonstrate that DCEM outperforms DEM in terms of stress accuracy and efficiency and has an advantage in dealing with complex displacement boundary conditions, which is supported by theoretical analyses and numerical simulations. 
	We extend DCEM to DCEM-Plus (DCEM-P), adding terms that satisfy partial differential equations. Furthermore, we propose a deep complementary energy operator method (DCEM-O) by combining operator learning with physical equations. Initially, we train DCEM-O using high-fidelity numerical results and then incorporate complementary energy.
	DCEM-P and DCEM-O further enhance the accuracy and efficiency of DCEM.
	}

%\begin{graphicalabstract}
%	\includegraphics{DCEMgraph_abstract}
%\end{graphicalabstract}


%\jnlcitation{\cname{%
%\author{Taylor M.},
%\author{Lauritzen P},
%\author{Erath C}, and
%\author{Mittal R}}.
%\ctitle{On simplifying ‘incremental remap’-based transport schemes.} \cjournal{\it J Comput Phys.} \cvol{2021;00(00):1--18}.}

\maketitle

%\renewcommand\thefootnote{}
%\footnotetext{\textbf{Abbreviations:} ANA, anti-nuclear antibodies; APC, antigen-presenting cells; IRF, interferon regulatory %factor.}

%\renewcommand\thefootnote{\fnsymbol{footnote}}
%\setcounter{footnote}{1}

\section{Introduction} 

The laws of nature can be described and approximated through various means, including partial differential equations (PDEs), which reflect the relationship of physical quantities with space and time. PDEs stand as a foundational tool for describing the laws of nature, playing a crucial role in fields like numerical weather prediction and computational solid mechanics   \cite{loss_is_minimum_potential_energy}.
Take the discoveries of Kepler and Newton as a specific example for illustrating the importance of PDEs, it can be observed that Newton's discovery of PDEs explained the underlying mechanisms responsible for elliptic orbits based on Kepler's data-driven models. As a result, the PDEs model is an abstract and symbolic representation of big data. PDEs have exhibited remarkable durability in engineering design, enabling the successful landing of spacecraft on the moon \cite{brunton2016discovering}.

Obtaining an analytical solution to PDEs is important in the process of engineering applications, which is often unlikely in real-world applications, necessitating the use of numerical methods to approximate the analytical solution.
There are several traditional numerical methods for approximating solutions of PDEs, including the well-established finite element method (FEM) \cite{zienkiewicz2005finite,hughes2012finite,bathe2006finite,reddy2019introduction}, the finite difference method \cite{darwish2016finitevolumemethod}, the finite volume method \cite{leveque2007finitedifferentialmethod}, and mesh-free method \cite{zhang2016material,liu2003mesh,rabczuk2004cracking,nguyen2008meshless,rabczuk2007three,rabczuk2019extended}.
These methods have been developed and used for a long time, and are considered reliable numerical methods, especially FEM in solid mechanics simulation.
Different trial and test functions produce various
finite element numerical methods. The essence of FEM is to find the
optimal solution in the approximate function space. However, constructing different elements requires a significant amount of cognitive cost \cite{wang2022cenn}. FEM has proven to be highly effective for many nonlinear problems, but solving the nonlinear problem by FEM involves assembling a complex tangent matrix and residual vector in the implicit FEM scheme. Furthermore, the explicit FEM scheme often faces conditional stability constraints, which demand small time increments \cite{koric2009explicit}.

The past decade has witnessed a significant impact of artificial intelligence across a range of fields, especially deep learning.
Various fields \cite{alphago,star_game, ALEXNET,speech_recognition,machine_translation,alphafold} are benefiting from deep learning,
and computational solid mechanics is no exception \cite{abueidda2022deep}.
One crucial aspect of integrating deep learning and mechanics is leveraging the powerful approximation capabilities of neural networks and the abundance of reliable big data to model complex relationships between inputs and outputs, such as constitutive models \cite{kirchdoerfer2016data,kirchdoerfer2018data, li2022equilibrium, hyper_constitutive_data_driven}, the prediction of the equivalent modulus of non-uniform materials \cite{li2019predicting}, inverse construction and topology optimization of metamaterials \cite{li2020designing, kollmann2020deep, liu2023multi}. A model that is already trained can often achieve high efficiency and accuracy, without accounting for the training time, but its drawback is usually the low interpretability and the large amount of data required.
Fitting the data with a neural network often results in poor model scalability, i.e., the neural network often needs to be retrained for different problems, and even the network structure may change drastically \cite{ill_gradient}. Moreover, the physical mechanism behind it is often not well understood \cite{li2022ref}. This aspect of combination faces challenges in solving high-dimensional problems because it relies on existing methods to obtain data, which leads to the curse of dimensionality \cite{bc-pinn}.

The universal approximation theorem states that multilayer feedforward neural networks with any nonconstant and bounded activation function and at least one hidden layer can act as universal approximators with arbitrary accuracy \cite{super_approximation, hornik1991approximation}. 
This theorem opens up a new way to solve PDEs, which is another aspect of combining deep learning and mechanics. Raissi et al. \cite{PINN_original_paper} proposed the most popular method in this direction, i.e., physics-informed neural networks (PINNs), although the idea of using the neural network to solve PDEs can be traced back to the last century \cite{admissible_earliest_paper,use_neural_network_to_solve_PDE}.
PINNs are neural networks that incorporate partial differential equations (PDEs) into their loss function. The original PINNs considered the strong form of PDEs in solid mechanics called the deep collocation method (DCM) \cite{loss_is_minimum_potential_energy, he2023deep}.  PINNs have been applied to various fields, such as solid mechanics \cite{PINN_solid_mechanics, li2022equilibrium, goswami2022physics}, fluid mechanics \cite{cai2021physics, raissi2020hidden}, and biomechanics \cite{yin2021non, yin2022simulating}. Moreover, Samaniego and Nguyen-Thanh et al. \cite{loss_is_minimum_potential_energy}  presented a deep energy method (DEM)\footnote{In this work, DCM means the strong form of PINNs, DEM means the energy form of PINNs} where the loss function can be reformulated as an energy functional in solid mechanics with applications to linear elasticity \cite{loss_is_minimum_potential_energy}, fracture mechanics \cite{goswami2020adaptive, goswami2020transfer, goswami2022physics}, hyperelasticity \cite{PINN_hyperelasticity}, viscoelasticity \cite{abueidda2022deep}, strain gradient elasticity \cite{paradeepenergy}, elastoplasticity \cite{he2023deep}. The core idea of DEM is that the principle of minimum potential energy can be used for optimization. 

PINNs have the advantage of generality, i.e.,  almost any PDEs can be solved by PINNs. However, they face the challenge of determining the optimal hyperparameters, especially in solid mechanics where the equations involve high-order tensors and derivatives. Although some recent studies have proposed methods for selecting hyperparameters \cite{NTK_PINN, NTK_to_get_hyperparameter_of_PINN,ill_gradient}, DEM has the advantage of having fewer hyperparameters and achieving higher efficiency and accuracy due to lower order derivative than PINNs, but it does not apply to all PDEs, as it is hard for some of them to derive an energy formulation. Moreover, DEM requires the prior construction of the field of interest due to the requirement of the variational principle \cite{the_comparision_of_strong_and_energy_form}. 
Fuhg et al. \cite{fuhg2022mixed} have shown that DEM cannot solve the concentration feature like stress very well for finite strain hyperelasticity, so they propose a mixed deep energy method (mDEM). The core idea of mDEM is to add some extra loss terms, including the stress obtained by neural network output and the constitutive law describing the relationship between displacement and stress. 
Abueidda et al. \cite{abueidda2023enhanced} improve mDEM by introducing the Fourier transform to the input of neural network for promoting the perception of high-frequency function and incorporating both the strong form and energy form.
To apply DEM, one needs to perform a successful integration over the domain defined by the integration points \cite{abueidda2022deep}.
The deep learning approach relies mainly on matrix-vector multiplications, which are highly optimized and greatly benefit from GPUs. However, it is difficult to fully exploit GPUs in an implicit FEM analysis.
In addition, developing nonlinear PINNs to solve PDEs is much simpler than FEM \cite{RN676}, attributing to the machine learning packages such as PyTorch \cite{paszke2019pytorch} and TensorFlow \cite{abadi2016tensorflow}.

In the linear theory of elasticity, the most critical variational principle is the dual extreme principle, i.e., the principle of minimum potential and complementary energy \cite{nonlinear_plate_theory_comple}.  However, DEM currently lacks a complementary energy principle. It is well-known that the stress finite element (equilibrium element) plays a vital role in computational mechanics \cite{fraeijs1963upper,de1968equilibrium,fraeijs1965displacement,fraeijs1966bending}. Thus, proposing a deep energy method based on the principle of complementary energy is of utmost importance. To date, purely data-driven methods require large amounts of data, and it is not easy to surpass traditional finite element methods in terms of efficiency and accuracy if they solely rely on physical equations. 
Although the theory of generalized approximate functions shows great theoretical prospects, significant work remains to achieve this goal \cite{lu2022comprehensive}.
 Recently, the practical application potential \cite{goswami2022physics,wang2021learning, wen2022u} of DeepONet \cite{DeepOnet} and FNO (Fourier neural operator) \cite{li2020fourier} has been receiving attention due to their basis in the universal approximation operator theory \cite{chen1995universal}.  Goswami et al. \cite{goswami2022physics} have shown the potential for a combination of DeepONet and DEM to improve computational efficiency and accuracy when predicting crack path by phase field modeling of fracture. Therefore, Semi-supervised learning, combined with operator and physical laws, has great potential in the future of data-driven approaches.

\begin{figure}
	\begin{centering}
		\includegraphics[scale=1.3]{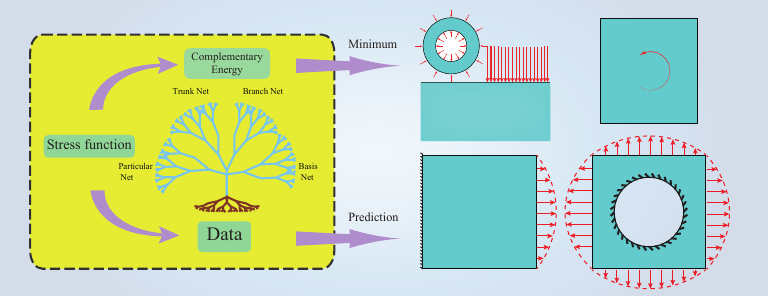}
		\par\end{centering}
	\caption{Schematic of DCEM: The process involves initially employing data, whether from experiments or highly accurate simulation results, to pre-train an operator learning model and obtain a good initial solution. Subsequently, the physical equations (in DCEM, we use the principle of the complementary energy, other PDEs are also feasible in the framework) are applied to refine the initial solution, enabling a faster acquisition of a reasonable reference solution. \label{fig:DCEM_schematic}}
\end{figure}

In this work, we propose a novel deep energy form based on the principle of minimum complementary energy (DCEM) as shown in \Cref{fig:DCEM_schematic}, instead of the traditional minimum potential energy principle. DCEM is primarily used to solve linear elastic problems in traditional solid mechanics and has limited effectiveness in solving nonlinear problems. There are two reasons for the difficulty of applying the stress function method to nonlinear problems including geometric and material nonlinearity. In nonlinear mechanics, the material nonlinearity makes it challenging to directly apply the stress function method. 
In the case of small strain linear elasticity, the nonlinear term in nonlinear geometric equations becomes negligible, leading us to disregard this nonlinear term and express the governing equations of the stress function by geometric equations of linear elasticity.
As a result, if the problem is nonlinear, the governing equations of stress function not only become complex but also different materials referred to different forms of constitutive law should have different corresponding governing equations of stress function in nonlinear problems. Thus, this makes the stress function approach difficult to apply to nonlinear problems. Hence, we only solve the linear problems by DCEM in this work.
Our output function is the stress function, which naturally satisfies the equilibrium equation. This is the first attempt to leverage the power of the physics-informed neural networks (PINNs) energy form based on complementary energy. We further extend the capabilities of DCEM and propose a DCEM-Plus (DCEM-P) algorithm, where we add terms that naturally satisfy the biharmonic equation in the Airy stress function for better accuracy and efficiency compared to DCEM. We develop DCEM based on the DeepONet operator (DCEM-O), including branch net, trunk net, basis net, and particular net.  The motivation behind DCEM-O is to harness and revive existing calculation results (data) and not waste previous calculation results to improve the computational efficiency of DCEM.  The data is from our current or future high-fidelity numerical results and experimental data.
The advantages of the DCEM are multi-fold  : 
\begin{itemize}
	\item \textbf{Accuracy in Stress}: DCEM introduces a new energy form based on the minimum complementary energy principle, distinct from traditional DEM based on the minimum potential energy. This makes it a valuable supplement to existing DEM methods, offering improved accuracy, especially in stress predictions.
	
	\item \textbf{Enhanced Accuracy}: DCEM-Plus (DCEM-P) extends DCEM by incorporating terms that satisfy the biharmonic equation in the Airy stress function. This idea of DCEM-P can be applied to other methods of PINNs as well.
	
	\item \textbf{Computational Efficiency}: DCEM can be combined with physical equations and existing data through operator learning (DCEM-O). Leveraging additional data improves computational efficiency, making it a powerful and efficient approach, particularly with big data. The reason is that we can train operator learning with big data to obtain an initial solution, which is then fine-tuned using physical equations. Since the initial solution provided by big data is close to the exact solution, the number of iterations required by the physical equations is greatly reduced.

	\item \textbf{Convenient for Complex Displacement Boundaries}: DCEM's theoretical superiority over DEM in constructing admissible functions is particularly advantageous when dealing with complex displacement boundary conditions. It eliminates the need for constructing an admissible function field on the displacement boundary, simplifying the implementation process.
	
\end{itemize}

The outline of the paper is as follows. In \Cref{sec: prerequisite}, we introduce the prerequisite knowledge, including the feed-forward network, DeepONet algorithm, deep energy method, and the important stress functions in solid mechanics. In \Cref{sec: method}, we describe the methodology of the proposed method DCEM, DCEM-P, and DCEM-O. In \Cref{sec: result}, we present the numerical results of the most common stress functions, Prandtl and Airy stress functions, with different boundary conditions under DCM, DEM, and DCEM. The numerical experiments are the circular tube, wedge problem, plate with hole, and torsion problem.
The results demonstrate that DCEM exhibits better accuracy and efficiency in terms of stress compared to DEM. DCEM has advantages in solving problems primarily influenced by the displacement boundary. Finally, \Cref{sec:Discussion} and \Cref{sec:Conclusion} present some discussions of DCEM, concluding remarks, and some limitations of DCEM such as we only deal with the linear elasticity problem without body force and possible future work. In the Appendices, we show some important proofs of DCEM. 

\section{Prerequisite knowledge} \label{sec: prerequisite}

\subsection{Introduction to feed-forward neural networks}
There are four primary types of neural networks: fully connected (also known as feed-forward neural networks), convolutional neural networks \cite{lecunCNN}, recurrent neural networks (with LSTM being the most commonly used), and the more recent Transformer \cite{vaswani2017attention}. The combination of machine learning and computational mechanics relies heavily on the powerful fitting capabilities of neural networks \cite{lu2022comprehensive, hornik1989multilayer}. As a result, feed-forward neural networks remain the mainstream choice for scientific applications involving machine learning. Therefore, this article will focus mainly on the feed-forward neural networks, which is essentially a multiple linear regression model enhanced with nonlinear capabilities through the activation function. The mathematical formula for the feed-forward neural networks can be expressed as follows:

\begin{equation}
	\begin{split}\boldsymbol{z}^{(1)} & =\boldsymbol{w}^{(1)}\cdot\boldsymbol{x}+\boldsymbol{b}^{(1)}\\
		\boldsymbol{a}^{(1)} & =\boldsymbol{\sigma}(\boldsymbol{z}^{(1)})\\
		\vdots\\
		\boldsymbol{z}^{(L)} & =\boldsymbol{w}^{(L)}\cdot\boldsymbol{a}^{(L-1)}+\boldsymbol{b}^{(L)}\\
		\boldsymbol{a}^{(L)} & =\boldsymbol{\sigma}(\boldsymbol{z}^{(L)})\\
		\boldsymbol{y} & =\boldsymbol{w}^{(L+1)}\cdot\boldsymbol{a}^{(L)}+\boldsymbol{b}^{(L+1)},
	\end{split}
\end{equation}
where $\boldsymbol{x}$ is input, $\boldsymbol{y}$ is output, $\boldsymbol{z}$ is linear output. Activation function $\boldsymbol{\sigma}$ is applied to $\boldsymbol{z}$, and trainable parameters are $\boldsymbol{w}$ and $\boldsymbol{b}$. Layers' neurons $\boldsymbol{a}^{(l-1)}$ are linearly transformed to $\boldsymbol{z}^{(l)}$. Activation function $\boldsymbol{a}^{(l)}$ is typically nonlinear (e.g., {\rm tanh}). In this work,  {\rm tanh} function is used as an activation function.

\subsection{Introduction to DeepONet} \label{sec: introduction to DeepONet}
DeepONet is a specific neural network architecture based on neural networks to learn operators \cite{DeepOnet}. The mathematical structure is similar to polynomial and periodic function fitting 

\begin{equation}
	f(\boldsymbol{y})=\sum_{i=1}^{n}\alpha_{i}\phi(\boldsymbol{y}_{i}).\label{eq:basisfunction approximation}
\end{equation}

The Trunk net is fitted with a neural network using the basis function $\phi(\boldsymbol{y})$. This approach shares a fundamental concept with PINNs, where the neural network fits from the coordinate space to the objective function. However, in DeepONet, the basis function is fitted instead of the objective function, and the constant weight $\alpha$ in front of the basis function is fitted by the branch net. Notably, the same input function of the branch net yields a fixed weight $\alpha$, which aligns with the function approximation concept in numerical analysis. In contrast to traditional function approximation algorithms that pre-select basis functions, DeepONet leverages neural networks to adaptively select suitable basis functions based on the data.

Since Trunk net resembles PINNs, we can apply the automatic differential algorithm \cite{automatic_differential} in PINNs to construct a partial differential operator and obtain a new loss function (strong form or energy principle) based on PDEs. In this work, we propose a semi-supervised combination of data operators and partial differential equations, which will be discussed in detail in DCEM-O \Cref{sec: DCEM-o}. Next, we will introduce the network structure of DeepONet.

The network structure of DeepONet is similar to the theoretical results of the article on universal operator approximation in 1995  \cite{DeepOnet, chen1995universal}
\begin{equation}
	|G(\boldsymbol{u})(\boldsymbol{y})-\sum_{k=1}^{p}\sum_{i=1}^{n}c_{i}^{k}\sigma(\sum_{j=1}^{m}w_{ij}^{k}\boldsymbol{u}(\boldsymbol{x}_{j})+b_{i}^{k})\sigma(W^{k}\cdot\boldsymbol{y}+B^{k})|<\epsilon,\label{eq: DeepONet theory}
\end{equation}
where $G$ is the operator that maps the function $\boldsymbol{u}$ to $G(\boldsymbol{u})$, and $\epsilon$ is the error between the exact operator $G$ and the approximate operator. We can use the neural network to replace the item in \Cref{eq: DeepONet theory}

\begin{equation}
	\begin{cases}
		NN_{B}^{k}(\boldsymbol{u}(\boldsymbol{x});\boldsymbol{\theta}_{NN}^{B}) & =\sum_{i=1}^{n}c_{i}^{k}\sigma(\sum_{j=1}^{m}w_{ij}^{k}\boldsymbol{u}(\boldsymbol{x}_{j})+b_{i}^{k})\\
		NN_{T}^{k}(\boldsymbol{y};\boldsymbol{\theta}_{NN}^{T}) & =\sigma(W^{k}\cdot\boldsymbol{y}+B^{k})\\
		NN_{O}(\boldsymbol{y};\boldsymbol{u}) & =\sum_{k=1}^{p}NN_{B}^{k}(\boldsymbol{u}(\boldsymbol{x});\boldsymbol{\theta}_{NN}^{B})\cdot NN_{T}^{k}(\boldsymbol{y};\boldsymbol{\theta}_{NN}^{T}),
	\end{cases}
\end{equation}
where $NN_{B}^{k}$, $NN_{T}^{k}$, and $NN_{O}$ are the output of the branch net, trunk net, and DeepONet respectively. $\boldsymbol{u}(\boldsymbol{x})$ can be approximated from the sensor $\boldsymbol{x}$, i.e., discrete points approximate continuous function. For example, if the input $\boldsymbol{u}(\boldsymbol{x})$ of $NN_{B}$  is the gravitational potential energy of a three-dimensional cube material ($[0,1]^{3}$), the gravitational potential energy field in $[0,1]^{3}$ can take $101^{3}$ at equal intervals of 0.01.

 The data structure of $\boldsymbol{u}(\boldsymbol{x})$ is a variety of different structures, such as the data structure close to the image. Therefore, we can use corresponding network structures as the network architecture of the branch net, e.g.,  CNN with local features to deal with the data similar to an image and time-related RNN to deal with data strongly related to time. Note that the essence of the problem must be considered while choosing the network structure of the branch net. Hence, we can find that $\boldsymbol{u}(\boldsymbol{x})$ determines the output of the branch net in DeepONet, which is related to the weight $\alpha$ in \Cref{eq:basisfunction approximation}, and does not correlate with the coordinates we are interested in. Similarly,  the coordinate completely determines the output of trunk net, which has the same characteristics as the basis function $\phi(\boldsymbol{y})$ in \Cref{eq:basisfunction approximation}. Therefore, DeepONet has many similar ideas to traditional function approximation.

If the input to DeepONet's branch network is a constant field, the constant value can be provided as input instead of using a variable field function, improving computational efficiency. 
Since the branch network of DeepONet typically uses discrete values to represent the field, knowing in advance that the field is constant eliminates the need for numerous discrete values. In such cases, the constant value can replace these discrete values in the branch network.
For details and extensions about DeepONet, please refer to this paper \cite{lu2022comprehensive}.

\subsection{Introduction to deep energy method}

The deep energy method (DEM) \cite{loss_is_minimum_potential_energy} is based on the principle of minimum
potential energy to obtain the true displacement solution. The principle
of minimum potential energy means that the true displacement field
minimizes the potential energy functional $J(\boldsymbol{u})$ among
all the admissible displacement fields satisfying the displacement
boundary conditions in advance. Note that the theory here is not only
for the case of linear elasticity, but also some nonlinear hyperelastic
problems \cite{PINN_hyperelasticity}. The formulation of DEM can be written as:
\begin{equation}
	\begin{aligned}
		\boldsymbol{\theta}^{*} & =\argmin_{\boldsymbol{\theta}}J(\boldsymbol{u}(\boldsymbol{x};\boldsymbol{\theta}))\\
		s.t. \quad   \boldsymbol{u}(\boldsymbol{x}) & =\boldsymbol{\bar{u}}(\boldsymbol{x}),\boldsymbol{x}\subseteq\Gamma^{u},
	\end{aligned}
\end{equation}
where $\boldsymbol{u}(\boldsymbol{x};\boldsymbol{\theta})$ is approximated by neural network, $\boldsymbol{\bar{u}}(\boldsymbol{x})$ is the given Dirichlet
boundary $\Gamma^{u}$ condition and 
\begin{equation}
	J(\boldsymbol{u})=\int_{\Omega}\psi(\boldsymbol{u})dV-\int_{\Omega}\boldsymbol{f}\cdot\boldsymbol{u}dV-\int_{\Gamma^{t}}\bar{\boldsymbol{t}}\cdot\boldsymbol{u}dA,
\end{equation}
where $\psi$, $\Omega$, $\boldsymbol{f}$, $\boldsymbol{\bar{t}}$ and $\Gamma^{t}$ is
the strain energy density function, the domain, the body force, the
surface force and Neumann boundary respectively. To be specific, the
strain energy density in linear elasticity is
\begin{equation}
	\begin{aligned}\psi & =\frac{1}{2}\boldsymbol{\varepsilon}:\boldsymbol{C}:\boldsymbol{\varepsilon}\\
		\boldsymbol{\varepsilon} & =\frac{1}{2}(\nabla\boldsymbol{u}+\boldsymbol{u}\nabla).
	\end{aligned}
\end{equation}

\begin{figure}
	\begin{centering}
		\includegraphics[scale=0.85]{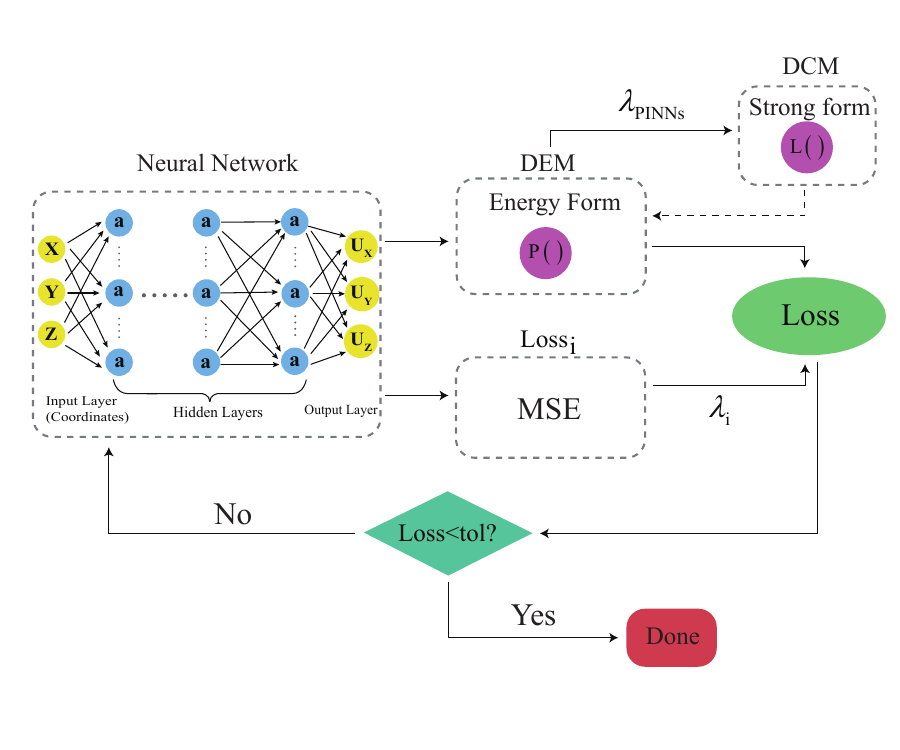}
		\par\end{centering}
	\caption{Schematic of the difference and connection between DCM and DEM, yellow circles in the Neural network are the inputs
		and outputs. Blue circles in the Neural network are the hidden neurons.
		$\text{P}()$ is the energy density of the functional. $\text{Loss}_{i}$ is
		the essential boundary loss. MSE is the mean square error to let
		the field of interest to satisfy the essential boundary. If the boundary
		condition is satisfied in advance, $\text{Loss}_{i}$ can be dismissed.  The index i in $\text{Loss}_{i}$ means $\text{Loss}_{i}$ can be the boundary condition loss and initial condition loss if temporal problem.  $\text{L}()$
		is the differential operator related to the strong form of PDEs. $\lambda_{DCM}$
		and $\lambda_{DEM}$ are the weight of the loss of DCM and DEM.
		$\lambda_{i}$ is the weight of $\text{Loss}_{i}$. Note that the number of $\lambda_{DCM}$ is determined
		by the number of PDEs. The dotted arrow means not all PDEs can
		be converted to the energy form. \label{fig:Schematic-of-PINN_DEM}}
\end{figure}

 DEM uses a neural network to
replace the approximation function like the trial function in FEM.
A neural network is used to approximate the displacement function over the physical domain of interest.
Due to the rich function space of neural networks, DEM theoretically holds great potential in solving PDEs, regardless of the optimization error.
 \Cref{fig:Schematic-of-PINN_DEM}
shows the difference and connection between DCM
and DEM. 

In DEM, we must construct the admissible displacement in advance
before using the principle of minimum potential energy.
The distance function $\boldsymbol{d}(\boldsymbol{x})$ can be applied to satisfy the requirement of the admissible displacement:
\begin{equation}
	\boldsymbol{u}(\boldsymbol{x})=\boldsymbol{\bar{u}}(\boldsymbol{x})+\boldsymbol{d}(\boldsymbol{x})\odot\boldsymbol{u}_{g}(\boldsymbol{x}),
\end{equation}
where $\odot$ is element-wise, i.e., multiplication of corresponding
elements does not change the shape, and $\boldsymbol{u}_{g}(\boldsymbol{x})$
is a general function, which is neural networks in DEM. The distance
function is a special and currently common way to
satisfy boundary condition in advance \cite{wang2022cenn,PINN_hyperelasticity,boundary_conditions_distance_functions,lu2021physics, admissible_in_PINN_energy_form,admissible_earliest_paper}. The distance function $d(\boldsymbol{x})$ in each dimension is to give the nearest
distance from the point $\boldsymbol{x}$ to Dirichlet boundary $\Gamma^{\boldsymbol{u}}$:
\begin{equation}
	d(\boldsymbol{x})=\min_{\boldsymbol{y}\in\Gamma^{\boldsymbol{u}}}\sqrt{(\boldsymbol{x}-\boldsymbol{y})\cdot(\boldsymbol{x}-\boldsymbol{y})}.
\end{equation}
 The label of the distance $y^{(i)}$ can
be obtained from the traditional method (dataset: $\{\boldsymbol{x}^{(i)},y^{(i)}\}_{i=1}^{n}$). Then we use
the NN to fit the label, i.e., 
\[
\begin{aligned}d(\boldsymbol{x}) & \approx NN(\boldsymbol{x};\boldsymbol{\theta})\\
	\boldsymbol{\theta} & =\argmin_{\boldsymbol{\theta}}\mathcal{L}=\sum_{i=1}^{n}[NN(\boldsymbol{x}^{(i)})-y^{(i)}].
\end{aligned}
\]
The neural network used to fit the distance function $\boldsymbol{d}(\boldsymbol{x})$ can also address the complex boundary problem \cite{complex_PINN_a_method_to_construct_admissible_function}.
Besides, there is an easier way to satisfy boundary conditions
than using the distance constraint, i.e., the soft constraint way \cite{deep_ritz}.
The soft constraint method is usually implemented by a penalty loss,
i.e. $\beta$$\int_{\Gamma}(\boldsymbol{u}-\bar{\boldsymbol{u}})^{2}d\Gamma$, where $\beta$ is the hyperparameter of the boundary loss, and $\bar{\boldsymbol{u}}$ is the given condition on the boundary.
However, the boundary condition cannot be satisfied exactly via soft constraint, and different values for penalty factor $\beta$ result in different numerical results.

\subsection{Introduction to stress function}\label{sec: introduction to stress function}
The stress function method is an extension of the well-known stress solution in solid mechanics, which aims to automatically satisfy the equilibrium equations \cite{the_foundation_of_solid_mechanics_feng}. When solving elasticity problems, there are two main methods: the displacement method and the stress method. While there is also a strain solution method, it is less commonly used because strain and stress are linearly related in elastic mechanics problems, making the difficulty of the strain solution equivalent to that of the stress solution.

In the stress method, the stress must satisfy the equilibrium equations, but by using a stress function, the equilibrium equations are automatically satisfied. This not only reduces the number of equations in the stress solution, but also maintains the advantages of the stress method. The advantages of the stress method yield more precise stress results, like in equilibrium elements. However, the stress method necessitates higher-order trial functions and complex integration processes to determine the displacement field, which is more complicated than taking derivatives. Therefore, the stress function solution plays a significant role in solving problems in elastic mechanics, especially stress. To better understand the stress function, we briefly
introduce displacement and stress solutions of elastic mechanics. For detailed
conceptualization of methods to solve the elastic mechanic's problems,
readers may refer to this book \cite{the_foundation_of_solid_mechanics_feng}. 

In the displacement method, the primary unknown quantity to be solved is the displacement field, denoted as $\boldsymbol{u}$.  In contrast, stress is the basic
unknown quantity in stress solution.  We substitute the strain tensor obtained by geometrical equations
\begin{equation}
	\boldsymbol{\varepsilon}=\frac{1}{2}(\nabla\boldsymbol{u}+\boldsymbol{u}\nabla) \label{eq:linear strain equation}
\end{equation}
into the linear constitutive equations, i.e., Hooke's law 
\begin{equation}
	\boldsymbol{\sigma}=\boldsymbol{C}:\boldsymbol{\varepsilon},
\end{equation}
where $\boldsymbol{u}$, $\nabla$, $\boldsymbol{\sigma}$,
$\boldsymbol{C}$ and $\boldsymbol{\varepsilon}$ is the displacement
field, gradient operator, stress tensor, elasticity tensor,
and strain tensor. Thus, we can express the stress $\boldsymbol{\sigma}$ in terms
of displacement $\boldsymbol{u}$ in the equilibrium equation
\begin{equation}
	\nabla\cdot\boldsymbol{\sigma}+\boldsymbol{f}=0, \label{eq:equilibrium equaions}
\end{equation}
where $\boldsymbol{f}$ is the body force, such as gravity. $\rho$
and $\ddot{\boldsymbol{u}}$ are the density and acceleration. $\nabla\cdot()$
means divergence operator.

We solve the equilibrium equations represented by the displacement,
called Navier-Lamé equations \cite{the_foundation_of_solid_mechanics_feng} (LN equations), which is the displacement
solution method. LN equations can be written in the form 
\begin{equation}
	\begin{cases}
		Gu_{i,jj}+(\lambda+G)u_{j,ji}+f_{i}=0 & \boldsymbol{x}\in\Omega\\
		u_{i}=\bar{u}_{i} & \boldsymbol{x}\in\partial\Omega,
	\end{cases}\label{eq:the LN equations}
\end{equation}
where the comma notation ``,'' indicates partial derivatives, such
as $u_{x,x}=\partial u_{x}/\partial x$, and the subscripts $i,j$
conforms to the Einstein notation, such as $\sigma_{kk}=\sigma_{xx}+\sigma_{yy}+\sigma_{zz}$
and $u_{i}$ means $i$ can be \{x, y, z\}. $\lambda$ and $G$ are
Lame parameters
\begin{equation}
	\begin{cases}
		\lambda=\frac{\nu E}{(1+\upsilon)(1-2\upsilon)}\\
		G=\frac{E}{2(1+\upsilon)},
	\end{cases}
\end{equation}
where $E$ and $\upsilon$ represent the Young’s modulus and Poisson’s ratio respectively. 
The solution for the displacement variable $\boldsymbol{u}$ can be obtained by solving \Cref{eq:the LN equations} subject to well-defined boundary conditions. This approach is referred to as the displacement method, because it exclusively considers the displacement variable \( u \) as the solution variable.
 When only displacement boundary conditions are applied, the displacement method is preferred over the stress method.

The stress method is an approach in which stresses $\boldsymbol{\sigma}$ are the primary unknown quantities to be determined.  We substitute the constitutive
equations into Saint-Venant compatibility equations \cite{the_foundation_of_solid_mechanics_feng}
\begin{equation}
	\nabla\times\boldsymbol{\boldsymbol{\varepsilon}}\times\nabla=0\label{eq:compatibility equations},
\end{equation}
where $\nabla\times$ denotes the curl operator.
In the constitutive
equations, strain is expressed in terms of stress 
\begin{equation}
	\varepsilon_{ij}=\frac{1+\upsilon}{E}\sigma_{ij}-\frac{\upsilon}{E}\sigma_{kk}\delta_{ij},\label{eq:strain-stress}
\end{equation}
where $\delta_{ij}$ is the Kronecker delta, which means $\delta_{ij}=1$, only when $i=j$ (not summed). The compatibility equations 
guarantee single-valued and continuous displacement. We can rewrite \Cref{eq:compatibility equations}
in terms of stress $\boldsymbol{\sigma}$ as
\begin{equation}
	\nabla^{2}\sigma_{ij}+\frac{1}{1+\upsilon}\theta_{,ij}=-(f_{i,j}+f_{j,i})-\frac{\upsilon}{1-\upsilon}\delta_{ij}f_{k,k},\label{eq:BM equations}
\end{equation}
where $\theta=\sigma_{kk}$.
\Cref{eq:BM equations} are known as the Beltrami-Michell compatibility
equations (BM equations). Note that the BM equations are dependent,
which means the independent equations are less than 6 if the problem
is 3D. Because the number of the basic unknowns (stresses) is 6, we
need to add the additional equations, i.e., the equilibrium equations
in terms of stress, to make the equation closed. Therefore, the complete set of equations for the elastostatics problem in terms of stress solutions can be summarized as follows:
\begin{equation}
	\begin{cases}
		\nabla^{2}\sigma_{ij}+\frac{1}{1+\upsilon}\theta_{,ij}+(f_{i,j}+f_{j,i})+\frac{\upsilon}{1-\upsilon}\delta_{ij}f_{k,k}=0 & \boldsymbol{x}\in\Omega\\
		\sigma_{ij,i}+f_{j}=0 & \boldsymbol{x}\in\Omega\\
		\sigma_{ij}n_{j}=\bar{t}_{i} & \boldsymbol{x}\in\partial\Omega,
	\end{cases},\label{eq:BM equtaions}
\end{equation}
where we only consider the Neumann boundary conditions (specified
surface tractions) because the stress solution is quite difficult to
deal with displacement boundary conditions due to boundary integrals. 

\begin{figure}
	\begin{centering}
		\includegraphics[scale=1.8]{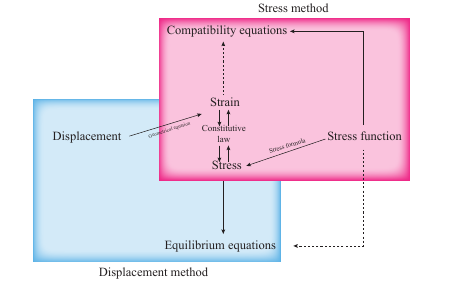}
		\par\end{centering}
	\caption{The relation between displacement and stress solution. The black solid line arrow is the derivation process;    the dotted arrow indicates that the equation is automatically satisfied \label{fig:the relation of different solution}}
\end{figure}

To solve the non-independence problem of the stress solution
in \Cref{eq:BM equtaions}, we can employ certain techniques,
i.e., Bianchi Identity \cite{the_foundation_of_solid_mechanics_feng}
\begin{equation}
	\begin{split}
		L_{ij,j} & =0\\
		L_{ij} & =e_{ink}e_{jml}\Phi_{mn,kl},
	\end{split}
\end{equation}
where $e_{ijk}$ is the permutation symbol. Since the characteristic of $L_{ij}$ satisfy
the similar form of equilibrium equation,  the stress function $\boldsymbol{\Phi}$
can be used to be the basic unknown variable
instead of stress. Then, \Cref{eq:BM equtaions}
in the absence of body forces can be expressed 
as:
\begin{equation}
	\begin{cases}
		\nabla^{2}(e_{ink}e_{jml}\Phi_{mn,kl})+\frac{1}{1+\upsilon}(e_{pnk}e_{pml}\Phi_{mn,kl}){}_{,ij}=0 & \boldsymbol{x}\in\Omega\\
		(e_{ink}e_{jml}\Phi_{mn,kl})n_{j}=\bar{t}_{i} & \boldsymbol{x}\in\partial\Omega
	\end{cases},\label{eq:stress equtaions 3d}
\end{equation}
where the stress function $\Phi_{ij}$ (symmetry) has 3 independent
components in three-dimensional elasticity. \Cref{eq:stress equtaions 3d}
above is the stress function method, which is the other form of the stress
solution. There are $C_{6}^{3}-3=17$ possible choices of the stress function.
The reason for subtracting 3 is that there are 3 combinations
that automatically satisfy \Cref{eq:stress equtaions 3d}. The most
common choices of the stress function are Maxwell and Morera stress functions,
\begin{equation}
	\Phi_{Maxwell}=\left[\begin{array}{ccc}
		\Phi_{11} & 0 & 0\\
		& \Phi_{22} & 0\\
		Sym &  & \Phi_{33}
	\end{array}\right];\Phi_{Morera}=\left[\begin{array}{ccc}
		0 & \Phi_{12} & \Phi_{13}\\
		& 0 & \Phi_{23}\\
		Sym &  & 0
	\end{array}\right]. \label{eq:Airy_Morera}
\end{equation}
We focus on the most common two-dimensional problems in
solid mechanics, i.e., the plane state of stress or strain and torsion
problems. \Cref{fig:the relation of different solution} shows the relation between displacement and stress solution.

\subsubsection{Airy stress function}\label{sec: Airy stress}
In the two-dimensional stress and strain problems, the Maxwell stress
function in  \Cref{eq:Airy_Morera} can be reduced to the Airy stress function, i.e., $\Phi_{11}=\Phi_{22}=0$, $\Phi_{33}\neq0$:
\begin{equation}
	\Phi_{Airy}=\left[\begin{array}{ccc}
		0 & 0 & 0\\
		& 0 & 0\\
		Sym &  & \Phi_{33}
	\end{array}\right].
\end{equation}
Airy stress function is a special and simple form of the Maxwell stress
function for plane problems. Only the stress function component
$\Phi_{33}$ of the Maxwell stress function is not zero, so we use
Airy function $\phi_{A}$ to replace the only Maxwell stress function
not equal to zero. Thus, the \Cref{eq:stress equtaions 3d} can be
simplified to 
\begin{equation}
	\begin{cases}
		\nabla^{2}\nabla^{2}\phi_{A}=-(1-\upsilon)\nabla^{2}V & \boldsymbol{x}\in\Omega\\
		l(\frac{\partial^{2}\phi_{A}}{\partial y^{2}}+V)-m(\frac{\partial^{2}\phi_{A}}{\partial x\partial y})=\bar{t}_{x} & \boldsymbol{x}\in\partial\Omega\\
		-l(\frac{\partial^{2}\phi_{A}}{\partial x\partial y})+m(\frac{\partial^{2}\phi_{A}}{\partial x^{2}}+V)=\bar{t}_{y} & \boldsymbol{x}\in\partial\Omega
	\end{cases},\label{eq: The Airy equations}
\end{equation}
where $l$ and $m$ are the direction cosines of the outer normal
to the boundary curve; $\bar{t}_{x}$ and $\bar{t}_{y}$ are
the surface tractions acting on the boundary surface; $\nabla^{2}\nabla^{2}$
is the biharmonic operator, which is expressed in the Cartesian coordinate
system as :
\begin{equation}
	\nabla^{2}\nabla^{2}=\frac{\partial}{\partial x^{4}}+2\frac{\partial}{\partial x^{2}\partial y^{2}}+\frac{\partial}{\partial y^{4}}.
\end{equation}
Note $V$ is the potential of the body force, and the relationship between $V$ and body force is
\begin{equation}
	f_{x}=-\frac{\partial V}{\partial x};f_{y}=-\frac{\partial V}{\partial y},
\end{equation}
and 
the Airy stress function can only solve
problems with the potential of the body force, and cannot directly
deal with displacement boundary conditions. If the problem has
displacement boundary conditions, we can use Saint Venant's
theory to transform the displacement (Dirichlet) boundary into the force boundary
condition. The stress is expressed by the Airy stress function as
\begin{equation}
	\begin{cases}
		\sigma_{xx}=\frac{\partial^{2}\phi_{A}}{\partial y^{2}}+V\\
		\sigma_{yy}=\frac{\partial^{2}\phi_{A}}{\partial x^{2}}+V\\
		\sigma_{xy}=-\frac{\partial^{2}\phi_{A}}{\partial x\partial y}
	\end{cases}.
\end{equation}

\begin{figure}
	\begin{centering}
		\includegraphics{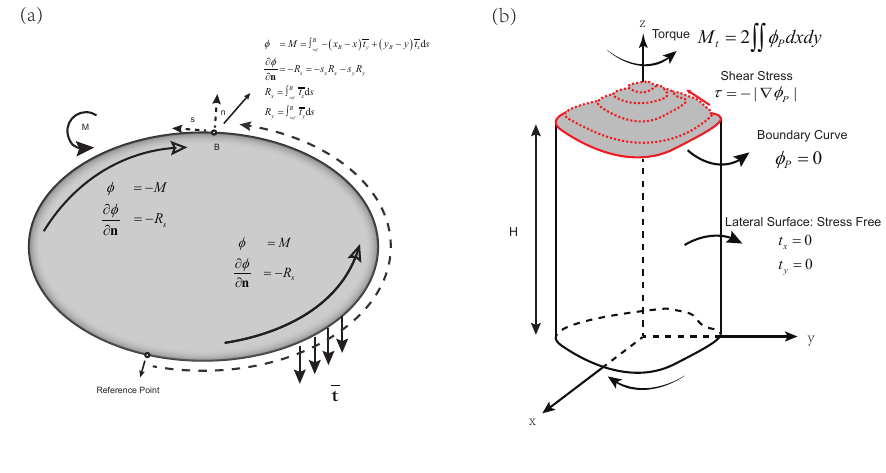}
		\par\end{centering}
	\caption{The illustration of Airy and Prandtl stress function: (a) The value
		of the Airy stress function: the reference point means $\phi_{A}=\partial\phi/\partial x=\partial\phi/\partial y=0$.
		$\bar{\boldsymbol{t}}$ is the surface force. $\boldsymbol{n}$ and
		$\boldsymbol{s}$ is the normal and tangential direction, respectively.
		$R_{x}$ ($R_{y}$) means the integral of the surface force in the
		x (y) direction at the boundary from the reference point to the current
		position B. $M$ is the moment of the current position with respect to the reference point at the boundary. Counterclockwise is
		the positive direction of the moment. 
		(b) The value of the Prandtl stress function: the red lines represent
		the contours of the Prandtl stress function. The value of shear stress
		is equal to the negative gradient of the Prandtl stress function.
		The direction of the shear stress points is the tangent direction
		of the contour of the Prandtl stress function. The Prandtl stress
		is equal to the constant (usually taken as zero) on the boundary. \label{fig:The-value-of stress function}}
\end{figure}

The Airy stress function has the following important properties :
\begin{enumerate}
	\item A difference of a linear function in Airy stress function does not
	affect its solution, as
	the lowest derivative order of \Cref{eq: The Airy equations} is
	two without body force, so the linear function will not affect the
	solution result. For example, if $\phi_{A}^{(1)}-\phi_{A}^{(2)}=kx+b$, $\nabla^{2}\nabla^{2}(\phi_{A}^{(1)}-\phi_{A}^{(2)})=0$.
	\item The value of the Airy stress function
	is the moment of the current position with respect to the reference point at the boundary, and the directional derivative of
	the Airy stress function is the integral of the external force vector.
	This property is significant to guess the formation of the Airy stress
	function, as we explain in \Cref{fig:The-value-of stress function}a.
\end{enumerate}

\subsubsection{Prandtl stress function}
In the torsion problem of a cylindrical body, the Morera stress
function in \Cref{eq:Airy_Morera} can be reduced to the Prandtl stress function, i.e., $\Phi_{12}=\Phi_{13}=0$,
$\partial\Phi_{23}/\partial x\neq0$:
\begin{equation}
	\Phi_{Prandtl}=\left[\begin{array}{ccc}
		0 & 0 & 0\\
		& 0 & \Phi_{23}\\
		Sym &  & 0
	\end{array}\right].
\end{equation}
Prandtl stress function is a special form of the Morera stress function
to deal with the torsion problem of a cylindrical body, as shown in
\Cref{fig:The-value-of stress function}b. Only the stress function
component $\Phi_{23}$ of the Morea stress function is not zero, so
we use Prandtl function $\phi_{P}$ to replace the only Morea stress
function not equal to zero. Thus, \Cref{eq:stress equtaions 3d}
can be simplified to 
\begin{equation}
	\begin{cases}
		\nabla^{2}\phi_{p}=-2G\alpha & \boldsymbol{x}\in\Omega\\
		M_{t}=2\iint\phi_{P}dxdy & \boldsymbol{x}\in\Omega\\
		\phi_{P}=C & \boldsymbol{x}\in\partial\Omega
	\end{cases},\label{eq: The Airy equations-1}
\end{equation}
where $\alpha$ is the rate of twist, i.e. $\alpha=d\theta/dz$. $G$
is one of the Lame parameters, i.e., the shear modulus, and $M_{t}$ (torque) is the boundary condition at the ends $z=H$.
$M_{t}=2\iint\phi_{P}dxdy$ is the boundary condition at the ends of
the cylinder. $\phi_{P}=C$ is the lateral surface boundary condition
without surface force, and no loss of generality is involved in setting
$\phi_{p}=0$ in a connected region ($\phi_{p}=0$ in a connected
region by default unless otherwise stated). Note that $\alpha$ is an
unknown constant, so we usually solve 
\begin{equation}
	\begin{cases}
		\nabla^{2}\phi_{p}=-2G\alpha & \boldsymbol{x}\in\Omega\\
		\phi_{P}=0 & \boldsymbol{x}\in\partial\Omega
	\end{cases}\label{eq: The Airy equations_at_first}
\end{equation}
first with $\alpha$. We can readily get $\phi_{p}$ with $\alpha$,
because the equation is the standard Poisson equation (there are many
well-establish methods in potential theory \cite{kellogg1953foundations}
to help us to solve it). Then we substitute $\phi_{p}$ with  $\alpha$ into $M_{t}=2\iint\phi_{P}dxdy$, and we can solve
for $\alpha$ eventually.
The relation between Prandtl stress function and the stress in the Cartesian coordinate system can be expressed as:
\begin{equation}
	\begin{cases}
		\tau_{zx} = \frac{\partial \phi_{p}}{\partial y}   \\
		\tau_{zy}  =- \frac{\partial \phi_{p}}{\partial x}.   \\
	\end{cases}
\end{equation}

The Prandtl stress function has important properties as shown
in \Cref{fig:The-value-of stress function}b. We can use the minimum complementary energy variation principle to convert it into the corresponding energy form. In the next section, we introduce the principle of minimum complementary energy we use.

\section{Method} \label{sec: method}
In this section, we introduce the principle of minimum complementary energy. 
The proposed DCEM mainly uses neural networks to fit the admissible stress function field. 
However, note that constructing the admissible Airy stress function field is more intricate compared to constructing the admissible displacement field based on the principle of minimum potential energy.

\subsection{Introduction to the principle of minimum complementary energy} \label{sec: minimum complementary energy}

So far, the displacement and stress solutions in \Cref{sec: introduction to stress function} are in strong forms.   From the variational formulation, we can get the energy form of the stress solution. The traditional deep energy form is based on the principle of minimum potential energy. However, there is another important energy principle in solid mechanics called the principle of minimum complementary energy. 

The minimum complementary energy principle is a variational principle with stress function or stress as the basic unknown variable. The complementary energy $\Pi_{c}$ is composed of the complementary strain energy $U_{c}$ and the complementary potential $V_{c}$:

\begin{equation} 
	\Pi_{c}=U_{c}+V_{c}.\label{eq:=006700=005C0F=004F59=0080FD=00539F=007406}
\end{equation}
The  complementary strain energy $U_{c}$ is the integrated overall energy of the  complementary strain energy density $W_{c}$ on the solution domain:

\begin{equation}
	U_{c}=\int_{\Omega}W_{c}d\Omega
\end{equation}

The complementary strain energy density $W_{c}$ is the energy density in the integral form of stress as the basic unknown variable:
\begin{equation}
	W_{c}=\int_{\sigma_{ij}}\varepsilon_{ij}d\sigma_{ij}.
\end{equation}
The complementary potential $V_{c}$ is expressed as  
\begin{equation}
	V_{c}=-\int_{\Gamma^{u}}\bar{u}_{i}p_{i}d\Gamma,
\end{equation}
where its calculation method is the given essential displacement boundary condition $\boldsymbol{\bar{u}}$ multiplied by the constraint reaction force $\boldsymbol{p}=\boldsymbol{n}\cdot\boldsymbol{\sigma}$ with the stress field as the basic unknown quantity, and its physical meaning is the excess energy absorbed by the support system or transmitted to other objects by the support system.

The admissible stress field must satisfy the given force boundary conditions and equilibrium equations 
\begin{equation}
	\begin{cases}
		\nabla\cdot\boldsymbol{\sigma}+\boldsymbol{f}=0\\
		\boldsymbol{\sigma}\cdot\boldsymbol{n}=\bar{\boldsymbol{t}},
	\end{cases},\label{eq:equail equations}
\end{equation}
in advance. However, it is challenging to construct the admissible stress field in advance because the equilibrium equations are not easy to satisfy. Thus, it is common to use the stress function method to replace the stress field because the stress function naturally satisfies the equilibrium equations. Only the force boundary conditions instead of equilibrium equations should be considered. We put  admissible stress function $\phi_{admiss}$ into  \Cref{eq:=006700=005C0F=004F59=0080FD=00539F=007406}, and optimize the complementary energy functional to minimize it:
\begin{equation}
	\phi_{true}=\argmin_{\phi_{admiss}}\Pi_{c}(\phi_{admiss}).
\end{equation}

\subsection{DCEM: Deep complementary energy method based on the principle of minimum complementary energy}\label{sec: DCEM_method}
Unlike the minimum potential energy using the neural network to approximate the displacement field, we use a neural network to approximate the stress function. Then the complementary energy is minimized by the principle of minimum complementary energy. 
Unfortunately,  not all stress functions satisfy the conditions of the given force boundary conditions in advance. 
Therefore, if we use the principle of minimum complementary energy,  admissible stress functions that satisfy the force boundary conditions must be constructed in advance. Therefore, the key to the DCEM lies in the construction of admissible stress functions. The expressions of the admissible stress functions can be written as:
\begin{equation}
	\phi_{i}(\boldsymbol{x})=NN_{p(i)}(\boldsymbol{x})+\sum_{m=1}^{n}NN^{(m)}_{g(i)}(\boldsymbol{x})*NN^{(m)}_{b(i)}(\boldsymbol{x}), \label{eq:distance funciton of the admissible stress function}
\end{equation}
where $NN_{p}$ is the particular solution network only satisfying the non-homogeneous force boundary condition as $\boldsymbol{x}$ at the force boundary condition; $NN_{g}$ is the generalized network, which is an unconstrained neural network, because the construction of the admissible stress functions in \Cref{eq:distance funciton of the admissible stress function} can satisfy the boundary constraints. $NN_{b}$ is the basis function network related to different stress functions such as the Airy and Prandtl stress functions. The three types are analogous to boundary networks, generalized networks, and distance networks, respectively, of admissible displacement fields in \cite{PINNstrong_form_in_elastodynamics, complex_PINN_a_method_to_construct_admissible_function, admissible_in_PINN_energy_form}. It is worth noting that distance networks are a special case of basis function networks. 
Any function can be used as the distance function as long as the distance function is zero at the given displacement boundary in the admissible displacement field. Thus, the distance function is not unique, which will be discussed in detail in \Cref{sec:Appendix-C.basis function}. It is possible to use different distance functions to construct the admissible field to decrease the approximation error theoretically \cite{DeepOnet}.
In the Prandtl stress function, the construction of the admissible stress function is the same as the construction of the displacement field, while the construction of the admissible Airy stress function is more complicated than the construction of admissible displacement. The admissible displacement field should consider the Dirichlet boundary condition (the displacement boundary), while the admissible stress function should consider the Neumann boundary condition (the force boundary).
But in the Airy stress function, not only the basis function is zero at the given force boundary, but also the normal derivative of the basis function is required to be zero at the force boundary, which is shown in \Cref{sec:Appendix-C.the basis function in Airy stress function}. Compared with the admissible displacement field, the basis function in the Airy stress function has the extra condition that the normal derivative is zero, mainly due to the relationship between the stress function and the stress. As a result, the basis function is related to different types of stress functions, such as the Airy and Prandtl stress functions. 

In the Airy stress function, the basis function can be obtained from the distance function, defined as:
\begin{equation}
	NN_{b} =\lambda*Dis^{order},
\end{equation}
where
\begin{equation}
	\begin{cases}
		Dis(x) & =0,x\in\Gamma^{t}\\
		Dis(x) & \neq0,x\in\Omega
	\end{cases},
\end{equation}
$\Gamma^{t}$ represents the boundary of the domain, and $\Omega$ is the domain itself. The parameter $order$ is a real number greater than 1 (in this example, $order=2$), and $\lambda$ is a scalable factor ($\lambda = 1/max(Dis^{order})$) used for the normalization of the basis function. The construction of the basis function ensures that it satisfies the conditions $B(\boldsymbol{x})=0$ and $\partial B(\boldsymbol{x})/\partial\boldsymbol{n}=0$ when $\boldsymbol{x}\in\Gamma^{t}$, because:
\begin{align}
	\frac{\partial NN_{b}}{\partial\boldsymbol{n}}=\lambda*order*Dis^{order-1}\frac{\partial Dis}{\partial\boldsymbol{n}}=0, & x\in\Gamma^{t}.
\end{align}
This property ensures that the basis function satisfies the natural boundary conditions, making it suitable for the basis function in the Airy stress function, as is shown in \Cref{sec: Wedge}.

	There are many ways of constructing an admissible stress function in theory, such as the Coefficient function method and \Cref{eq:distance funciton of the admissible stress function} (Basis function method). The coefficient function method is to construct many admissible basis function fields (such as the orthogonal polynomials, i.e., Legendre polynomials, Chebyshev polynomials, periodic polynomials) satisfying the homogeneous boundary conditions in the given force boundary condition, then fit the constants in front of the basis functions. \Cref{eq:distance funciton of the admissible stress function} is based on the idea of distance function, but it needs to meet the condition that the normal derivative is zero additionally in the Airy stress function, as explained in \Cref{sec:Appendix-C.the basis function in Airy stress function}. Of course,  the "omnipotent" penalty function method can also be used to construct the admissible stress function. \Cref{tab:The-different-methods} shows different methods for constructing admissible stress functions.

\begin{table}	
	\caption{Different methods for constructing admissible stress function in Airy stress function.
		$\Gamma^{t}$ means the force boundary condition. Homogeneous force
		boundary condition (HFB) means when $\boldsymbol{t}(\boldsymbol{x})=0$
		when $\boldsymbol{x}\in\Gamma^{t}$, and non-homogeneous force boundary
		condition (non-HFB) means when $\boldsymbol{t}(\boldsymbol{x})=\bar{\boldsymbol{t}}$
		when $\boldsymbol{x}\in\Gamma^{t}$ \label{tab:The-different-methods}}
	\begin{centering}
		\resizebox{\textwidth}{15mm}{
			\begin{tabular}{cccc}
				\toprule 
				\multirow{2}{*}{Methods in Airy stress function} & \multicolumn{3}{c}{$\phi(\boldsymbol{x})=A(\boldsymbol{x})+B(\boldsymbol{x})*C(\boldsymbol{x})$}\tabularnewline
				\cmidrule{2-4} \cmidrule{3-4} \cmidrule{4-4} 
				& $A(\boldsymbol{x})$ & $B(\boldsymbol{x})$ & $C(\boldsymbol{x})$\tabularnewline
				\midrule 
				Coefficient function method & Satisfy non-HFB  ($\boldsymbol{x}\in\Gamma^{t}$) & Constant trainable parameters & Satisfy HFB when $\boldsymbol{x}\in\Gamma^{t}$\tabularnewline
				Basis function method & Satisfy non-HFB  ($\boldsymbol{x}\in\Gamma^{t}$)  & $B(\boldsymbol{x})=0;\partial B(\boldsymbol{x})/\partial\boldsymbol{n}=0$
				($\boldsymbol{x}\in\Gamma^{t}$) & Neural network\tabularnewline
				Penalty function method & \multicolumn{3}{c}{$\mathcal{L}_{p}=\beta_{1}[\hat{\phi}(\boldsymbol{x})-\phi(\boldsymbol{x})]^{2}+\beta_{2}[\partial\hat{\phi}(\boldsymbol{x})/\partial\boldsymbol{n}-\partial\phi(\boldsymbol{x})/\partial\boldsymbol{n}]^{2}$}\tabularnewline
				\bottomrule
		\end{tabular}} 
		\par\end{centering}	
\end{table}

The algorithm flow of DCEM is:

\begin{enumerate}
	\item According to the force boundary conditions, we construct the admissible stress function. When dealing with the Airy stress function, we can choose one of the methods for constructing the admissible stress function in \Cref{tab:The-different-methods}. If we choose the basic function method in \Cref{tab:The-different-methods}. The particular solution network should be trained, and training points are $\{\boldsymbol{x}_{i}\in\Gamma^{t},\bar{\boldsymbol{t}}_{i} \}_{i=1}^{n}$, so that the particular solution network can fit the force boundary condition. For example, the force boundary condition expressed by the Airy stress function for the plane stress problem is:
	\begin{equation}
		\begin{split}n_{x}(\frac{\partial^{2}NN_{p}}{\partial y^{2}}+V)-n_{y}(\frac{\partial^{2}NN_{p}}{\partial x\partial y}) & =\bar{t}_{x}\\
			-n_{x}(\frac{\partial^{2}NN_{p}}{\partial x\partial y})+n_{y}(\frac{\partial^{2}NN_{p}}{\partial x^ {2}}+V) & =\bar{t}_{y},
		\end{split}\label{eq: particular equations}
	\end{equation}
	and the property of the Airy stress function can be used to construct the particular solution network, i.e., $\phi=M$ and $\partial\phi/\partial\boldsymbol{n}=-R_{s}$, as shown in \Cref{fig:The-value-of stress function} and proof in \Cref{sec:Appendix-B.-proof of the particular solution network}.
	If it is a free torsion problem, the particular solution network must ensure that the Prandtl stress function at the boundary conditions is constant. It is convenient to set the particular solution network $NN_{p}$ to zero when dealing with the connected region. Noting that through the properties of the Airy stress function in \Cref{sec: Airy stress}, it is feasible to obtain the particular solution, which is equivalent to \Cref{eq: particular equations},  as shown in proof  \Cref{sec:Appendix-B.-proof of the particular solution network}.
	
	\item Using the principle of the minimum complementary energy, we get a certain admissible stress function by the AD algorithm in Pytorch when the loss function (complementary energy is obtained by using numerical integration, we use Simpson integration in this paper)   converges.

	\item According to the relation between the stresses and the stress functions, we estimate the stresses by the AD algorithm.
\end{enumerate}

It is worth noting that, according to the minimum complementary energy principle, the admissible stress function needs to satisfy the force boundary condition in advance; according to the minimum potential energy principle, the admissible displacement field in the deep energy method needs to satisfy the displacement boundary condition in advance. Therefore, in theory, if displacement boundary conditions mainly dominate the boundary conditions of this problem, it would be more appropriate to use the principle of minimum complementary energy because there are fewer constraints on the admissible stress function field than the admissible displacement field in the minimum potential energy principle, so the powerful fitting ability of the neural network can be released because the form of the admissible function field through the distance function will limit the function space of the neural network to a certain extent. In contrast, if force boundary conditions dominate the problem, it will be more advantageous to use the principle of minimum potential energy because the assumptions on the displacement function are reduced.

\subsection{DCEM-P: DCEM with biharmonic function in Airy stress function}

The strong form of the Airy stress function satisfies the biharmonic equation 

\begin{equation}
	\nabla^{4}\phi=0. \label{eq:the biharmonic equation_first}
\end{equation}
We can add some terms $\phi_{bh}$ naturally satisfying \Cref{eq:the biharmonic equation_first}, such as $x^{2}$, $y^{2}$. The trainable parameters $a_{i}$ before these terms can be determined by the loss function (the complementary energy $\Pi_{c}$), i.e.

\begin{equation}
	\begin{split}
		\phi^{true}  =&\argmin_{a_{i}, \boldsymbol{\theta}_{NN^{(m)}_{g}}}\Pi_{c}(\phi)\\
		\phi(\boldsymbol{x})  =&NN_{p}(\boldsymbol{x}; \boldsymbol{\theta}_{NN^{(m)}_{p}})+\sum_{m=1}^{n}NN^{(m)}_{g}(\boldsymbol{x};\boldsymbol{\theta}_{NN^{(m)}_{g}})*NN^{(m)}_{b}(\boldsymbol{x};\boldsymbol{\theta}_{NN^{(m)}_{b}}) + \sum_{i=1}^{s}a_{i}*\phi_{bh}^{i}.
	\end{split}
\end{equation}
DCEM-P can converge faster and be more accurate than DCEM  because some terms that pre-satisfy the PDE equations are included (add the inductive bias). Because the DCEM is similar to DCEM-P, we only present the algorithm of DCEM-P in \Cref{alg:DCEM-P}.

\begin{algorithm}
	\caption{The algorithm of DCEM-P}\label{alg:DCEM-P}
	\begin{algorithmic}[1]
		\State \textbf{Step 1}: Training the particular net $NN_{p}$.
		\State  Allocate training points on the force boundary conditions, i.e.,   $\{\boldsymbol{x}_{i}\in\Gamma^{t},\bar{\boldsymbol{t}}_{i} \}_{i=1}^{n}$.
		\If{Prandtl stress function: }
		\State $NN_{p} = 0 $ (connected region).
		\EndIf
		\If{Airy stress function:} 
		\State $Loss_{p} = 
		\sum_{i=1}^{n} [l(\partial^{2}NN_{p}(\boldsymbol{x}_{i})/\partial y^{2})-m(\partial^{2}NN_{p}(\boldsymbol{x}_{i}) / \partial x\partial y)- \bar{t}_{xi}]^{2}
		+\sum_{i=1}^{n} [-l(\partial^{2}NN_{p}(\boldsymbol{x}_{i}) / \partial x\partial y)+m(\partial^{2}NN_{p}(\boldsymbol{x}_{i})/\partial x^{2})- \bar{t}_{yi}]^{2}$.
		\State Minimize $Loss_{p}$ to get the optimal parameters $\boldsymbol{\theta}_{p}$ of $NN_{p}$.  
		\EndIf
		\State \textbf{Step 2}: Training the basis net $NN_{b}$.
		\State  Allocate training points $\{\boldsymbol{x}_{i}\in\Gamma^{t}, d_{i}\}_{i=1}^{m}$ on the domain and boundary, where $d_{i}$ is the nearest distance from $\boldsymbol{x}_{i}$ to the force boundary conditions.
		\If{Prandtl stress function: }
		\State $Loss_{b} = 
		\sum_{i=1}^{m}[NN_{b}(\boldsymbol{x}_{i})- d_{i}]^{2}$.
		\EndIf
		\If{Airy stress function:} 
		\State $Loss_{b} = 
		\sum_{i=1}^{m}[NN_{b}(\boldsymbol{x}_{i})- d_{i}]^{2}
		+\sum_{i=1}^{r}[NN_{b}(\boldsymbol{x}_{i})]^{2}
		+\sum_{i=1}^{r}[\partial NN_{b}(\boldsymbol{x}_{i}) / \boldsymbol{n}]^{2}$, where $r$ is the number of points on the force boundary condition and $\boldsymbol{n}$ is the normal direction of the force boundary.
		\EndIf		
		\State Minimize $Loss_{b}$ to get the optimal parameters $\boldsymbol{\theta}_{b}$ of $NN_{b}$.   
		\State \textbf{Step 3}: Select the terms of biharmonic function $\phi_{bh}$ in Airy stress function.  
		\State \textbf{Step 4}: Training the general net $NN_{g}$.
		\State $\phi(\boldsymbol{x})  =NN_{p}(\boldsymbol{x})+NN_{g}(\boldsymbol{x};\boldsymbol{\theta}_{NN_{g}})*NN_{b}(\boldsymbol{x}) + \sum_{i=1}^{s}a_{i}*\phi_{bh}^{i}$.
		\State Minimize the complementary energy and get the optimal parameters of $\boldsymbol{\theta}_{NN_{g}}$ and $a_{i}$ (freeze the $NN_{p}$ and $NN_{b}$).
	\end{algorithmic}
\end{algorithm}

\subsection{DCEM-O: Deep operator energy method based on the principle of minimum complementary energy } \label{sec: DCEM-o}
This idea's starting point is to reuse the existing calculation results fully and not waste the previous calculation results to improve the computational efficiency of DCEM.
The idea of DCEM-O is to use the DeepONet framework to help DCEM make operator learning (DeepONet serves as the framework for our operator learning), as mentioned in \Cref{sec: introduction to DeepONet}. Trunk net can build partial differential operators to construct the physical loss.
The algorithmic flow of DCEM-O is shown in \Cref{alg:DCEM-O}

\begin{algorithm}
	\caption{The algorithm of DCEM-O}\label{alg:DCEM-O}
	\begin{algorithmic}[1]
		\State \textbf{Training:}
		\State \textbf{Data:} The high-fidelity stress function
		$\{\boldsymbol{x}^{j}_{s},\phi_{s}^{j} \}_{s=1}^{n}$
		and variable field
		$\{\boldsymbol{x}^{j}_{i},f^{j}_{i} \}_{i=1}^{m}$ such as geometry, material information as the input of the branch net and $j=1,2,\cdots, t$.
		\State \textbf{Step 1}: Training the branch $NN_{B}$ and trunk net $NN_{T}$. 
		\State  input the variable field
		$\{\boldsymbol{x}^{j}_{i},f^{j}_{i} \}_{i=1}^{m}$ to the  branch net;
		input the $\boldsymbol{x}^{j}_{s}$ to the  trunk net and 
		$\phi_{s}^{j}$ to the output stress function.
		\State Minimize the operation loss.
		\State \textbf{Test:}
		\State \textbf{Step 2}: Training the particular net $NN_{p}$ as DCEM. 
		\State \textbf{Step 3}: Training the basis net $NN_{b}$	
		as DCEM. 
		\State \textbf{Step 4}: Training the branch $NN_{B}$ and trunk net $NN_{T}$ again by minimizing the complementary energy to obtain the optimal parameters $\boldsymbol{\theta}_{B}$ and $\boldsymbol{\theta}_{T}$ (freeze the particular and basis net).
	\end{algorithmic}
\end{algorithm}

\begin{figure}
	\begin{centering}
		\includegraphics[scale=0.75]{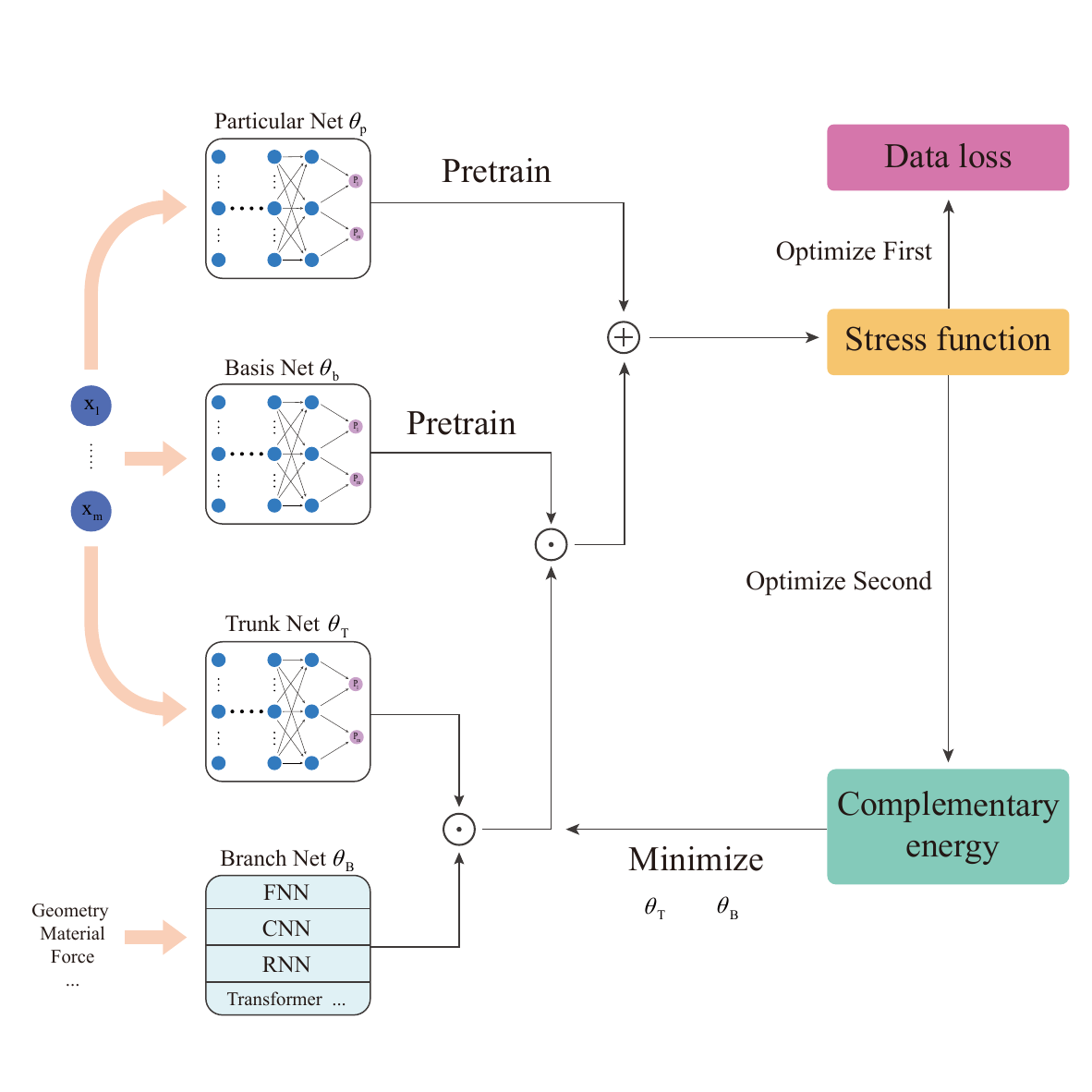}
		\par\end{centering}
	\caption{The schematic of deep operator energy method based on the principle of minimum complementary energy (DCEM-O) \label{fig:DCEM-O}}
\end{figure}

\Cref{fig:DCEM-O} shows the framework of DCEM-O. We first construct the particular solution and basis function network at the force boundary condition, which is the same as the steps of DCEM. We only change the general network to the network structure of DeepONet. The training process is employed in a serial fashion; the training method is first to learn the operator in the existing data, i.e., not to construct the physical-based complementary energy functional first. After training based on the existing data, we consider optimization based on the principle of complementary energy. Our input data in the branch net can include different geometric shapes, materials, and force boundary conditions (other field variables can also be fed into the branch net), and the output is the existing high-fidelity stress function result, such as the stress-function element method. Finally, we optimize the parameters of the trunk and branch net based on the principle of minimum complementary energy.

The key to the operator learning with physical law lies in effectively integrating the principles of physics with data-driven approaches.  Various methods have been proposed to bridge the gap between physical laws and data-driven techniques \cite{goswami2022physics, li2021physics, wang2021learning}.   The AD and numerical differentiation can be used to construct the physical loss. In our study, we opt for the AD algorithm to create the physical loss. This choice is attributed to the fact that finite difference methods (numerical differentiation) necessitate a fine-resolution uniform grid. In the absence of such conditions, numerical errors in derivatives can become magnified in the output solution. As a result, the AD algorithm is preferred for constructing the physical loss.
The hybrid optimization \cite{goswami2022physics, li2021physics, wang2021learning}, involves combining data loss and physical loss into a single loss function for optimization, using different hyperparameters. Specifically, the hybrid optimization strategy can be expressed as:
\begin{equation} 
	\begin{cases}
		\mathcal{L}(\boldsymbol{\theta})=\mathcal{L}_{data}(\boldsymbol{\theta})+\lambda*\mathcal{L}_{phy}(\boldsymbol{\theta})\\
		\mathcal{L}_{data}(\boldsymbol{\theta})=\frac{1}{I*J}\sum_{j=1}^{J}\sum_{i=1}^{I}|NN_{O}(\boldsymbol{y}^{(i)},\boldsymbol{u}^{(j)};\boldsymbol{\theta}))-\boldsymbol{u}^{(j)}(\boldsymbol{y}^{(i)})|^{2}\\
		\mathcal{L}_{phy}(\boldsymbol{\theta})=\frac{1}{I*J}\sum_{j=1}^{J}\sum_{k=1}^{K}|Phy(NN_{O}(\boldsymbol{x}^{(k)},\boldsymbol{u}^{(j)};\boldsymbol{\theta}))|^{2} 
	\end{cases}, \label{eq:DEEPONET_PHY}
\end{equation}
where the details of \Cref{eq:DEEPONET_PHY} are elaborated in \Cref{sec: introduction to DeepONet}.
This approach combines data loss (operator loss) and physical loss with different hyperparameters into the same loss function for optimization together. 
However, in our research, we have chosen a sequential optimization strategy instead of hybrid optimization. This choice is due to the challenge of selecting appropriate hyperparameters. We first optimize the data loss and subsequently optimize the physical loss.

\section{Result}\label{sec: result}
In this section, we introduce several benchmark problems including Airy and Prandtl stress function to assess the robustness and effectiveness of the DCEM. Therefore, we will compare the solutions obtained using DCEM with those from DCM and DEM, using the reference solution from FEM as a basis for comparison when analytical solutions are not available. Although the geometry and boundary conditions may not be complex, the FEM solution is considered to represent an accurate approximation of the exact solution.

It is worth noting that DCEM-O first trains the network parameters using a large dataset. Then, for a specific physical problem, it provides an initial solution and fine-tunes the parameters based on physical equations. For another specific physical problem, DCEM-O uses the network parameters trained on the large dataset and does not use the parameters optimized from the previous specific physical problem, thus avoiding catastrophic forgetting \cite{goodfellow2013empirical}.

\subsection{Airy stress function}

\subsubsection{Circular tube: full displacement boundary conditions} \label{sec: circular}
The circular tube is relevant to the study of thick pressure vessels, which are commonly used as containers for gas and storage in engineering \cite{vullo2014circular}.
If all the boundary conditions are force boundary conditions, the admissible displacement is not required in DEM. In this example, all the boundary conditions are displacement boundary conditions. As a result,  the admissible stress function is not required in DCEM. The DEM can also be used to solve the full displacement boundary problem, but we have to require strong assumptions about admissible displacement fields.

\begin{figure}
	\begin{centering}
		\includegraphics{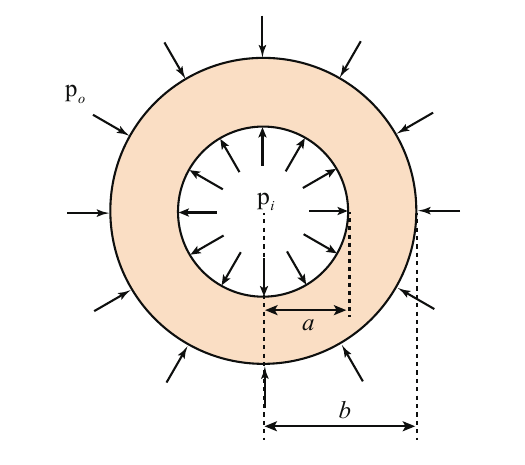}
		\par\end{centering}
	\caption{Schematic diagram of the circular tube:  the value of parameters are $p_{o}=10$,  $p_{i}=5$, $a=0.5$,  $b=1.0$,  Young's modulus $E=1000$, and Poisson's ratio $\nu=0.3$ \label{fig:the schematic of circular tube}}
\end{figure}

As is shown in \Cref{fig:the schematic of circular tube}, we utilize \Cref{eq:lame analysical solution} to convert the full pressure boundary conditions to the full displacement boundary conditions. The transformation enables a comparison of DEM and DCEM performance, highlighting varied advantages and disadvantages associated with different energy principles under full displacement boundary conditions. The DCEM  does not need to assume the admissible stress function because there are no force boundary conditions. We use the displacement solution of the famous Lame's formula to construct the displacement boundary condition. The analytical solution of the Lame's stress and displacement is:

\begin{equation}
	\begin{split}
		\sigma_{r} & =\frac{a^{2}}{b^{2}-a^{2}}(1-\frac{b^{2}}{r^{2}})p_{i}-\frac{b^{2}}{b^{2}-a^{2}}(1-\frac{a^{2}}{r^{2}})p_{o}\\
		\sigma_{\theta} & =\frac{a^{2}}{b^{2}-a^{2}}(1+\frac{b^{2}}{r^{2}})p_{i}-\frac{b^{2}}{b^{2}-a^{2}}(1+\frac{a^{2}}{r^{2}})p_{o}\\
		u_{r} & =\frac{1}{E}[\frac{(1-\nu)(a^{2}p_{i}-b^{2}p_{o})}{b^{2}-a^{2}}r+\frac{(1+\nu)a^{2}b^{2}(p_{i}-p_{o})}{b^{2}-a^{2}}\frac{1}{r}]\\
		u_{\theta} & =0, \label{eq:lame analysical solution}
	\end{split}
\end{equation}
where $a$ and $b$ are inner and outer radius respectively, $E$ and $\nu$ are Young's  modulus and Poisson's ratio, $p_{i}$ and $p_{o}$ are  the uniform pressure at $r=a$ and $r=b$. The compressive stress is positive. 

To use the DCEM  to solve the problem, we need to express the complementary strain energy and complementary potential in the form of stress function. The stresses expressed by the stress function of the axisymmetric problem in polar coordinates are:
\begin{equation}
	\begin{split}
		\sigma_{r}  =\frac{1}{r}\frac{\partial\phi}{\partial r}\\
		\sigma_{\theta} =\frac{\partial^{2}\phi}{\partial r^{2}}.
	\end{split}
\end{equation}

According to the constitutive law, the strains in polar coordinates are:
\begin{equation}
	\begin{aligned}\varepsilon_{r} & =\frac{1}{E}(\sigma_{r}-\nu\sigma_{\theta})=\frac{1}{E}(\frac{1}{r}\frac{\partial\phi}{\partial r}-\nu\frac{\partial^{2}\phi}{\partial r^{2}})\\
		\varepsilon_{\theta} & =\frac{1}{E}(\sigma_{\theta}-\nu\sigma_{r})=\frac{1}{E}(\frac{\partial^{2}\phi}{\partial r^{2}}-\nu\frac{1}{r}\frac{\partial\phi}{\partial r})
	\end{aligned}
\end{equation}

The complementary strain energy is expressed as a stress function:
\begin{equation}
	\begin{split}
		W_{c} & =\int_{\Omega}\frac{1}{2}\sigma_{ij}\varepsilon_{ij}d\Omega=\int_{\Omega}\frac{1}{2}(\sigma_{r}\varepsilon_{r}+\sigma_{\theta}\varepsilon_{\theta})2\pi rdr\\
		& =\int_{\Omega}\frac{1}{2}[\frac{1}{r}\frac{\partial\phi}{\partial r}\frac{1}{E}(\frac{1}{r}\frac{\partial\phi}{\partial r}-\nu\frac{\partial^{2}\phi}{\partial r^{2}})+\frac{\partial^{2}\phi}{\partial r^{2}}\frac{1}{E}(\frac{\partial^{2}\phi}{\partial r^{2}}-\nu\frac{1}{r}\frac{\partial\phi}{\partial r})]2\pi rdr
	\end{split}
\end{equation}

The complementary potential is:
\begin{equation}
	\begin{split}
		V_{c} & =u_{r}\sigma_{r}2\pi r|_{r=b}-u_{r}\sigma_{r}2\pi r|_{r=a}\\
		& =u_{r}\frac{\partial\phi}{\partial r}2\pi|_{r=b}-u_{r}\frac{\partial\phi}{\partial r}2\pi|_{r=a}
	\end{split}
\end{equation}
\begin{figure}
	\begin{centering}
		\includegraphics[scale=1.0]{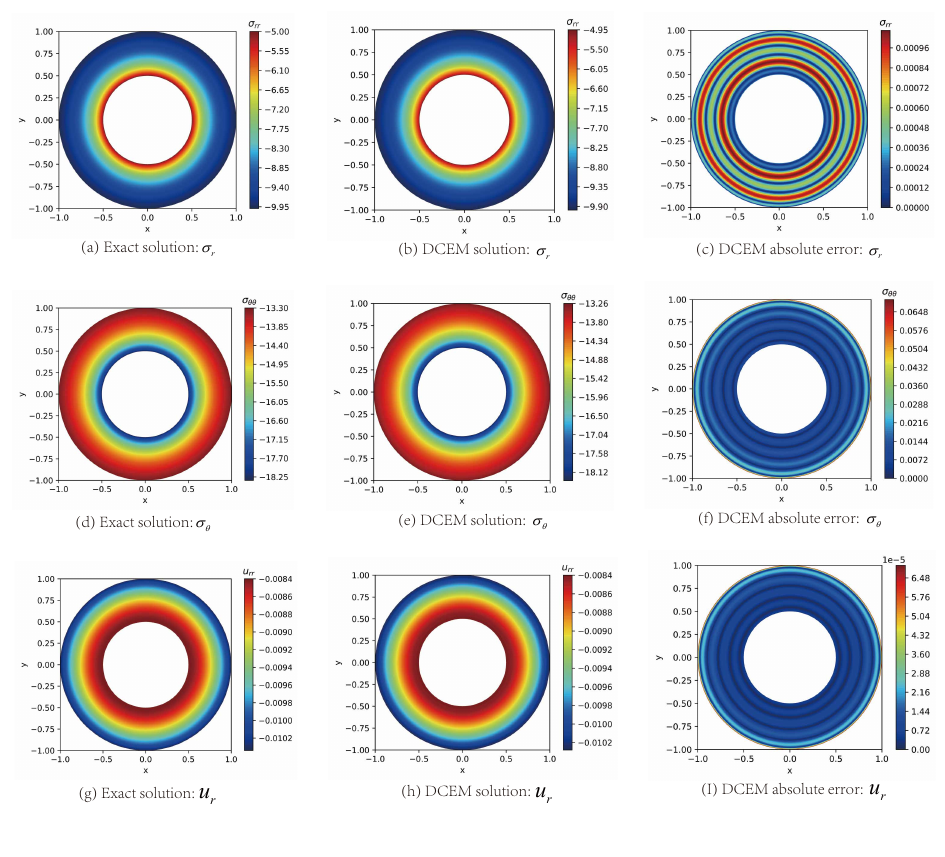}
		\par\end{centering}
	\centering{}\caption{The result of stresses $\sigma_{r}$, $\sigma_{\theta}$ and displacement $u_{r}$ of the circular tube by DCEM: the exact solutions of $\sigma_{r}$ (a), $\sigma_{\theta}$ (d) and $u_{r}$ (g); the predictions of DCEM in $\sigma_{r}$ (b), $\sigma_{\theta}$ (e) and $u_{r}$ (h); the absolute errors of DCEM in $\sigma_{r}$ (c), $\sigma_{\theta}$ (f) and $u_{r}$ (i). \label{fig:DCEM result of circular tube}}
\end{figure}

To research the accuracy of  DCEM, the analytical solution is brought into the complementary strain energy $W_{c}=79\pi/960$ and complementary potential $V_{c}=79 \pi/480$ ($p_{o}=10$,  $p_{i}=5$, $a=0.5$,  $b=1.0$,  Young's modulus $E=1000$, and Poisson's ratio $\nu=0.3$). \Cref{fig:DCEM result of circular tube} shows the stress prediction of DCEM compared with the analytical solution. The points are uniformly distributed. This problem is axisymmetric so the problem can be reduced to a one-dimensional problem. Thus, the number of the input and output neurons are both 1 (the input is $r$, the output is Airy stress function), 3 hidden layers, each layer 20 neurons, optimization function Adam, 2000 iterations. The absolute error Abs(Pred-Exact) of $\sigma_{r}$ is smaller than $\sigma_{\theta}$, the primary error of $\sigma_{r}$ is on the inner ring boundary, the primary error of $\sigma_{\theta}$ is on inner and outer ring boundary.

\begin{figure}
	\begin{centering}
		\includegraphics[scale=1.0]{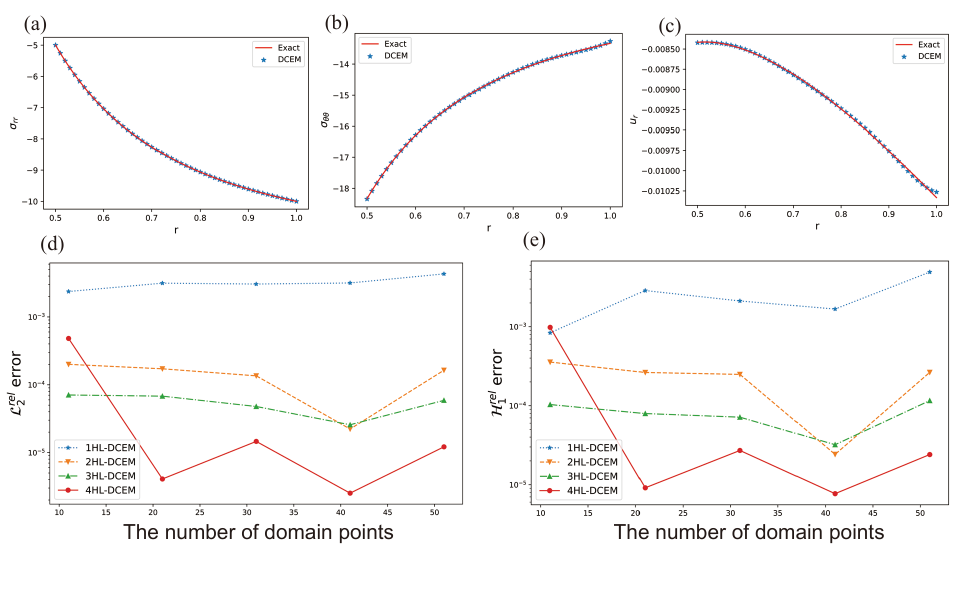}
		\par\end{centering}
	\centering{}\caption{The stresses $\sigma_{r}$ (a), $\sigma_{\theta}$ (b) and $u_{r}$ (c) of the circular tube in the radial direction under prediction solution of the DCEM compared with analytical solution. The errors in terms of $\mathcal{L}_{2}^{rel}$ norm (d) and $\mathcal{H}_{1}^{rel}$  seminorm (e) of DCEM with respect to the training steps in Airy stress function. 	Four different neural networks were constructed, each having a different number of hidden layers (HL): 1HL, 2HL, 3HL, and 4HL. Each hidden layer consists of 30 neurons. The number of domain points is the number of uniformly distributed points in the radius.  The neural network's ability to generalize may not improve with more points if the network architecture, learning rate, or other hyperparameters are not well-tuned. Also, neural networks often have complex loss landscapes with many local minima due to the non-convex nature.\label{fig:the radial directio of circular cube}}
\end{figure}

\Cref{fig:the radial directio of circular cube} shows the comparison between the stress and the analytical solution in the radial direction.
The $\mathcal{L}_{2}^{rel}$ relative error norm and $\mathcal{H}_{1}^{rel}$ relative error seminorm in DCEM are calculated as follows
\begin{equation}
	\begin{split} \Vert e \Vert_{\mathcal{L}_{2}^{rel}} &= \frac{\Vert \phi^{pred} - \phi^{exact}\Vert_{\mathcal{L}_{2}}}{\Vert \phi^{exact}\Vert_{\mathcal{L}_{2}}} =  \frac{\int_{\Omega}(\phi^{pred} - \phi^{exact})^2 dr}{\int_{\Omega} (\phi^{exact})^{2} dr} \\
		\Vert e \Vert_{\mathcal{H}_{1}^{rel}} &= \frac{\Vert \phi^{pred} - \phi^{exact}\Vert_{\mathcal{H}_{1}}}{\Vert \phi^{exact}\Vert_{\mathcal{H}_{1}}} =  \frac{\int_{\Omega}( E^{pred} - E^{exact})^2 dr}{\int_{\Omega} (E^{exact})^{2} dr} \\
		E &= \int_{\Omega} (\frac{\partial \phi}{\partial r})^2 dr.
	\end{split}
\end{equation}
 The $\mathcal{L}_{2}^{rel}$ errors  of $\sigma_{r}$, $\sigma_{\theta}$ and $u_{r}$ are $\sim10^{-5}$, $\sim10^{-4}$, and $\sim10^{-3}$ respectively. The error of $\sigma_{r}$ is smaller than $\sigma_{\theta}$, mainly because $\sigma_{r}$ is the first derivative of the stress function, but $\sigma_{\theta}$ is the second derivative of the stress function. With increasing derivative order, the error in general tends to become larger. More importantly, the relative error of the stress is less than the relative error of the displacement in DCEM. In FEM, stress-based FEM provides more accurate stress results compared to displacement-based FEM, aligning with the conclusion drawn from the comparison between DEM and DCEM.  We also compute the $\mathcal{H}_{1}^{rel}$ and $\mathcal{L}_{2}^{rel}$ in a different number of domain points and neural network in \Cref{fig:the radial directio of circular cube}d,e. 
The results show that all neural network architectures exhibit promising results compared to the exact solution. However, it is important to note that all methods do not exhibit convergence concerning the number of hidden layers, i.e., adding more hidden layers doesn't consistently lead to improved accuracy. Increasing the number of points does not necessarily enhance the neural network's ability to generalize, especially if the network architecture, learning rate, or other hyperparameters are not optimally tuned. Furthermore, due to their non-convex nature, neural networks often exhibit complex loss landscapes with numerous local minima. Because all boundaries are displacement boundaries, the admissible stress function does not need to be constructed in advance.

\begin{figure}
	\begin{centering}
		\includegraphics[scale=1.5]{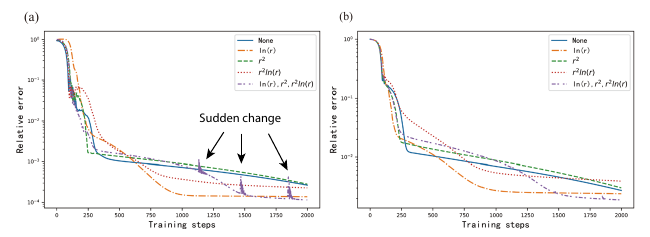}
		\par\end{centering}
	
	\centering{}\caption{The overall relative $\mathcal{L}_{2}^{rel}$ errors of stress $\sigma_{r}$ (a) and $\sigma_{\theta}$ (b) in different biharmonic functions of the circular tube under the DCEM-P. "None": the result of DCEM. "${\rm ln}(r)$", "$r^{2}$", and "$r^{2}{\rm ln }(r)$": add respective one term in DCEM-P. "${\rm ln}(r)$, $r^{2}$, $r^{2}{\rm ln }(r)$": add all terms in DCEM-P.
		\label{fig:biharmonic result of circular}}
\end{figure}

Considering the strong form of the stress function:
\begin{equation}
	\nabla^{4}\phi=0. \label{eq:the biharmonic equation}
\end{equation}
Adding some basis functions that naturally satisfy the biharmonic equation as shown in \Cref{eq:the biharmonic equation} will improve the accuracy and convergence speed of DCEM. Thus, DCEM-P is proposed by adding some basis functions to the DCEM model. Take the circular tube as an example, and the biharmonic equation for the axisymmetric problem is simplified as follows:
\begin{equation}
	\frac{d^{4}\phi}{dr^{4}}+\frac{2}{r}\frac{d^{3}\phi}{dr^{3}}-\frac{1}{r^{2}}\frac{d^{2}\phi}{dr^{2}}+\frac{1}{r^{3}}\frac{d\phi}{dr}=0.
\end{equation}
The biharmonic function satisfying this equation is ${\rm ln}(r)$, $r^{2}$ and $r^{2}{\rm ln}(r)$. The stresses of these biharmonic functions are shown in \Cref{tab:biharmonic stress}, and we add the biharmonic function to the stress function field:
\begin{equation}
	\phi=NN(r;\theta)+a_{1}{\rm ln}(r)+a_{2}r^{2}+a_{3}r^{2}{\rm ln}(r),
\end{equation}
\begin{table}
	\caption{The stress fields (polar coordinates) are obtained by the Airy stress function satisfying the biharmonic function.\label{tab:biharmonic stress}}
	
	\centering{}%
	\begin{tabular}{cccc}
		\toprule 
		Biharmonic function & ${\rm ln}(r)$ & $r^{2}$ & $r^{2}{\rm ln}(r)$\tabularnewline
		\midrule
		$\sigma_{r}$ & $1/r^{2}$ & 2 & $1+2{\rm ln}(r)$\tabularnewline
		$\sigma_{\theta}$ & $-1/r^{2}$ & 2 & $3+2{\rm ln}(r)$\tabularnewline
		$\tau_{r\theta}$ & 0 & 0 & 0\tabularnewline
		\bottomrule
	\end{tabular}
\end{table}
where $a_{1}$, $a_{2}$ and $a_{3}$ are set as optimization variables for optimization. \Cref{fig:biharmonic result of circular} compares the stress $\mathcal{L}_{2}^{rel}$ error with the biharmonic function, i.e., adding ${\rm ln}(r)$, $r^{2}$, $r^{2}{\rm ln }(r)$ separately and all biharmonic functions. After adding the biharmonic function, the stress accuracy has been improved to a certain extent, and the convergence speed has been accelerated at the same time. The improved effect of the biharmonic function ${\rm ln}(r) $ is the most obvious because the analytical solution contains  $1/r^{2}$. Thus it is equivalent to naturally constructing the stress function that satisfies the analytical solution, and the accuracy and convergence speed are naturally fast. In addition, all biharmonic functions (${\rm ln}(r)$, $r^{2}$, $r^{2}{\rm ln }(r)$) are added, and the accuracy and convergence speed is almost the same as just adding ${\rm ln}(r)$, which shows that DCEM-P can well adjust the best biharmonic terms adaptively, making it prominent ${\rm ln}(r )$. After adding the biharmonic functions, there will be a sudden change shown in \Cref{fig:biharmonic result of circular} in a certain iteration step, which is caused by a dynamic game between the neural network approximation and the trainable coefficient of the biharmonic function. The gradient of the coefficient of the biharmonic function is larger than that of trainable parameters of the neural network. As a result, we adjust the learning rate of different parameters according to the gradient. To be specific, 
\begin{equation}
	lr_{bh}=lr_{nn}*\frac{mean\{|\nabla_{nn}Loss|\}}{mean\{|\nabla_{bh}Loss|\}},
\end{equation}
where  $lr_{bh}$ is the learning rate of the coefficient of the biharmonic function, and $lr_{nn}$ is the learning rate of the trainable parameters of the neural network. $\nabla_{nn}Loss$ means the gradient of the trainable parameters including weight but without bias of neural network. $\nabla_{bh}Loss$ means the gradient of the coefficient of the biharmonic function.

Next, we compare the accuracy and efficiency of the DEM, the PINNs displacement strong form (LN equation), and the DCEM (without the biharmonic function) under full displacement boundary conditions. The strong form of the PINNs stress function is not compared. This is because the strong form of the stress function is inconvenient for handling displacement boundary conditions, as it needs to be given in the form of boundary integrals.
 Thus, the strong form of the stress function is not considered. The construction method of the admissible displacement field in DEM and the strong form of PINNs displacement is:

\begin{equation}
	u=(1-\frac{r}{a})\frac{a}{a-b}u_{r}|_{r=b}+(1-\frac{r}{b})\frac{b}{b-a}u_{r}|_{r=a}+(r-a)(r-b)*NN(r;\boldsymbol{\theta}).\label{eq:=005747=00538B=005706=007B52=0053EF=0080FD=004F4D=0079FB=0089E3=005F62=005F0F}
\end{equation}

\begin{figure}
	\begin{centering}
		\includegraphics[scale=0.95]{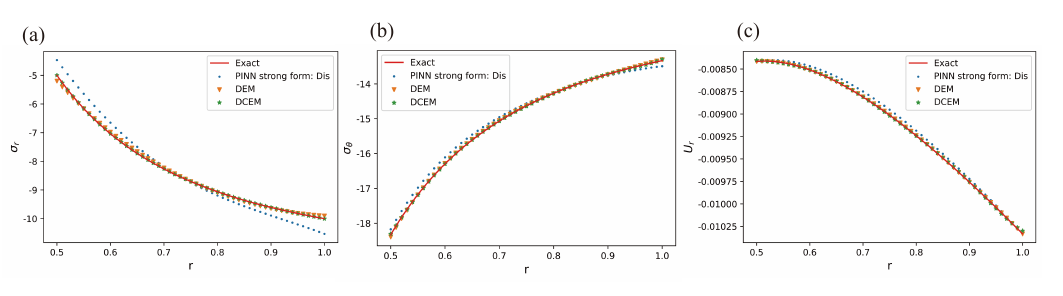}
		\par\end{centering}
	
	\centering{}\caption{Comparison of the predicted stresses $\sigma_{r}$ (a), $\sigma_{\theta}$ (b) and $u_{r}$ (c) with the analytical solution of the circular tube under the strong form of PINNs displacement, DEM,  and DCEM  in the radial direction  \label{fig:stress_different method_line_circular tube}}
\end{figure}

\begin{figure}
	\begin{centering}
		
		\includegraphics[scale=1.00]{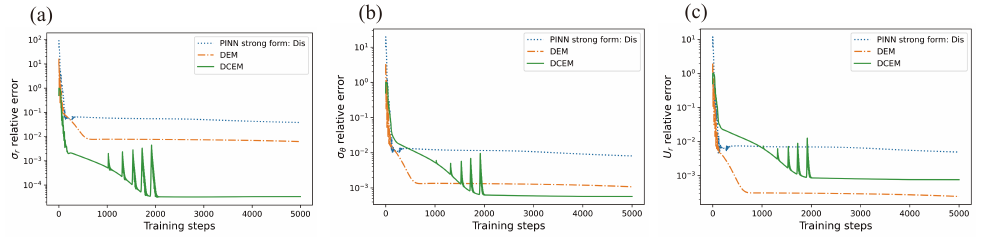}
		\par\end{centering}
	
	\centering{}\caption{Relative error $\mathcal{L}_{2}^{rel}$ of the stresses $\sigma_{r}$ (a), $\sigma_{\theta}$ (b) and $u_{r}$ (c) of the circular tube under the strong form of PINNs displacement, DEM and DCEM  \label{fig:stress_different method_line_circular tube_error}}
\end{figure}

It is not difficult to verify that the above admissible displacement field satisfies the given displacement boundary condition. The configurations of these three methods are the same (10,000 points in the radial direction, optimizer Adam, 3 hidden layers of neural network, 20 neurons in each layer). Since the strong form of PINNs displacement involves the second-order derivative, the computation time is larger than that of the DEM with only the first-order derivative. In addition, the DCEM   and DEM also involve numerical integrals. As a result, the calculation of DCEM is required for precise numerical integration.  
In terms of the stress, \Cref{fig:stress_different method_line_circular tube}  and \Cref{fig:stress_different method_line_circular tube_error} show that DCEM is the closest to the analytical solution, and the precision is the highest, while DEM has the most accurate in terms of displacement. The precision of the DCEM is the highest because there is no assumption on the form of the stress function, and DCEM completely releases the strong ability of the neural network to fit. However, the DEM and the PINNs displacement strong form are optimized under a certain displacement assumption space, and the optimization space is not as large as the DCEM. Thus, the approximation error of DEM is greater than DCEM in terms of stress. Because DCEM is based on stress directly and DEM is based on displacement directly, the accuracy of displacement obtained by DEM is higher than DCEM. The phenomenon is similar to FEM based on displacement or stress. In sum, for the problem of dealing with full-displacement boundary conditions, the DCEM is theoretically more suitable than DEM and PINNs for stress estimation.

\begin{figure}
	
	\begin{centering}
		\includegraphics[scale=0.8]{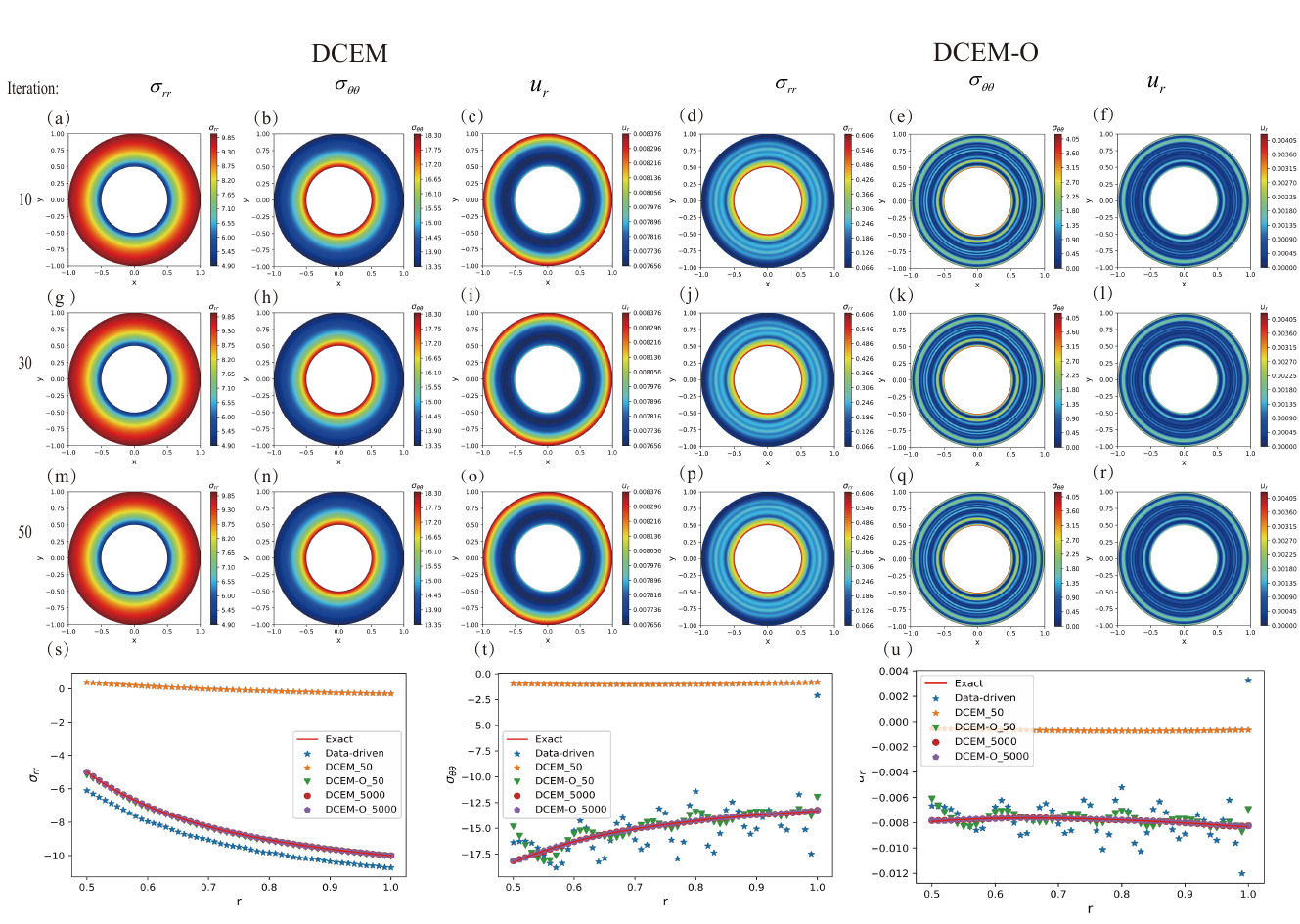}
		\par\end{centering}
	\caption{Comparison of DCEM and DCEM-O in Airy stress function of circular tube: (a, b, c, d, e, f), (g, h, i, j, k, l), and (m, n, o, p, q, r) the absolution error in the number of iterations 10, 30, and 50 respectively; (a, g, m, d, j, p) $\sigma_{rr}$; (b, h, n, e, k, q) $\sigma_{\theta \theta}$; (c, i, o, f, l, r) $u_{r}$; (a, b, c, g, h, i, m, n, o) the result of DCEM; (d, e, f, j ,k ,l, p, q, r) the result of DCEM-O; (s) the prediction of  $\sigma_{rr}$ by DCEM and DCEM-O of initial 50 and converged 5000 iterations; (t) the prediction of  $\sigma_{\theta \theta}$ by DCEM and DCEM-O of initial 50 and converged 5000 iterations; (u) the prediction of  $u_{r}$ by DCEM and DCEM-O of initial 50 and converged 5000 iterations. Data-driven means DCEM-O predicts the stress without fine-tuning by the physical laws.  \label{fig: the DCEM and DCEM-P in circular}}
	
\end{figure}

We compare the DCEM-O (operator learning based on DCEM) with DCEM in this full-displacement boundary example. The input of the branch net is pressure, $p_{i}$ and $p_{o}$. Because the problem does not have the force boundary condition, the basis and particular solution are not required. The output of the DCEM-O can be written as:
\begin{equation}
	\phi=NN_{B}(p_{i}, p_{o};\boldsymbol{\theta})*NN_{T}(r;\boldsymbol{\theta}).
\end{equation}
The architecture of the trunk net is 6 hidden layers and 30 neurons in every hidden layer. The architecture of the branch net is 3 hidden layers and 30 neurons in every hidden layer. We use 10000 different $p_{i}$, $p_{o}$ (Boundary conditions), Poisson's ratio (Material property), and b (Geometry) as the input of the branch net. The input of trunk net is all 10 equal spacing points: $p_{i}$ from 4.0 to 6.0, $p_{o}$ from 9.0 to 11.0, Poisson's ratio from 0.3 to 0.49 and b from 0.9 to 1.1, respectively. Every input of $p_{i}$, $p_{o}$, Poisson's ratio, and b is divided into all 10 equally spaced points, resulting in a total of 10*10*10*10 = 10,000 data points. Every data point costs 2.46s for training DCEM-O. The output of DCEM-O is the Airy stress function $\phi$, and the output training dataset is from the analytical solution of the problem or the high fidelity numerical experiment (FEM based on the stress function  \cite{cen20118}). The analytical solution of the circular tube can be written as:
\begin{equation}
	\phi=\frac{a^{2}}{b^{2}-a^{2}}[\frac{r^{2}}{2}-b^{2}ln(r)]p_{i}-\frac{b^{2}}{b^{2}-a^{2}}[\frac{r^{2}}{2}-a^{2}ln(r)]p_{o}.
\end{equation}
Note that the test dataset (we test $p_{i}=5$, $p_{o}=10$, Poisson's ratio 0.5, and $b=1.0$) is out of the training set. The test is interesting because the Poisson's ratio is incompressible and the extrapolation capability of DCEM-O can be tested. \Cref{fig: the DCEM and DCEM-P in circular} shows the comparison of DCEM with DCEM-O, and the absolute error of DCEM-O is lower than DCEM in the same iteration. \Cref{fig: the DCEM and DCEM-P in circular}s,t,u shows the DCEM-O can converge faster than the DCEM in the initial iteration step, and both can converge to the exact solution at the 5000 iteration. The time of DCEM-O and DCEM in one epoch are 0.011s and 0.0058s respectively due to different architecture of neural networks.

\begin{figure}
	
	\begin{centering}
		\includegraphics[scale=0.8]{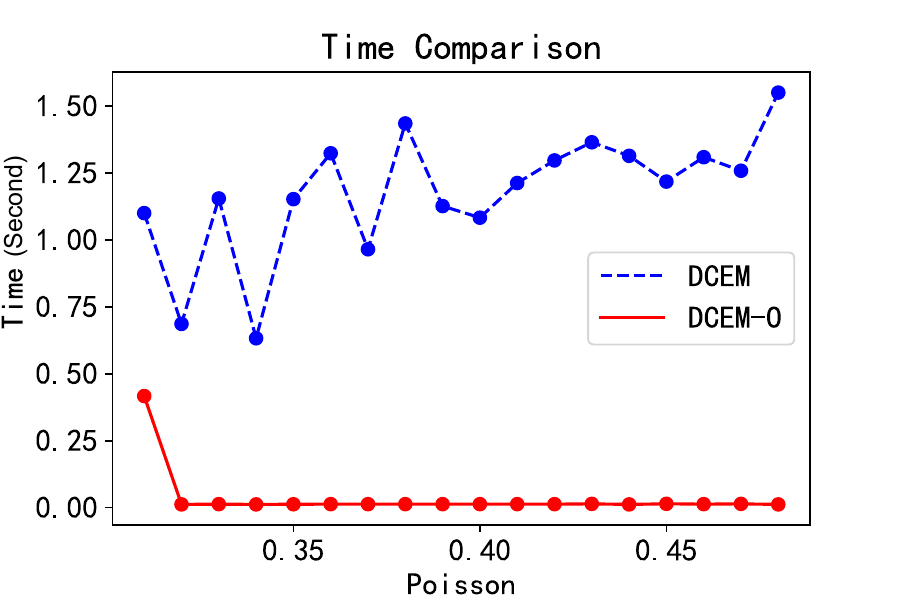}
		\par\end{centering}
	\caption{Comparison of DCEM and DCEM-O at different Poisson's ratios: the time is recorded when the $\mathcal{L}_{2}^{rel}$ is  1 \% \label{fig:DCEM_o_PARA}}
	
\end{figure}

Since the above-mentioned DCEM-O only demonstrates a specific result, we conducted statistical experiments on DCEM-O. We tested DCEM-O for Poisson’s ratio at 18 equally spaced points in the range [0.31, 0.32, ..., 0.46, 0.47, 0.48]. Please note that the Poisson’s ratios tested are outside the training set, as our training set contains Poisson’s ratios at 10 equally spaced points in the range from 0.3 to 0.49 (inclusive of endpoints 0.3 and 0.49). We conducted tests using both DCEM and DCEM-O algorithms, with the iteration stopping when the $\mathcal{L}_{2}^{rel}$ reached 1 \% and recorded the computation time. \Cref{fig:DCEM_o_PARA} shows that DCEM-O is significantly faster than DCEM. 
At Poisson’s ratio = 0.31, we found that the time required by DCME-O is greater than that for other Poisson's ratios. Because our training set's Poisson's ratios range from 0.3 to 0.49, Poisson’s ratio = 0.31 approaches the extrapolation level. However, at Poisson's ratio = 0.48, which is also close to the extrapolation level, the time required is less. This discrepancy could be due to the highly non-convex nature of neural networks, and therefore some conclusions are not necessarily deterministic.

\subsubsection{Wedge: infinite problem} \label{sec: Wedge}

\begin{figure}
	\begin{centering}
		\includegraphics{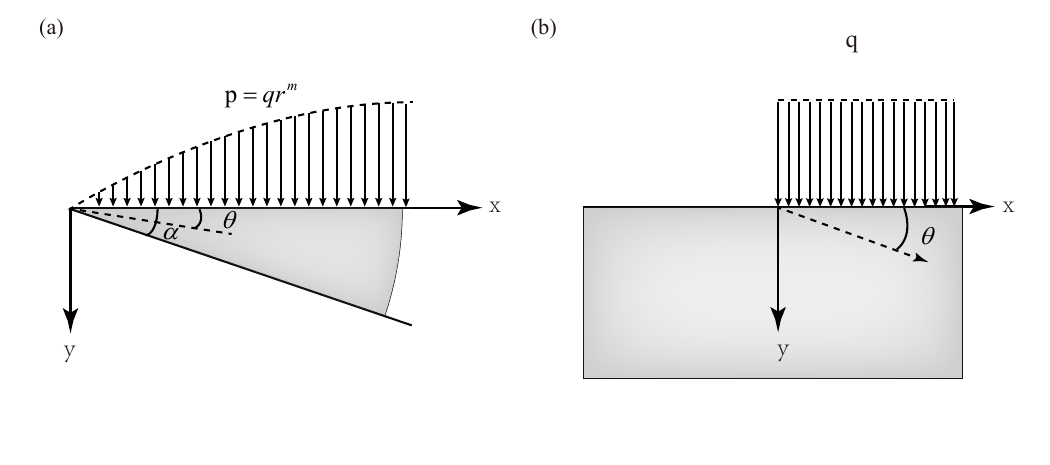}
		\par\end{centering}
	\caption{(a) The schematic of the problem of the wedge. (b) Semi-infinite uniform pressure on the half-space infinite elastic medium.  \label{fig:wedge}}
\end{figure}

The wedge problem is a common problem in engineering, and it can be encountered in various scenarios such as the curling up of the sharp leading edge of a supersonic wing \cite{fung1953behavior}. In elasticity mechanics, the stress function method is commonly used to solve this problem \cite{the_foundation_of_solid_mechanics_feng}. In this problem, a normal pressure distribution $p=qr^{m}$ acts on the wedge, as depicted in \Cref{fig:wedge}a.

The Airy stress function for this problem can be expressed analytically as follows:
\begin{equation}
	\phi=r^{m+2}\{a*cos[(m+2)\theta]+b*sin[(m+2)\theta]+c*cos(m\theta)+d*sin(m\theta)\},
\end{equation}
where the constants $a$, $b$, $c$, and $d$ are determined by the shape and the load of the boundary. For the specific case of $m=0$ and $\alpha=\pi$, which corresponds to a semi-infinite uniform pressure on the upper boundary of an elastic medium of infinite extent, the Airy stress function simplifies to:
\begin{equation}
	\phi=cr^{2}[\alpha-\theta+{\rm sin}(\theta){\rm cos}(\theta)-{\rm cos}^{2}(\theta){\rm tan}(\alpha)], \label{eq: Airy analytical stress function in wedge}
\end{equation}
where $c=q/[2({\rm tan}(\alpha)-\alpha)]$, $\alpha=\pi$, $q=5$, $E=1000$, and $\mu=0.3$, as shown in \Cref{fig:wedge}b.

Using the stress formulas based on the polar coordinates, we can obtain the analytical solutions for stress components:
\begin{equation}
	\begin{aligned}\sigma_{r} & = \frac{1}{r^{2}} \frac{\partial^{2} \phi}{\partial\theta^{2}}+\frac{1}{r}\frac{\partial\phi}{\partial r}=2c[\alpha-\theta- \text{sin}^{2}(\theta){\rm tan}(\alpha)-{\rm sin}(\theta){\rm cos}(\theta)]\\
		\sigma_{\theta} & = \frac{\partial^{2}\phi}{\partial r^{2}} =2c[\alpha-\theta+{\rm sin}(\theta){\rm cos}(\theta)-{\rm cos}^{2}(\theta){\rm tan}(\alpha)]\\
		\tau_{r\theta} & =-\frac{\partial}{\partial r}(\frac{1}{r}\frac{\partial\phi}{\partial\theta}) =c[1-{\rm cos}(2\theta)-{\rm sin}(2\theta){\rm tan}(\alpha)].
	\end{aligned} \label{eq: the relationship between stress and stress function}
\end{equation}
To verify the accuracy of the DCEM algorithm, we use the analytical solution and compare it with the PINNs strong form of the stress function as shown in \Cref{eq:the biharmonic equation}. Due to the infinite characteristic of the problem, it is challenging to solve it using the displacement method, so we do not attempt to solve it using the DEM.

The key to DCEM lies in constructing the admissible stress function that satisfies both the boundary value condition and the derivative condition, which is different from the admissible displacement field. According to the properties of the Airy stress function as explained in \Cref{sec:Appendix-B.-proof of the particular solution network}, the form of the stress function is:
\begin{equation}
	\phi=r^{2}f(\theta) \label{eq:Airy stress function in Widge assumption}
\end{equation}
We can use the neural network to approximate $f(\theta)$. Since DCEM must satisfy the force boundary condition in advance before we use the principle of minimum complementary energy,  the basis function method in \Cref{tab:The-different-methods} is used to construct the admissible stress function to satisfy the requirement of the DCEM. The admissible stress function can be expressed as: 

\begin{equation}
	f_{NN}(\theta; \boldsymbol{\theta}_{nn}) = NN_{p}(\theta; \boldsymbol{\theta}_{p}) + NN_{b}(\theta; \boldsymbol{\theta}_{b}) * NN_{g}(\theta; \boldsymbol{\theta}_{g}),
\end{equation}
where $\boldsymbol{\theta}_{nn}$ is the trained parameter of neural networks including the particular, basic, and general neural network. $\boldsymbol{\theta}_{p}$, $\boldsymbol{\theta}_{b}$, and $\boldsymbol{\theta}_{g}$ are trained parameters of the particular, basic, and general neural network respectively. Particular network $NN_{p}$ satisfies the boundary condition when on the natural boundary (cubic polynomial is used in $NN_{p}$), i.e.,
\begin{equation}
	\begin{cases}
		NN_{p}(0)=-\frac{1}{2}q\\
		NN_{p}(\alpha)=0\\
		NN_{p}^{'}(0)=0\\
		NN_{p}'(\alpha)=0,
	\end{cases}\label{eq:=007F5A=0051FD=006570=0053EF=0080FD=005E94=00529B=0051FD=006570=007684=0065BD=0052A0}
\end{equation}
where $NN_{p}^{'}$ is the derivative with respect to the angle  $\theta$.

The basis network $NN_{b}$ is constructed based on the distance function, and $NN_{b}$ is expressed as follows:
\begin{equation}
	NN_{b}(\theta)= \lambda * [\theta * (\alpha - \theta)]^2,
\end{equation}
where $\lambda=\alpha^4/16$ is used for normalization because the maximum of $NN_{b}$ without $\lambda$ is $16/\alpha^4$ when $NN_{b}(\alpha/2)$. $NN_{g}$ is a generalized network, the trainable parameters of which are determined by the principle of the minimum complementary energy. It is easy to verify that the $f_{NN}(\theta)$ satisfy the requirement of the admissible stress function, i.e. $f_{NN}(0)= -1/(2q), f_{NN}(\alpha)=0, f_{NN}^{'}(0)=0, f_{NN}'(\alpha)=0$.

\begin{figure}
	\begin{centering}
		\includegraphics[scale=1.5]{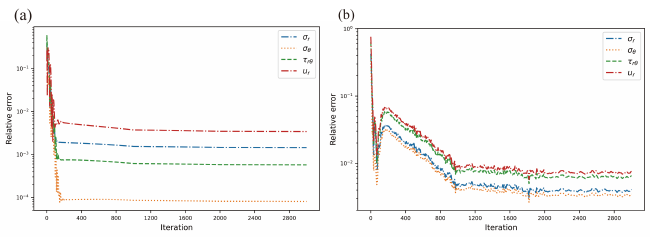}
		\par\end{centering}
	
	\centering{}\caption{The overall relative errors $\mathcal{L}_{2}^{rel}$ of the wedge in $\sigma_{r}$,$\sigma_{\theta}$, and $\tau_{r\theta}$ under DCEM (a) and PINNs strong form (b). \label{fig:DEM and DCEM in wedge L2 stress}}
	
\end{figure}

\begin{figure}
	\begin{centering}
		\includegraphics[scale=0.8 ]{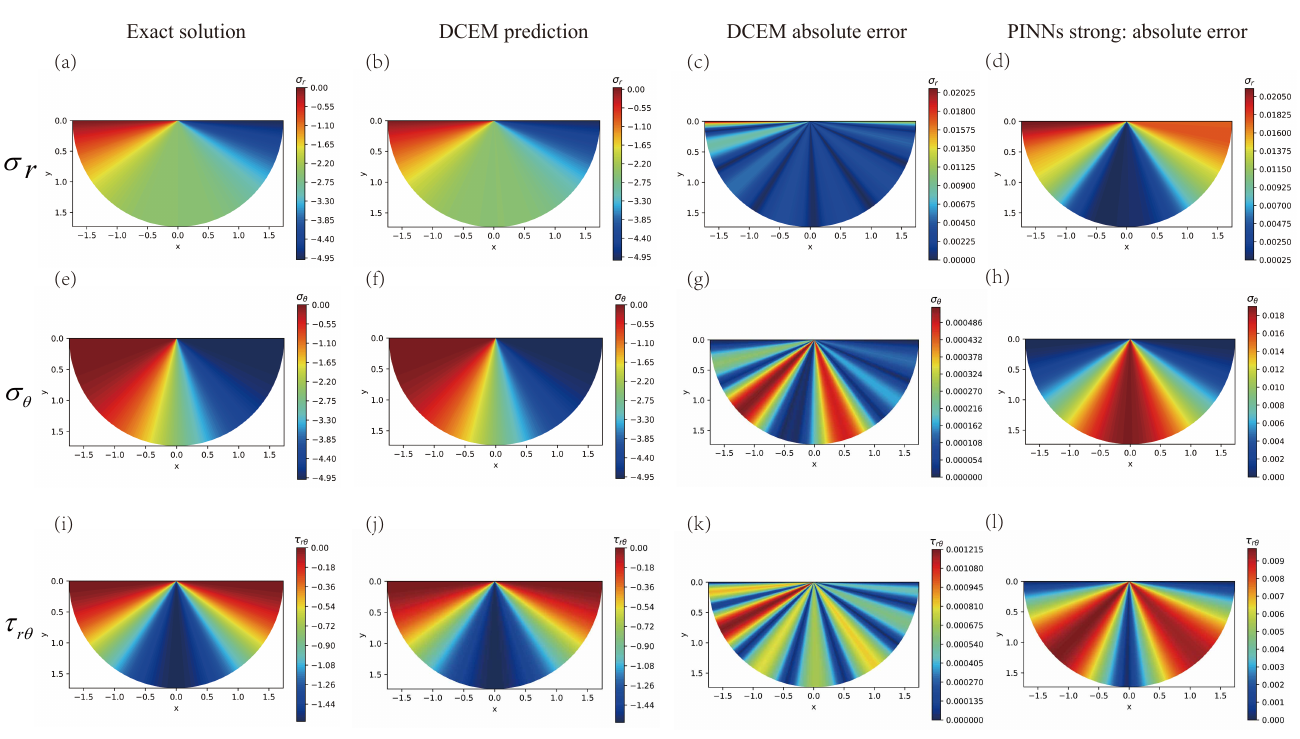}
		\par\end{centering}
	\centering{}\caption{Comparison of 	DCEM and PINNs strong form of stress function: (a, b, c, d) $\sigma_{r}$;  (e, f, g, h) $\sigma_{\theta}$;  (i, j, k, l) $\tau_{r\theta}$; (a, e, i) exact solution; (b, f, j) DCEM prediction; (c, g, k) the absolute error of DCEM; (d, h, l) the absolution error of PINNs strong form of stress function.
		\label{fig:DCEM_prediction_wedge}}
\end{figure}

\begin{figure}
	\begin{centering}
		\includegraphics[scale=1.0]{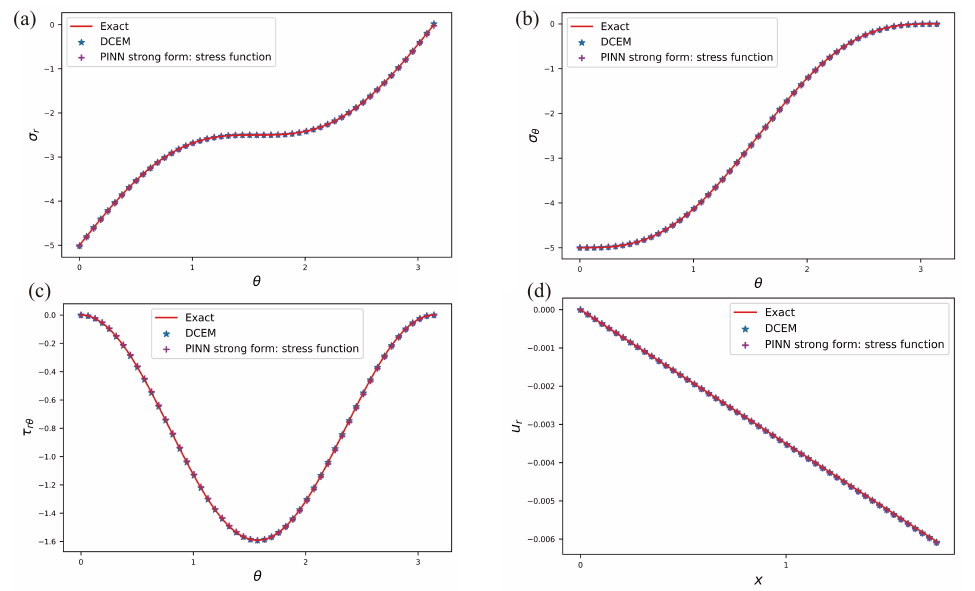}
		\par\end{centering}
	
	\centering{}\caption{Comparison of the stresses $\sigma_{r}$ (a), $\sigma_{\theta}$ (b),  $\sigma_{r\theta}$ (c) and $u_{r}$ (d) of the DCEM and PINNs stress functions of semi-infinite body subjected to a semi-infinite uniform load. \label{fig:Comparison of different stress of different models with analytical solution}}
\end{figure}

To better quantify the overall error of the DCEM, we analyze the variation trend of its $\mathcal{L}^{rel}_{2}$ error with the number of iterations. 
As shown in \Cref{fig:DEM and DCEM in wedge L2 stress}a, the relative errors of $\sigma_{r}$, $\sigma_{\theta}$, $\tau_{r\theta}$, and $u_{r}$ in DCEM are approximately $\sim10^{-3}$, $\sim10^{-5}$, $\sim10^{-4}$, and $\sim10^{-5}$, respectively.
Compared with PINNs strong form of stress function as shown in \Cref{fig:DEM and DCEM in wedge L2 stress}b, DCEM  exhibits better accuracy and efficiency, especially in efficiency, as demonstrated in \Cref{tab:Accuracy and efficiency comparison of DCEM}. This improved efficiency is mainly attributed to DCEM's lower-order derivatives, which result in higher computational efficiency. Additionally, the convergence speed of DCEM is significantly faster than that of PINNs strong form of stress function.
The accuracy of $\sigma_{\theta}$ is observed to be the highest among the other stress components. This is because of the relationship between the Airy stress function and the stress components, as shown in \Cref{eq: the relationship between stress and stress function}. The accuracy is inversely proportional to the order of the derivative.
\Cref{fig:DCEM_prediction_wedge} further illustrates the performance of DCEM and PINNs strong form. The contour plots demonstrate that DCEM exhibits higher accuracy than PINNs in terms of the maximum absolute error and the distribution of the error, although both methods are actually quite accurate as shown in \Cref{fig:Comparison of different stress of different models with analytical solution}.

For further comparison, we analyze different angles $\alpha$ as shown in \Cref{tab: different angle wedge}. The results demonstrate that DCEM consistently exhibits better accuracy and efficiency compared to PINNs strong form of stress function across different angles of the wedge. This indicates that DCEM is a robust and reliable method for solving wedge problems with varying geometries. The improved accuracy and efficiency of DCEM make it a promising approach for practical engineering applications where accurate stress analysis is crucial.

\begin{table}
	\caption{Accuracy and efficiency comparison of DCEM, DEM and strong form of stress function for wedge with uniform load\label{tab:Accuracy and efficiency comparison of DCEM}}
	\centering{}%
	\begin{tabular}{ccccccccc}
		\toprule 
		\multirow{2}{*}{\makecell[c]{Relative error \\ and converge time}} & \multicolumn{4}{c}{DCEM} & \multicolumn{4}{c}{PINNs}\tabularnewline
		\cmidrule(lr){2-5} \cmidrule(lr){6-9}  
		& $\alpha=\frac{\pi}{3}$ & $\alpha=\frac{\pi}{4}$ & $\alpha=\frac{\pi}{5}$ & $\alpha=\pi$ & $\alpha=\frac{\pi}{3}$ & $\alpha=\frac{\pi}{4}$ & $\alpha=\frac{\pi}{5}$ & $\alpha=\pi$ \tabularnewline
		\midrule 
		$\sigma_{r}$ & 6.85e-04  & 7.35e-04  &  8.23e-04 & 1.43e-03 & 2.13e-03  & 8.71e-03  & 4.84e-03 & 4.38e-03\tabularnewline
		$\sigma_{\theta}$ & 1.92e-05  & 3.01e-05 & 3.72e-05  & 8.04e-05 &  2.98e-03 & 4.41e-03  & 4.67e-03 & 3.77e-03 \tabularnewline
		$\tau_{r\theta}$ &1.19e-04  & 1.59e-04 & 1.92e-04 & 5.70e-04 & 3.41e-03  & 2.77e-03 & 1.62e-03 & 7.03e-03 \tabularnewline
		$u_{r}$ & 2.32e-03 &  2.04e-03  & 1.88e-3  & 3.40e-03 & 7.89e-03 & 7.07e-03 & 6.17e-3 & 8.18e-03 \tabularnewline
		Time(s) & 3.49  & 3.61  & 3.41  & 3.35  & 28.87  & 30.27  & 36.11 & 40.57 \tabularnewline
		\bottomrule \label{tab: different angle wedge}
	\end{tabular}
\end{table}

\begin{figure}
	
	\begin{centering}
		\includegraphics[scale=1.1]{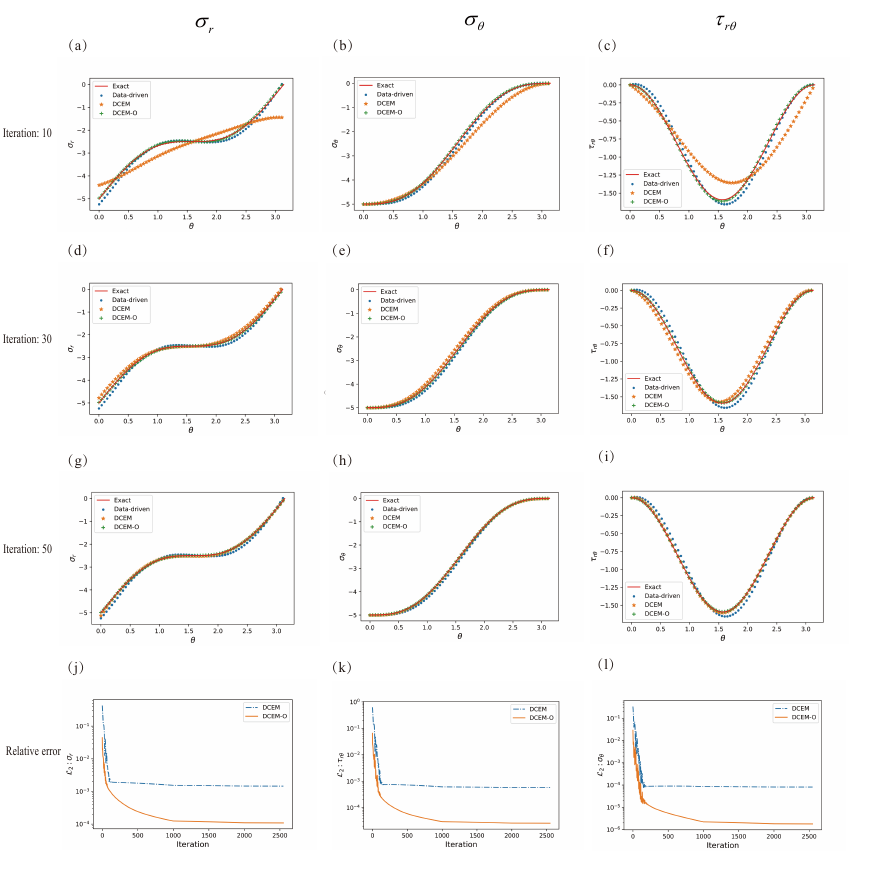}
		\par\end{centering}
	\caption{Comparison of DCEM and DCEM-O in Airy stress function of the wedge: (a, b, c), (d, e, f), and (g, h, i) the prediction of $\sigma_{r}$,  $\sigma_{\theta}$, and  $\tau_{r \theta}$ in the number of iterations 100, 500, and 1000 respectively; Data-driven means pure data-driven by DeepONet.  The evolution of the overall relative errors $\mathcal{L}_{2}^{rel}$ by DCEM and DCEM-O:  $\sigma_{r}$ (a, d, g, j);  $\sigma_{\theta}$ (b, e, h, k);  $\tau_{r \theta}$ (c, f, i, l). (j, k, l) is evolution of the relative error $\mathcal{L}^{rel}_{2}$ of  $\sigma_{r}$,  $\sigma_{\theta}$, and  $\tau_{r \theta}$ respectively with the number of iterations.  \label{fig: Comparison of DCEM and DCEM-O in Airy stress function of wedge}}
	
\end{figure}

We compared DCEM-O (operator learning based on DCEM) with DCEM in solving the wedge problem. In DCEM-O, the branch net takes the geometry information $\alpha$ and external force $q$ as input.  The output of the DCEM-O can be written as:
\begin{equation}
	\phi = r^{2} * [NN_{p}(\theta) + NN_{b}(\theta) * NN_{B}(\alpha, q;\boldsymbol{\theta})*NN_{T}(\theta ;\boldsymbol{\theta})].
\end{equation}
The architecture of the trunk net $NN_{T}(\theta ;\boldsymbol{\theta})$ consists of 6 hidden layers, each containing 30 neurons, while the branch net $NN_{B}(\alpha, q;\boldsymbol{\theta})$ contains 3 hidden layers with 30 neurons in each layer. We use the 110 different $\alpha$ and external force $q$  as the input of the branch net. The input of the trunk net is 100 equal spacing points. The output of DCEM-O is the Airy stress function $\phi$, and the training dataset for DCEM-O is obtained from the analytical solution in \Cref{eq: Airy analytical stress function in wedge}, while the test dataset (with $\alpha=\pi$ and $q=5$) is different from the training set.
\Cref{fig: Comparison of DCEM and DCEM-O in Airy stress function of wedge} illustrates the comparison of DCEM with DCEM-O. DCEM-O demonstrates significantly higher accuracy in a very small number of iterations, indicating that it converges much faster than DCEM in terms of stress results at the same iteration level. This suggests that the combination of data and physical law is a very efficient and promising approach to solving PDEs.

\subsubsection{Non-uniform tension of plate and central hole: irregular domains} \label{sec: plate and central hole}
\begin{figure}
	
	\begin{centering}
		\includegraphics[scale=1.1]{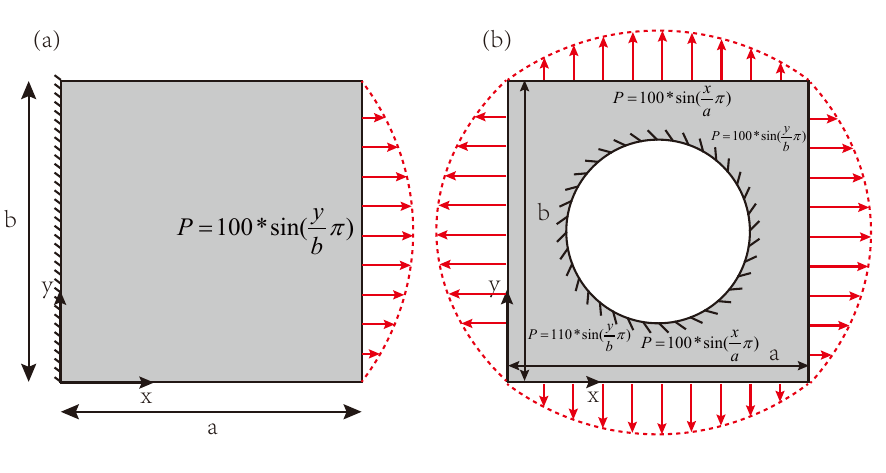}
		\par\end{centering}
	\caption{Schematic of non-uniform tensile forces on a plate with a central hole, where the material properties are set with Young's modulus $E=70$ MPa and Poisson's ratio $\nu=0.3$: (a) Non-uniform tensile forces on the plate with $a=b=1$. The applied force follows a sinusoidal distribution on the right boundary, while the left boundary is fixed. (b) Non-uniform tensile forces on the central hole with $a=b=1$. The boundary of the hole is fixed, while the other four sides are subjected to sinusoidal distributions. Specifically, the upper and lower boundaries have $P=100\sin(\pi x/a)$, the right boundary has $P=100\sin(\pi y/b)$, and the left boundary is subjected to a higher force with $P=110\sin(\pi y/b)$. \label{fig: The schematic of Non-uniform tension of plate and central hole}}
	
\end{figure}

The plate problem is a common issue in solid mechanics, often using a square plate with a central hole as a benchmark for stress concentration phenomena, as shown in \Cref{fig: The schematic of Non-uniform tension of plate and central hole}. For the square plate depicted in \Cref{fig: The schematic of Non-uniform tension of plate and central hole}a, the left boundary has fixed boundary conditions, meaning both displacement and rotation are restricted to zero.
In the
circular hole case, DCEM does not require the construction of admissible displacement fields for satisfying Dirichlet boundary conditions, so we set the boundary
of the hole in the square plate problem to be a fixed boundary. Additionally,
non-uniform stretching forces are applied along the boundaries (both
the square plate and central hole cases have a non-uniform sinusoidal
distribution of load), and the specific loading is as shown in \Cref{fig: The schematic of Non-uniform tension of plate and central hole}b. To break the symmetry of the square plate with a hole, the magnitude
of the load on the left boundary is set to 110, while the other three
sides have the same load magnitude of 100. The dimensions are chosen
as $a=b=1$mm, the radius of the circular hole is 0.25 mm, and the
center of the circle is at the center of the square plate (0.5, 0.5).
The Young's modulus is 1000 MPa, and the Poisson's ratio is 0.3.

In the first application of non-uniform tension of plate as shown in \Cref{fig: The schematic of Non-uniform tension of plate and central hole}a, since there is no analytical solution to the problem of non-uniform
stretching of square plates, we rely on the Finite Element Method
(FEM) as a reference solution.
An extremely fine mesh size of 0.01 mm (200*200: 40,000 elements in total) is utilized in Abaqus/standard. After further mesh refinement, the results of stresses do not change except for the corner of the rectangle due to the singularity.
The element type is CPS8R, an eight-node plane stress element known as the 8-node serendipity element in FEM. It is a general-purpose plane stress element, with 'R' indicating reduced integration.
In this straightforward mechanical problem, the obtained results are considered as the exact solution to evaluate the accuracy of DEM and DCEM. The boundary condition of the problem is 
\begin{equation}
	\begin{aligned}u_{x}|_{x=0}&=0,  &u_{y}|_{x=0}=0\\
		t_{x}|_{y=0}&=0,  &t_{y}|_{y=0}=0\\
		t_{x}|_{y=b}&=0,  &t_{y}|_{y=b}=0\\
		t_{x}|_{x=a}&=100sin(\frac{\pi y}{b}),  &t_{y}|_{x=a}=0.
	\end{aligned}
\end{equation}

To compare the performance
of DEM and DCEM, we focus on the stress predictions. In DEM, the admissible
displacement is constructed as
\begin{equation}
	\begin{aligned}u_{x} & =x*NN_{1}(x, y;\boldsymbol{\theta})\\
		u_{y} & =x*NN_{2}(x, y;\boldsymbol{\theta}).
	\end{aligned}
\end{equation}
The loss function of DEM is 
\begin{equation}
	\begin{aligned}\mathcal{L}_{DEM} & =U+V\\
		U & =\int_{0}^{a}\int_{0}^{b}\frac{1}{2}\varepsilon_{ij}\sigma_{ij}dxdy\\
		& =\int_{0}^{a}\int_{0}^{b}\frac{E}{2(1-\upsilon^{2})}[(\frac{\partial u_{x}}{\partial x})^{2}+2\upsilon\frac{\partial u_{x}}{\partial x}\frac{\partial u_{y}}{\partial y}+(\frac{\partial u_{y}}{\partial y})^{2}+(1-\upsilon)(\frac{\partial u_{x}}{\partial y}+\frac{\partial u_{y}}{\partial x})^{2}]dxdy\\
		V & =\int_{0}^{b}t_{i}u_{i}dy=\int_{0}^{b}t_{x}|_{x=a}u_{x}dy=\int_{0}^{b}100sin(\frac{\pi y}{b})u_{x}dy
	\end{aligned}
	\label{eq:DEM loss}
\end{equation}
The outputs are 2D displacements ($u_{x}$, $u_{y}$), because outputs of two independent networks with different parameters are more accurate than the outputs of a single network \cite{PINN_solid_mechanics}.

In DCEM, the particular neural network $NN_{p}$ is trained in advance
to satisfy the force boundary condition on the upper, right, and lower
boundaries. These conditions are mathematically expressed as follows:
\begin{equation}
	\begin{cases}
		n_{x}\frac{\partial^{2}\phi}{\partial y^{2}}-n_{y}\frac{\partial^{2}\phi}{\partial x\partial y}=t_{x}\\
		n_{y}\frac{\partial^{2}\phi}{\partial x^{2}}-n_{x}\frac{\partial^{2}\phi}{\partial x\partial y}=t_{y}
	\end{cases},\label{eq:force_boundary_condition}
\end{equation}
where $n_{x}$ and $n_{y}$ represent the components of the outward
normal vector at each boundary point. The quantities $t_{x}$ and
$t_{y}$ correspond to the prescribed forces along the x and y directions,
respectively. This formulation ensures that the force boundary conditions
are fulfilled within the DCEM framework. The particular network is
set to 4 hidden layers and 20 neurons in each layer. The dimension
of input is two, i.e., x and y coordinates, and the output is only
the Airy stress function and the learning rate is 1e-3. Through the
\Cref{eq: particular equations}, the basis function is defined as 
\begin{equation}
	F_{b}(x, y)=\frac{1}{64ab^{2}}[(x-a)(y)(y-b)]^{2}. \label{eq:basic_f}
\end{equation}
As a result, the admissible stress function is followed as 
\begin{equation}
	\phi=NN_{p}(x, y)+F_{b}(x, y)*NN_{g}(x, y;\boldsymbol{\theta})
\end{equation}
It is easy to verify that the force boundary condition in \Cref{eq:force_boundary_condition} is satisfied
due to the formula of the basis function in \Cref{eq:basic_f}. $NN_{g}$ is the general
network and is trained using the principle of complementary energy.
Its architecture and training scheme are identical to the particular
neural network $NN_{p}$. 
The loss function of DCEM is 
\begin{equation}
	\begin{aligned}\mathcal{L}_{DCEM} & =U_{c}+V_{c}\\
		U_{c} & =\int_{0}^{a}\int_{0}^{b}\frac{1}{2}\varepsilon_{ij}\sigma_{ij}dxdy\\
		& =\int_{0}^{a}\int_{0}^{b}\frac{1}{2E}[(\frac{\partial^{2}\phi}{\partial x^{2}})^{2}+(\frac{\partial^{2}\phi}{\partial y^{2}})^{2}-2\upsilon\frac{\partial^{2}\phi}{\partial x^{2}}\frac{\partial^{2}\phi}{\partial y^{2}}+2(1+\upsilon)(\frac{\partial^{2}\phi}{\partial x\partial y})^{2}]dxdy\\
		V_{c} & =\int_{0}^{b}t_{i}u_{i}dy=\int_{0}^{b}(t_{x}u_{x}|_{x=a}+t_{y}u_{y}|_{x=a})dy=0.
	\end{aligned}
	\label{eq:DCEM loss}
\end{equation}

To ensure a fair comparison between DEM
and DCEM, all the conditions including the training scheme, network architecture,
and point distribution are kept the same. The point distribution is
a uniform grid with a size of 101{*}101. 

\begin{figure}
	\begin{centering}
		\includegraphics[scale=1.15]{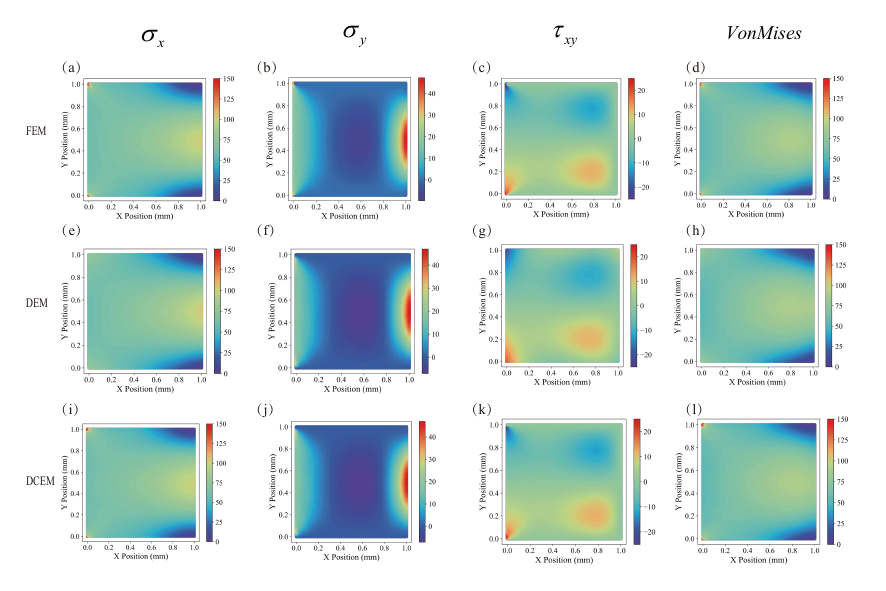}
		\par\end{centering}
	\caption{Comparison of FEM, DEM, and DCEM in terms of stress in non-uniform
		tension of plate: The result of FEM (a, b, c, d), DEM (e, f, g, h)
		and DCEM (i, j, k ,l). The result of $\sigma_{x}$ (a, e, i), $\sigma_{y}$
		(b, f, j), $\tau_{xy}$ (c, g, k), and Von-Mises stress (d, h, l).
		\label{fig: Comparison of FEM DEM and DCEM in term of stress in non-uniform tension of plate}}
\end{figure}

\begin{figure}
	\begin{centering}
		\includegraphics[scale=1.2]{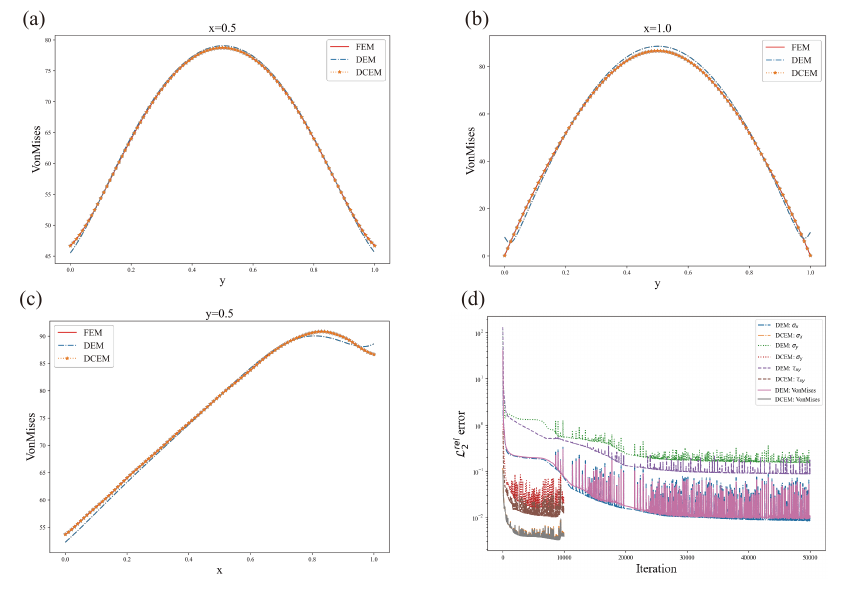}
		\par\end{centering}
	\caption{Different location comparison among FEM, DEM, and DCEM in non-uniform
		tension of plate: Von-Mises on the line of $x=0.5$ (a), right boundary
		$x=1.0$ (b), $y=0.5$ (c). The overall relative error $\mathcal{L}_{2}^{rel}$
		in $\sigma_{x}$, $\sigma_{y}$, $\tau_{xy}$, and Von-Mises (d).
		\label{fig: Different location comparison among FEM DEM and DCEM in non-uniform tension of plate}}
\end{figure}

\Cref{fig: Comparison of FEM DEM and DCEM in term of stress in non-uniform tension of plate}
illustrate the stress prediction results. Both DEM and DCEM have achieved
good results, but when it comes to capturing singular stress at the
corners, DCEM demonstrates a superior stress prediction performance.
\Cref{fig: Comparison of FEM DEM and DCEM in term of stress in non-uniform tension of plate}a,b,c
show the Von-Mises stress at different locations obtained by both
methods. DCEM is notably more accurate than DEM in capturing stress
values. \Cref{fig: Comparison of FEM DEM and DCEM in term of stress in non-uniform tension of plate}d
shows the relative overall $\mathcal{L}_{2}^{rel}$ error where DCEM outperforms
DEM in all stress components. 
\Cref{fig: Different location comparison among FEM DEM and DCEM in non-uniform tension of plate} presents a comparison of different locations among FEM, DEM, and DCEM. The results indicate that DCEM demonstrates better accuracy compared to DEM.
Besides, DCEM has a faster convergence
speed compared to DEM. For an efficiency comparison between DEM and
DCEM, \Cref{tab:Comparison of efficiency and accuracy between FEM DEM and DCEM in non-uniform tension of plate and central hole proplem} demonstrates that the calculation time of DCEM is one-tenth
of that of DEM.


\begin{table}
	\caption{Comparison of efficiency and accuracy between FEM, DEM, and DCEM in non-uniform tension of plate and central hole proplem.\label{tab:Comparison of efficiency and accuracy between FEM DEM and DCEM in non-uniform tension of plate and central hole proplem}}
	\centering{}%
	\begin{tabular}{ccccccc}
		\toprule 
		\multirow{2}{*}{\makecell[c]{Relative error \\ and converge time}} & \multicolumn{3}{c}{Plate} & \multicolumn{3}{c}{Central hole}\tabularnewline
		\cmidrule(lr){2-4} \cmidrule(lr){5-7}  
		& Points & \makecell[c]{Relative error\\  of VonMises} & Time(s)  & Points & \makecell[c]{Relative error\\  of VonMises} & Time(s)  \tabularnewline
		\midrule 
		FEM &\makecell[c]{CPS8R:  \\ 40000(ele)}   & \makecell[c]{Reference\\   solution} & 42.09s & \makecell[c]{CPS8R:  \\ 39806(ele)} & \makecell[c]{Reference\\   solution} & 45.44s  \tabularnewline
		DEM & 101*101 & 0.01109 & 525.7s   & 101*101 & 0.04802 & 438.7s  \tabularnewline
		DCEM & 101*101 & 0.003728 & 54.4s  & 101*101 & 0.01991 & 48.1s  \tabularnewline
		\bottomrule 
	\end{tabular}
\end{table}

\begin{figure}
	
	\begin{centering}
		\includegraphics[scale=1.4]{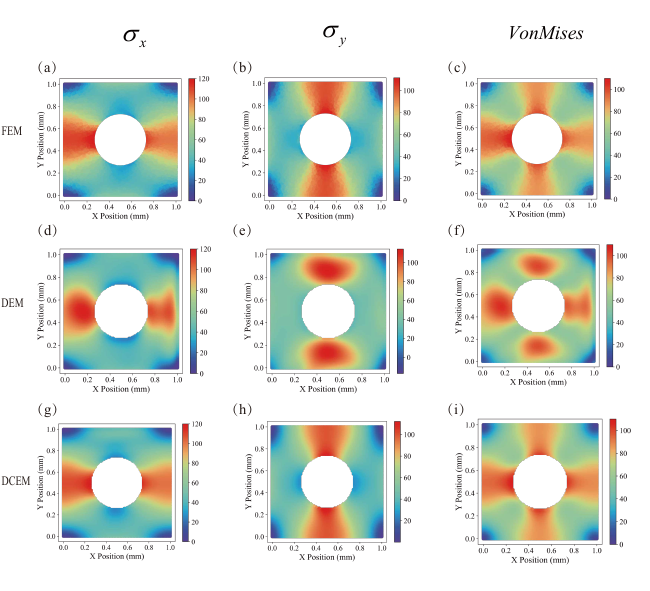}
		\par\end{centering}
	\caption{Comparison of FEM, DEM, and DCEM in terms of stresses in non-uniform tension of central hole: the reference solution $\sigma_{x}$  (a), $\tau_{xy}$  (b),  VonMises (c) by FEM. The prediction $\sigma_{x}$  (d), $\tau_{xy}$  (e),  VonMises (f) by DEM. The prediction $\sigma_{x}$  (g), $\tau_{xy}$  (h),  VonMises (i) by DCEM. \label{fig: Comparison of FEM DEM and DCEM in term of stress in non-uniform tension of central hole}}
	
\end{figure}

The second numerical example involves the non-uniform tension of a
specimen with a central hole, as depicted in \Cref{fig: The schematic of Non-uniform tension of plate and central hole}b.
This problem possesses an irregular geometry and can be challenging
to solve with DEM when using uniform point distribution and basic
Monte Carlo integration \cite{the_comparision_of_strong_and_energy_form}. The FEM solution obtained from Abaqus/standard
serves as the reference solution, utilizing 39,806 CPS8R elements.
The governing equation remains the same as the rectangular problem,
as expressed in \Cref{eq:DEM loss} and \Cref{eq:DCEM loss}. All settings
are consistent with the plate problem, except for the boundary
conditions:
\begin{equation}
	\begin{aligned}t_{x}|_{x=0}=-110sin(\frac{\pi y}{b}), & t_{y}|_{x=0}=0\\
		t_{x}|_{y=0}=0, & t_{y}|_{y=0}=-100sin(\frac{\pi x}{a})\\
		t_{x}|_{y=b}=0, & t_{y}|_{y=b}=100sin(\frac{\pi x}{a})\\
		t_{x}|_{x=a}=100sin(\frac{\pi y}{b}), & t_{y}|_{x=a}=0.
	\end{aligned}
\end{equation}

In DEM, the admissible displacement is constructed as 
\begin{equation}
	\begin{aligned}u_{x} & =[(\frac{x-0.5}{0.25})^{2}+(\frac{y-0.5}{0.25})^{2}-1]*NN_{1}(x,y;\boldsymbol{\theta})\\
		u_{y} & =[(\frac{x-0.5}{0.25})^{2}+(\frac{y-0.5}{0.25})^{2}-1]*NN_{2}(x,y;\boldsymbol{\theta}).
	\end{aligned}
\end{equation}

\begin{figure}
	
	\begin{centering}
		\includegraphics[scale=1.1]{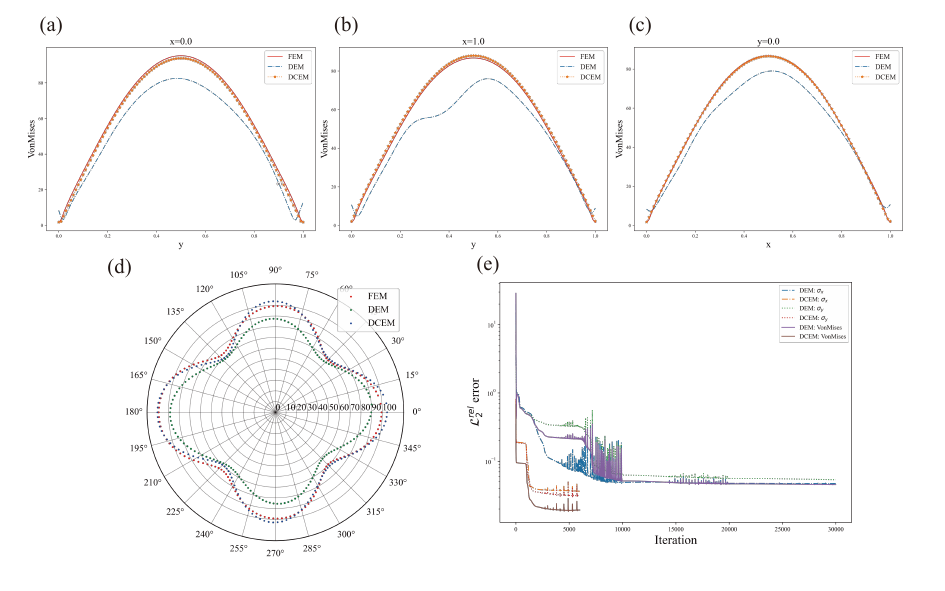}
		\par\end{centering}
	\caption{Different location comparison among FEM, DEM, and DCEM in non-uniform tension of central hole: the VonMises prediction by FEM, DEM, and DCEM on $x=0$ (a), $x=1.0$ (b), and $y=0.0$ (c),  and the circumferential stress around the hole (d). (e) The evolution of overall relative error $\mathcal{L}_{2}^{rel}$ by DEM and DCEM.   \label{fig: Different location comparison among FEM DEM and DCEM in non-uniform tension of central hole}}
	
\end{figure}
In DCEM, due to the fixed displacement boundary (Dirichlet boundary),
there's no requirement to handle the boundary of the hole, giving
DCEM an advantage in dealing with complex Dirichlet boundaries, attributed
to the principle of complementary energy. \Cref{fig: Comparison of FEM DEM and DCEM in term of stress in non-uniform tension of central hole}
illustrate the highly accurate stress results obtained by DCEM. DCEM
effectively captures stress concentration, demonstrating its efficiency
as shown in \Cref{tab:Comparison of efficiency and accuracy between FEM DEM and DCEM in non-uniform tension of plate and central hole proplem}. \Cref{fig: Different location comparison among FEM DEM and DCEM in non-uniform tension of central hole}a,b,c
show the Von-Mises stress at different locations obtained by DEM and
DCEM. \Cref{fig: Different location comparison among FEM DEM and DCEM in non-uniform tension of central hole}d
indicates DCEM's accurate results near the hole boundary, even with
uniform distribution (In FEM, the fine mesh is needed). \Cref{tab:Comparison of efficiency and accuracy between FEM DEM and DCEM in non-uniform tension of plate and central hole proplem} and
\Cref{fig: Different location comparison among FEM DEM and DCEM in non-uniform tension of central hole}e
display the faster convergence speed of DCEM compared to DEM.

\begin{figure}
	
	\begin{centering}
		\includegraphics[scale=1.4]{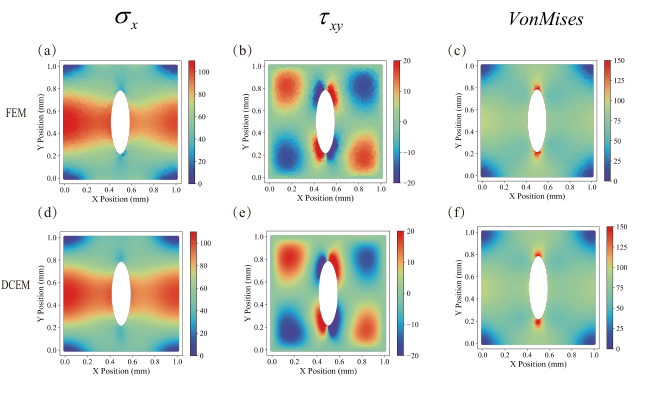}
		\par\end{centering}
	\caption{Comparison of FEM and DCEM in terms of stresses in non-uniform tension of ellipse hole: the reference solution $\sigma_{x}$  (a), $\tau_{xy}$  (b),  VonMises (c) by FEM. The prediction $\sigma_{x}$  (d), $\tau_{xy}$  (e),  VonMises (f) by DCEM.  \label{fig: Different location comparison among FEM DEM and DCEM in non-uniform tension of ellipse hole}}
	
\end{figure}

In conclusion, DCEM yields superior stress results in elasticity problems.
We also present results for other shapes, such as an ellipse, in \Cref{fig: Different location comparison among FEM DEM and DCEM in non-uniform tension of ellipse hole}.

\subsection{Prandtl stress function: full natural boundary conditions}\label{sec: prandtl}
\begin{figure}
	\begin{centering}
		\includegraphics{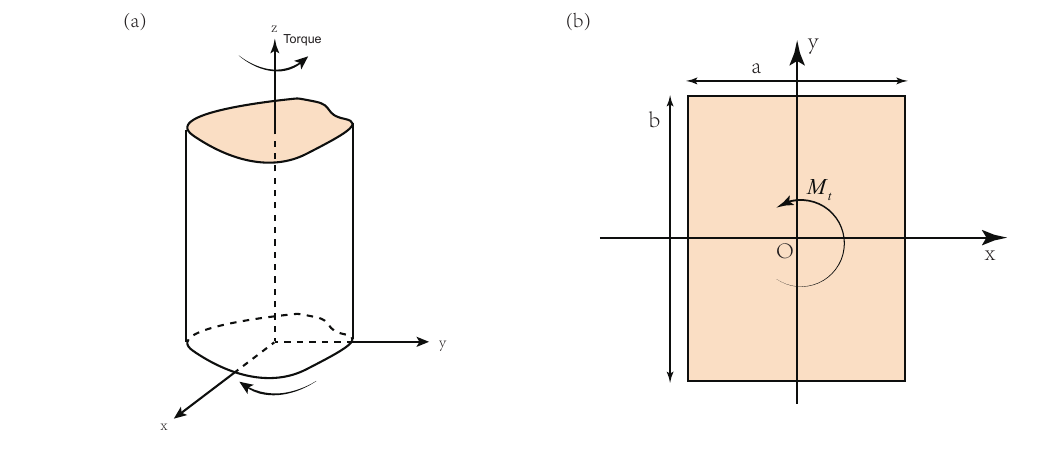}
		\par\end{centering}
	\caption{The schematic of the problem of torsion of a cylindrical body (a) and square cross-section of the cylinder (b). \label{fig:cylinder}}
\end{figure}

A cylindrical rod is a common engineering component that can withstand different loads such as tension, compression, bending, and torsion. The free torsion problem of cylindrical rods is a common problem in engineering \cite{the_foundation_of_solid_mechanics_feng}, which is often used to transmit torque, as shown in \cref{fig:cylinder}a.

Two common solutions are introduced in {\Cref{sec: introduction to stress function}, namely the displacement solution and the stress function solution. The strong form expression of the displacement solution is:
	\begin{equation}
		\begin{cases}
			\frac{\partial^{2}\psi}{\partial^{2}x}+\frac{\partial^{2}\psi}{\partial^{2}y}=0 & \boldsymbol{x}\in\Omega\\
			\text{\ensuremath{\frac{\partial\psi}{\partial x}n_{x}}+\ensuremath{\frac{\partial\psi}{\partial y}n_{y}}=y\ensuremath{n_{x}}-x\ensuremath{n_{y}}} & \boldsymbol{x}\in\partial\Omega \\
			\alpha G\oiint(x^{2}+y^{2}+x\frac{\partial\psi}{\partial y}-y\frac{\partial\psi}{x})dxdy=M_{t},
		\end{cases}
	\end{equation}
	where $\psi$ is the warping function and $\alpha$ is rate of twist. The displacement assumption can be written as:
	\begin{equation}
		\begin{split}
			u &=-\alpha zy\\
			v &=\alpha zx\\
			w  &=\alpha\psi(x,y)\label{eq:=0067F1=005F62=006746=004F4D=0079FB=0051FD=006570}.
		\end{split}
	\end{equation}
	We usually solve the Laplace equation with Neumann boundary conditions first, and get the $\psi$. Then we substitute the $\psi$  into the end boundary condition $\alpha G\iint(x^{2}+y^{2}+x \partial\psi / \partial y - y \partial\psi /\partial x)dxdy=M_{t}$ to get the unknown constant $\alpha$.
	
	The energy form of the displacement solution based on the principle of minimum potential energy (DEM) is: 
	
	\begin{equation}
		\Pi=L\int_{\Omega}(\frac{1}{2}\gamma_{zx}\tau_{zx}+\frac{1}{2}\gamma_{zy}\tau_{zy})dxdy-\alpha LM_{t},
	\end{equation}
	where $L$ is the length of the cylinder. We substitute the displacement expressed as \Cref{eq:=0067F1=005F62=006746=004F4D=0079FB=0051FD=006570} into the potential, and obtain
	\begin{equation}
		\Pi=L\int_{\Omega}\frac{1}{2}G\alpha^{2}[(\frac{\partial\psi}{\partial x}-y)^{2}+(\frac{\partial\psi}{\partial y}+x)^{2}]dxdy-\alpha LM_{t}.
	\end{equation}
	The constant term $\alpha LM_{t}$ and $G\alpha^{2}/2$ are removed, because it does not affect the optimization results of $\psi$. The above optimization problem is transformed into:
	\begin{equation}
		\psi=\argmin_{\psi}\{\int_{\Omega}[(\frac{\partial\psi}{\partial x}-y)^{2}+(\frac{\partial\psi}{\partial y}+x)^{2}]dxdy\}.
	\end{equation}
	It is worth noting that the warping function $\psi$ will have an undetermined constant in the free torsion problem of the cylindrical rod. This undetermined constant is the displacement of the rigid body and determined by the given boundary condition. If it involves strain and the derivative of the displacement function, this constant will not have an impact.
	
	In the other solution of stress, the strong form of the stress function solution is (considering simply connected regions):
	\begin{equation}
		\begin{cases}
			\frac{\partial^{2}\phi}{\partial^{2}x}+\frac{\partial^{2}\phi}{\partial^{2}y}=-2G\alpha & \boldsymbol{x}\in\Omega\\
			\phi=0 & \boldsymbol{x}\in\partial\Omega\\
			M_{t}=2\int_{\Omega}\phi d\Omega.
		\end{cases}
	\end{equation}
	
	In the stress function solution, the $\alpha$ is unknown in advance, so it cannot be solved directly. Here, the pre-assumption method is used, assuming $C_{1}=-2G\alpha$, to solve:
	
	\begin{equation}
		\begin{cases}
			\frac{\partial^{2}\phi}{\partial^{2}x}+\frac{\partial^{2}\phi}{\partial^{2}y}=C_{1}  & \boldsymbol{x}\in\Omega\\
			\phi=0  & \boldsymbol{x}\in\partial\Omega.
		\end{cases}
	\end{equation}
	
	The solution result is recorded as $\phi_{1}$, and $\phi_{1}$ is brought into the torque formula $M_{1}=2\int_{\Omega}\phi_{1}d\Omega$ to correct
	\begin{equation}
		\phi=\frac{M_{t}}{M_{1}}\phi_{1} \label{eq: revised fai}
	\end{equation}
	
	According to the uniqueness of the elastic mechanics' solution, the above revised $\phi$ in \Cref{eq: revised fai} is the true solution to the problem. By plugging the corrected $\phi$ into the domain equation, and correcting $C_{1}$
	\begin{equation}
		\frac{\partial^{2}\phi}{\partial^{2}x}+\frac{\partial^{2}\phi}{\partial^{2}y}=\frac{M_{t}}{M_{1}}(\frac{\partial^{2}\phi_{1}}{\partial^{2}x}+\frac{\partial^{2}\phi_{1}}{\partial^{2}y})=\frac{M_{t}}{M_{1}}C_{1}.
	\end{equation}
	Thus $\alpha$ can be written as:
	\begin{equation}
		\alpha=-\frac{M_{t}C_{1}}{2M_{1}G}.
	\end{equation}
	
	Next, We consider DCEM, i.e., the minimum complementary energy form of the stress function solution, which can be written as (considering simply connected regions):
	\begin{equation}
		\phi=\argmin_{\phi}\{\Pi_{c}=\frac{L}{G}\int_{\Omega}\frac{1}{2}[(\frac{\partial\phi}{\partial x})^{2}+(\frac{\partial\phi}{\partial y})^{2}]dxdy-\alpha L\int_{\Omega}2\phi dxdy\}.
	\end{equation}
	The minimum complementary energy solution also encounters the same problem as the strong form of the stress function ($\alpha$ is not known in advance). Since the first-order variation of the complementary energy is equivalent to the strong form, the same approach can be adopted, setting $\alpha=\alpha_{1}$, to solve the minimum complementary energy form, and bring the obtained $\phi_{1}$ into the torque formula $M_{1}=2\int_{\Omega}\phi_{1} d\Omega$, and we correct
	\begin{equation}
		\begin{split}
			\phi & =\frac{M_{t}}{M_{1}}\phi_{1}\\
			\alpha & =\frac{M_{t}}{M_{1}}\alpha_{1}.
		\end{split}
	\end{equation}
	
	We take a rectangular cylinder as an example in \Cref{fig:cylinder}b to show the comparison between the prediction and the analytical solution of the torsion angle $\alpha$  of the four adopted methods: PINNs strong form of displacement, DEM, PINNs strong form of stress function, and DCEM. They share identical structures of neural networks, optimization schemes, and point allocation strategies. The neural network structure consists of two neurons in the input layer, representing the x and y coordinates. It has two hidden layers with 20 neurons in each layer. The output layer has one neuron, with the activation function being \({\rm tanh}\), and no activation function in the output layer. The optimization scheme used is Adam, and the initialization method is Xavier. All internal point allocation methods of the four methods are randomly distributed, with 10,000 points. Additionally, the strong displacement form needs to allocate extra boundary points, with 1,000 points per side, also randomly distributed. \Cref{fig: Comparison of the numerical solution and analytical solution of the rotation angle}  show that, for different $a/b$ aspect ratios, the four methods converge to the exact solution by increasing with the number of iterations, which shows that the four methods can predict the rotation angle $\alpha$ of unit length. It is worth noting that the DEM does not require to assume an admissible displacement field because all boundaries are force boundary conditions; However, the DCEM based on the minimum complementary energy principle needs to assume an admissible stress function. The stress function at the boundary is assumed to be a constant zero. Due to the regular geometry of the problem, the admissible stress function of the boundary condition is satisfied by the coordinate construction method \cite{principleofminimumcomplementary_energy} in advance:
	\begin{equation}
		\phi=(x^{2}-\frac{a^{2}}{4})(y^{2}-\frac{b^{2}}{4})NN(x,y;\boldsymbol{\theta}).
	\end{equation}
	The loss function of the strong form of displacement in PINNs is
	\begin{equation}
		\begin{split}\mathcal{L}_{strongdis}&=  \lambda_{1}\frac{1}{N_{dom}}\sum_{i=1}^{N_{dom}}|\frac{\partial^{2}\psi(\boldsymbol{x}_{i})}{\partial^{2}x}+\frac{\partial^{2}\psi(\boldsymbol{x}_{i})}{\partial^{2}y}|^{2}\\
			&+\lambda_{2}\frac{1}{N_{b}}\sum_{i=1}^{N_{b}}|\frac{\partial\psi(\boldsymbol{x}_{i})}{\partial x}n_{x}+\frac{\partial\psi(\boldsymbol{x}_{i})}{\partial y}n_{y}-y_{i}n_{x}+x_{i}n_{y}|^{2},\end{split}
	\end{equation}
	where we set $\lambda_{1}=\lambda_{2}=1$. The strong form of the PINNs stress function does not use a penalty function to satisfy the boundary conditions but uses an admissible stress function consistent with the DCEM to satisfy the force boundary conditions, so there are no additional hyperparameters in the strong form of PINNs stress function and DCEM. The strong form of PINNs stress function and  DCEM both assume $\alpha=0.0005$.

	\begin{figure}
		\begin{centering}
			\includegraphics[scale=2.0]{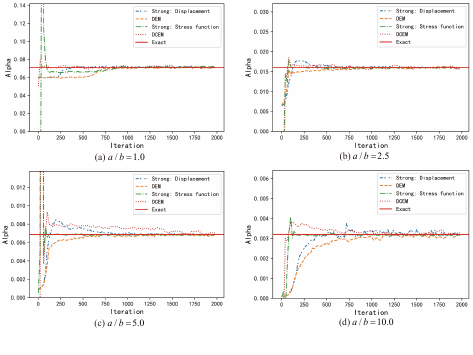}
			\par\end{centering}
		\caption{ Comparison of the numerical solution and analytical solution of the rotation angle $\alpha$ in the different rectangular cylinder (aspect ratio $a/b$) under the four models of PINNs displacement strong form, DEM based on the minimum potential energy, PINNs stress function strong form and DCEM based on the minimum complementary energy. \label{fig: Comparison of the numerical solution and analytical solution of the rotation angle}}
	\end{figure}

	In order to accurately compare the accuracy of the four methods, we analyze different aspect ratios  $a/b$ corresponding to different exact solutions $\alpha$ and maximum shear stress $\tau_{max} $, as shown in  \Cref{tab:=005728=004E0D=00540C=00957F=005BBD=006BD4=004E0B=005355=004F4D=00957F=005EA6=00626D=0089D2=007684=0089E3=006790=0089E3}. The analytical expressions of $\alpha$ and $\tau_{max}$ are:
	\begin{equation}
		\begin{split}
			\alpha =\frac{M_{t}}{\beta Gab^{3}} \\ 
			\tau_{max}  =\frac{M_{t}}{\beta_{1}ab^{2}}
		\end{split}
	\end{equation}
	
	\begin{table}
		\caption{Analytical solutions of  $\alpha$ and the maximum shear stress $\tau_{max}$ of a rectangular cylinder under different $a/b$ aspect ratios\label{tab:=005728=004E0D=00540C=00957F=005BBD=006BD4=004E0B=005355=004F4D=00957F=005EA6=00626D=0089D2=007684=0089E3=006790=0089E3}}
		
		\centering{}%
		\begin{tabular}{ccccc}
			\toprule 
			Aspect ratio $a/b$ & $\beta$ & Torsion angle $\alpha$ & $\beta_{1}$ & $\tau_{max}$\tabularnewline
			\midrule
			1.0 & 0.141 & 70.922 & 0.208 & 48076.923\tabularnewline
			1.2 & 0.166 & 50.201 & 0.219 & 38051.750\tabularnewline
			1.5 & 0.196 & 34.014 & 20231 & 28860.029\tabularnewline
			2.0 & 0.229 & 21.834 & 0.246 & 20312.203\tabularnewline
			2.5 & 0.249 & 16.064 & 0.258 & 15503.876\tabularnewline
			3.0 & 0.263 & 12.672 & 0.267 & 12484.395\tabularnewline
			4.0 & 0.281 & 8.897 & 0.282 & 8865.248\tabularnewline
			5.0 & 0.291 & 6.873 & 0.291 & 6872.852\tabularnewline
			10.0 & 0.312 & 3.205 & 0.312 & 3205.128\tabularnewline
			\bottomrule
		\end{tabular}
	\end{table}
	
	\Cref{fig:Prantl_alpha_tau_inital}a gives the relative error $|pred-exact|/exact$ of $\alpha$ under different aspect ratio $a/b$. The results show that all four methods (strong and energy form of displacement and stress function) can obtain satisfactory results under different rectangular sizes. The number of iterations of all methods is 2000, and all configurations related to training are the same. The results show that, under the same number of iterations, the accuracy of the four methods is similar. From the perspective of computational efficiency, DEM and DCEM can be computationally more efficient due to the lower order of the derivative than the corresponding strong form of displacement and stress functions theoretically.  Since all boundaries are force boundary conditions, DEM does not require additional construction of admissible displacement fields in advance, which is a great advantage compared to the DCEM. From the other extreme, if all the boundaries are displacement boundary conditions which we show in \Cref{sec: circular}, the DCEM does not need to construct the admissible stress function field in advance. Therefore, different energy principles have different advantages under different types of boundary conditions.
	\begin{figure}
		\begin{centering}
			\includegraphics[scale=1.1]{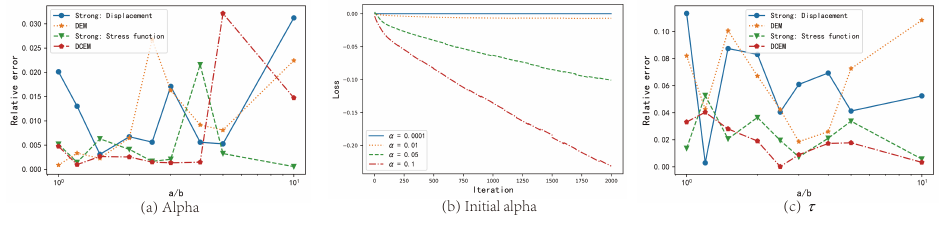}
			\par\end{centering}
		\caption{(a) Relative error of rotation angle $\alpha$ of a rectangular cylinder under four models under PINNs displacement strong form, DEM, PINNs stress function strong form, and DCEM. (b) The effect of different initial $\alpha$ on the convergence rate under DCEM. (c) Relative error of the maximum shear stress $\tau$ of the rectangular cylinder under the strong form of PINNs displacement, DEM, the strong form of PINNs stress function and DCEM.\label{fig:Prantl_alpha_tau_inital}}
	\end{figure}
	
	In order to further analyze the potential of the four methods, we cancel the limitation of the fixed number of iterations and calculate the relative error and convergence time of $\alpha$ when convergence, as shown in  \Cref{tab:=0056DB=0079CD=0065B9=006CD5=005728=004E0D=00540C=00957F=005BBD=006BD4=004E0B=005355=004F4D=00957F=005EA6=00626D=0089D2=007CBE=005EA6=00548C=006536=00655B=0065F6=0095F4=007684=005BF9=006BD4}. The computer hardware for all the numerical experiments is RTX2060, 6GB GPU, and 16GB RAM. Generally, the DCEM and the DEM have higher calculation efficiency because of the low order of derivatives, especially DCEM.
	\begin{center}
		\begin{table}
			\caption{Comparison of accuracy and convergence time of torsion angle $\alpha$ under different aspect ratios $a/b$ under the four models of PINNs displacement strong form, DEM, PINNs stress function strong form and DCEM\label{tab:=0056DB=0079CD=0065B9=006CD5=005728=004E0D=00540C=00957F=005BBD=006BD4=004E0B=005355=004F4D=00957F=005EA6=00626D=0089D2=007CBE=005EA6=00548C=006536=00655B=0065F6=0095F4=007684=005BF9=006BD4}}
			
			\centering{}%
			\begin{tabular}{ccccc}
				\toprule 
				\multirow{2}{*}{Aspect ratio $a/b$} & \multicolumn{4}{c}{Torsion angle $\alpha$: relative error and convergence time}\tabularnewline
				\cmidrule{2-5} \cmidrule{3-5} \cmidrule{4-5} \cmidrule{5-5} 
				& PINNs strong form: Dis & DEM & PINNs strong form: Stress & DCEM\tabularnewline
				\midrule
				1.0 & 1.51\%, 25.9s & 0.73\%, 19.4s & 1.02\%, 12.4s & 0.65\%, 3.4s\tabularnewline
				1.2 & 1.32\%, 12.1s & 0.65\%, 9.1s & 0.97\%, 7.4s & 0.12\%, 9.2s\tabularnewline
				1.5 & 1.26\%, 8.1s & 0.74\%, 9.1s & 1.04\%, 18.4s & 0.64\%, 9.4s\tabularnewline
				2.0 & 1.01\%, 15.2s & 0.93\%, 9.5s & 0.95\%, 5.7s & 0.27\%, 4.7s\tabularnewline
				2.5 & 1.03\%, 15.1s & 1.07\%, 15.7s & 0.52\%, 8.7s & 0.13\%, 3.7s\tabularnewline
				3.0 & 0.95\%, 21.1s & 1.04\%, 17.4s & 0.67\%, 23.2s & 0.42\%, 6.7s\tabularnewline
				4.0 & 1.02\%, 25.6s & 1.12\%, 12.2s & 0.82\%, 21.4s & 0.13\%, 2.6s\tabularnewline
				5.0 & 1.14\%, 31.2s & 1.14\%, 20.7s & 0.89\%, 19.2s & 0.43\%, 8.2s\tabularnewline
				10.0 & 1.57\%, 16.1s & 1.04\%, 24.2s & 0.79\%, 21.4s & 0.27\%, 8.3s\tabularnewline
				Mean error & $1.21\%$ & $0.94\%$ & $0.85\%$ & $\boldsymbol{0.34\%}$\tabularnewline
				Mean time & $18.9s$ & $15.3s$ & $15.3s$ & $\boldsymbol{6.2s}$\tabularnewline
				\bottomrule
			\end{tabular}
		\end{table}
		\par
	\end{center}
	
	\Cref{fig:=0077E9=005F62=0067F1=005F62=006746=0081EA=007531=00626D=008F6Ctaumax=005BF9=006BD4} presents four methods for predicting the maximum shear stress  $\tau$. The effectiveness of these methods in predicting the maximum shear stress is demonstrated to be satisfactory. The PINNs strong form of displacement and the DEM fluctuate greatly, but the PINNs strong form of stress function and the DCEM fluctuate less because the pre-setting $\alpha$ affects the result of the stress solution, i.e., the initial prediction value will affect the accuracy and convergence speed.  Compared to PINNs and DEM based on displacement, \Cref{fig:Prantl_alpha_tau_inital}c  shows that PINNs and DCEM based on stress function more closely capture the essence of stress. As a result, methods based on stress function achieve higher accuracy than those based on displacement.
	
	\begin{figure}
		\begin{centering}
			\includegraphics[scale=2.0]{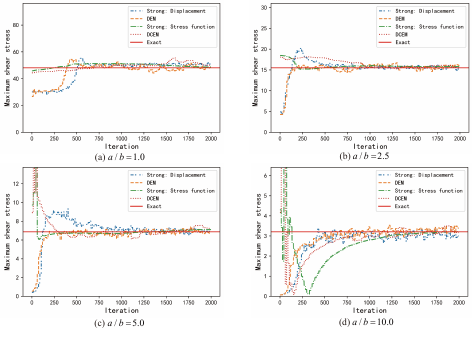}
			\par\end{centering}
		\caption{Comparison of the numerical and analytical solutions of the maximum shear stress $\tau$ of the rectangular cylinder under the strong form of PINNs displacement, DEM, the strong form of PINNs stress function and DCEM\label{fig:=0077E9=005F62=0067F1=005F62=006746=0081EA=007531=00626D=008F6Ctaumax=005BF9=006BD4}}
	\end{figure}

	DEM has been discussed in detail in \cite{loss_is_minimum_potential_energy}, so now we focus on DCEM only. \Cref{fig:=0077E9=005F62=0067F1=005F62=006746=0081EA=007531=00626D=008F6Ctau=004E91=0056FE}c,f show that the maximum absolute error is mainly concentrated on the boundary, because the maximum shear stress is concentrated on the boundary, so it cannot show that the learning of the boundary is worse than that of the interior simply. On the whole, the predictive shear stress effect of DCEM is close to the exact solution. The $\mathcal{L}_{2}^{rel}$ relative error norm and $\mathcal{H}_{1}^{rel}$ relative error seminorm in DCEM are calculated as follows 
	
	\begin{equation}
		\begin{split} \Vert e \Vert_{\mathcal{L}_{2}^{rel}} &= \frac{\Vert \phi^{pred} - \phi^{exact}\Vert_{\mathcal{L}_{2}}}{\Vert \phi^{exact}\Vert_{\mathcal{L}_{2}}} =  \frac{\int_{\Omega}(\phi^{pred} - \phi^{exact})^2 dxdy}{\int_{\Omega} (\phi^{exact})^{2} dxdy} \\
			\Vert e \Vert_{\mathcal{H}_{1}^{rel}} &= \frac{\Vert \phi^{pred} - \phi^{exact}\Vert_{\mathcal{H}_{1}}}{\Vert \phi^{exact}\Vert_{\mathcal{H}_{1}}} =  \frac{\int_{\Omega}( E^{pred} - E^{exact})^2 dxdy}{\int_{\Omega} (E^{exact})^{2} dxdy} \\
			E &= \int_{\Omega} (\frac{\partial \phi}{\partial x})^2 + (\frac{\partial \phi}{\partial y})^2 dxdy.
		\end{split}
	\end{equation}
	
	Four different neural networks were constructed, each having a different number of hidden layers (HL): 1HL, 2HL, 3HL, and 4HL. Each hidden layer consists of 30 neurons. Additionally, the NNs were trained for various steps ranging from 100 to 1000. As shown in \Cref{fig:=L2_H1_norm_prantl}, all neural network architectures demonstrated promising results when compared with the exact solution. Moreover, the DCEM exhibited convergence concerning the training steps in both $\mathcal{L}_{2}^{rel}$ and $\mathcal{H}_{1}^{rel}$ norms. However, it should be noted that the DCEM did not converge with respect to the number of hidden layers, as having more hidden layers did not necessarily result in improved accuracy.
	
	\begin{figure}
		\begin{centering}
			\includegraphics[scale=1.2]{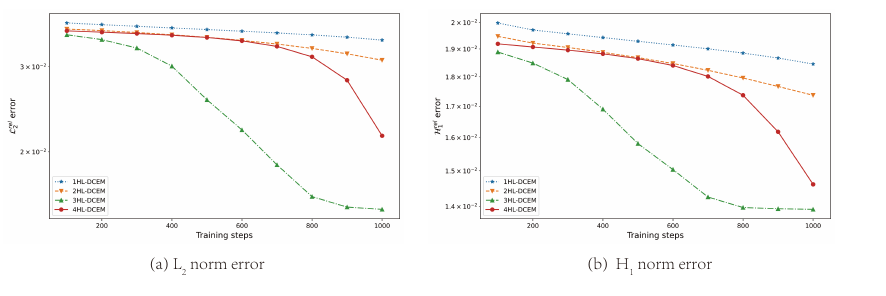}
			\par\end{centering}
		\caption{The error in terms of $\mathcal{L}_{2}^{rel}$ norm and $\mathcal{H}_{1}^{rel}$ seminorm of DCEM with respect to the training steps in Prandtl stress function.\label{fig:=L2_H1_norm_prantl}}
	\end{figure}	
	\begin{figure}
		\begin{centering}
			\includegraphics[scale=1.0]{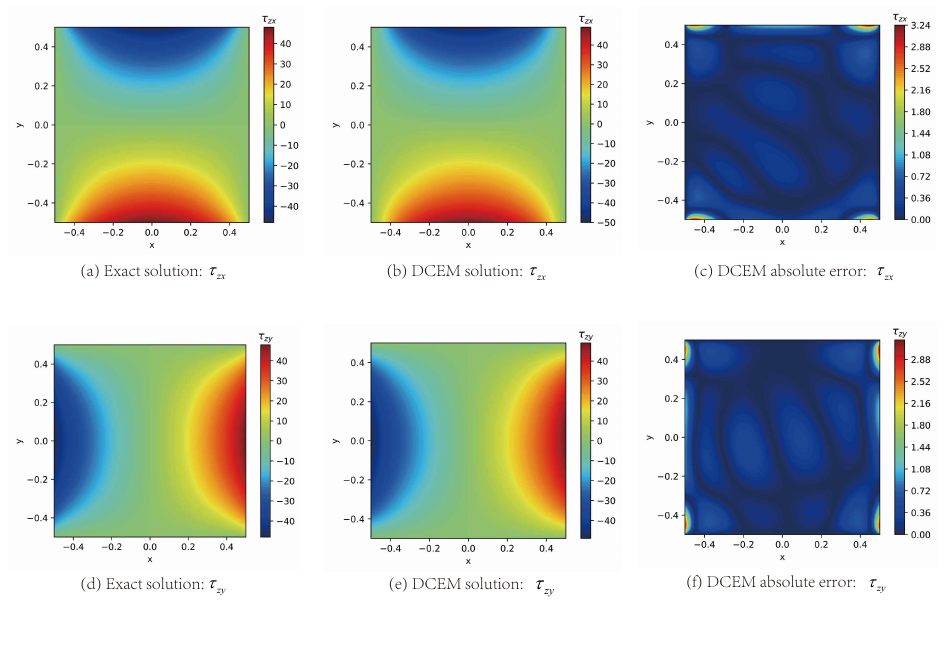}
			\par\end{centering}
		\centering{}\caption{Comparison of numerical and analytical solution of the shear stress in cylinder $\tau$ in DCEM \label{fig:=0077E9=005F62=0067F1=005F62=006746=0081EA=007531=00626D=008F6Ctau=004E91=0056FE}}
	\end{figure}
	
	Considering that the initial $\alpha$ of the DCEM  may affect the convergence rate and precision, the influence of different initial values of $\alpha$ on the convergence rate and precision is analyzed, as shown in \Cref{fig:Prantl_alpha_tau_inital}b. The convergence rate is slower with increasing $\alpha$. We explain it from the perspective of the strong form, this is because of the larger constant of the Poisson equation. The neural network is relatively more difficult to fit since the Poisson equation is more apparent in ups and downs.
	The initial value of $\alpha$  will not only affect the convergence speed but also significantly influence the accuracy of $\alpha$. The DCEM  needs to assume an admissible stress function since it is a full-force boundary condition problem, while the DEM does not need to assume an admissible displacement field. As a result, DEM is more convenient to deal with the full-force boundary condition problem, and this advantage can use the enormous approximate function space of the neural network to optimize and achieve remarkable results.

	We compare the DCEM-O (operator learning based on DCEM) with DCEM in this full-force boundary example. The input of the branch net is geometry information, $a$ and $b$. We deal with the $a$ and $b$ as the constant field function. The basis function is 
	$(x^{2}-\frac{a^{2}}{4})(y^{2}-\frac{b^{2}}{4})$,
	and the particular function is zero. The output of the DCEM-O can be written as:
	\begin{equation}
		\phi=(x^{2}-\frac{a^{2}}{4})(y^{2}-\frac{b^{2}}{4})NN_{B}(a,b;\boldsymbol{\theta})*NN_{T}(x,y;\boldsymbol{\theta}).
	\end{equation}
	The architecture of the trunk net consists of 6 hidden layers and 30 neurons in every hidden layer. The architecture of the branch net is 3 hidden layers and 30 neurons in every hidden layer. We use the 11 different $a/b$ ratios as the input of the branch net. The input of the trunk net is 100*100 equal spacing points. The output of DCEM-O is the Prandtl stress function $\phi$, and the output training dataset is from the analytical solution of the problem \cite{fung1953behavior} or the high fidelity numerical experiment (FEM based on the stress function element \cite{cen20118}). In this example, the data comes from the analytical solution of the rectangle, which can be written as:
	\begin{equation}
		\begin{split}\phi(x,y) & =G\alpha(\frac{b^{2}}{4}-y^{2})-\frac{8G\alpha b^{2}}{\pi^{3}}\sum_{i=0}^{\infty}\frac{(-1)^{i}}{(2i+1)^{3}}\frac{cosh(\lambda_{i}x)}{cosh(\lambda_{i}a/2)}cos(\lambda_{i}y)\\
			\lambda_{i} & =(2i+1)\frac{\pi}{b}
		\end{split}
	\end{equation}
	Note that the test dataset (we only test $a=b=1$) differs from the training set. \Cref{fig: the DCEM and DCEM-P in Prantl stress function} shows the comparison of DCEM with DCEM-O, and the absolute error of DCEM-O is lower than DCEM in the same iteration. The absolute error of DCEM is almost the same in the initial 250, 500, and 750 iterations, but DCEM-O converges to the exact solution dramatically in those iterations. \Cref{fig: the DCEM and DCEM-P in Prantl stress function}m,n shows that the DCEM-O converges faster than DCEM in terms of the $\mathcal{L}_{2}^{rel}$ and $\mathcal{H}_{1}^{rel}$.
	
	\begin{figure}
		
		\begin{centering}
			\includegraphics[scale=1.1]{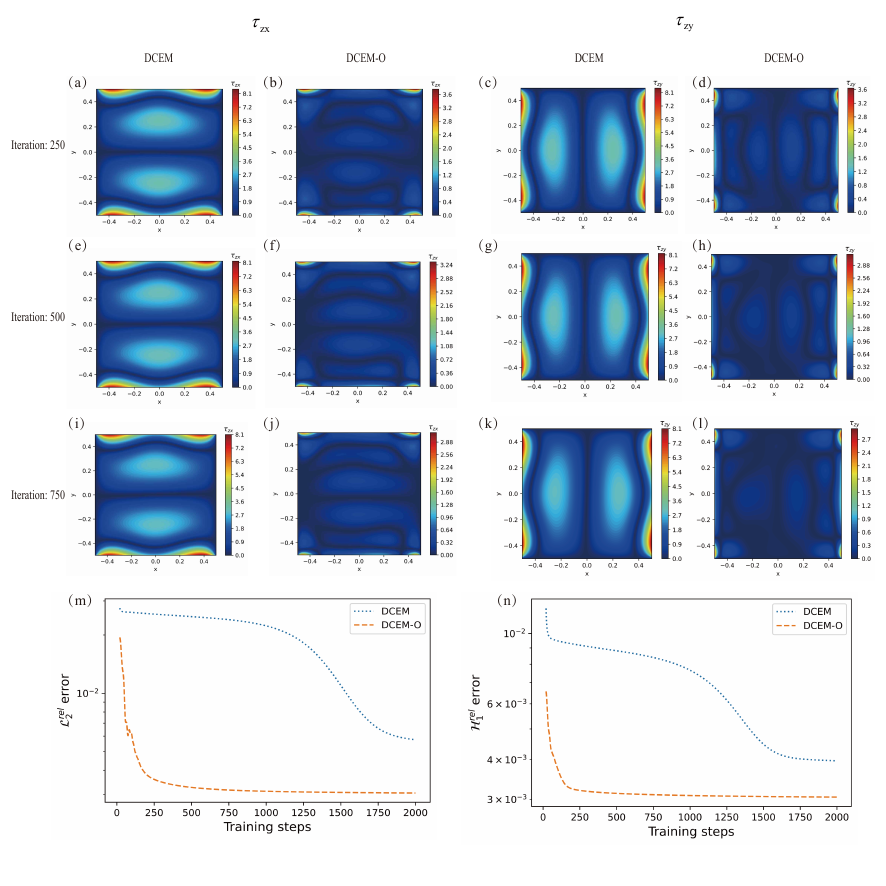}
			\par\end{centering}
		\caption{The absolute error of DCEM and DCEM-O in Prandtl stress function of cylinder: (a, b, c, d), (e, f, g, h), and (i, j, k, l) the absolution error in the number of iterations 250, 500, and 750 respectively; (a, e, i, c, g, k) the result of DCEM ; (b, f, j, d, h, l) the result of DCEM-O. (a, b, e, f, i, j) the result of $\tau_{zx}$; (c, d, g, h, k, l) the result of $\tau_{zy}$. (m) the evolution of  $\mathcal{L}_{2}^{rel}$  by DCEM and DCEM-O; (n) the evolution of  $\mathcal{H}_{1}^{rel}$ by DCEM and DCEM-O \label{fig: the DCEM and DCEM-P in Prantl stress function}}
		
	\end{figure}
	
	\section{Discussion}\label{sec:Discussion}

	\subsection{The comparison of DEM, PINNs strong form, and DCEM}
	PINNs strong form of stress function is suitable for dealing with full-force boundary conditions. However, it is not convenient to solve the problem with the displacement boundary because displacement boundary conditions are formulated in the form of boundary integral equations. On the other hand, DCEM based on the minimum complementary energy principle only needs to convert the displacement boundary conditions into the complementary potential for optimization when dealing with the displacement boundary conditions, which is difficult for the PINNs strong form of the stress function. From the construction of the admissible field, DEM is more suitable for dealing with the problems dominated by the force boundary conditions because it reduces the requirement of the construction of the admissible displacement field on the force boundary condition. In contrast, DCEM is suitable for dealing with displacement boundary conditions. 
	
	Therefore, from the perspective of admissible field construction, different problems are suitable for different deep energy methods (DEM or DCEM). One needs to understand the problem's nature so we can choose the suitable deep energy method based on potential or complementary energy.
	
	If we use the penalty method to satisfy the boundary condition, results show PINNs energy form depends more on the hyperparameters for the boundary condition than PINNs strong form empirically, because the energy methods would solve the wrong target if the admissible field is not satisfied. As a result, it is more important to construct the admissible field in DCEM and DEM than PINNs strong forms. 
	
	Mathematically, the choice of the form of PDEs can indeed affect the accuracy of the solution for a given physical problem. In the context of solid mechanics,  stress as an unknown and basic variable in the DCEM is generally considered to be more accurate than displacement as an unknown and basic variable in the DEM in terms of stress, and vice versa.

	\subsection{Applicability of Machine Learning in Solving PDEs}

	In many cases, linear problems can be efficiently solved using standard approaches such as the Finite Element Method (FEM). Moreover, for linear problems, reduced-order models can be employed to significantly reduce simulation time. On the other hand, when applying machine learning (ML) techniques to solve linear models, the originally well-posed linear problem, which involves finding a solution to a linear system of equations, is transformed into a nonlinear nonconvex optimization problem \cite{wang2022cenn}. This transformation introduces challenges as a unique solution cannot be guaranteed. Therefore, ML-based solutions for partial differential equations (PDEs) should primarily focus on nonlinear problems or optimization/inverse analysis problems. However, it is crucial to demonstrate that these methods can also effectively handle linear problems.
	
	\subsection{Energy Bounds in Variational Principles of Elasticity in terms of DEM and DCEM}
	In elasticity, the variational principles have upper and lower bounds on energy. The approximate displacement solution obtained using the principle of minimum potential energy is theoretically associated with lower strain energy compared to the exact displacement solution. Consequently, the structure exhibits slightly smaller displacements and strains, resulting in lower stresses derived from the constitutive model. On the other hand, the principle of complementary energy yields stresses that are theoretically associated with greater strain energy compared to the exact stress distribution, leading to a softer structural response and higher stress levels. However, it's worth noting that these upper and lower bounds do not directly apply to methods like DEM and DCEM, as shown in \Cref{fig: Different location comparison among FEM DEM and DCEM in non-uniform tension of central hole}. This is because these bounds are achieved in the context of optimization at the extremum, which is challenging to reach in the case of neural network-based methods due to the non-convex nature of the optimization landscape. Therefore, the energy bounds from the variational principles of elasticity may not be fully applied in DEM and DCEM.
	
	\subsection{Shear and volumetric locking problem in terms of DEM and DCEM}
	In the low-order elements of displacement finite elements, the deformation mode (interpolation function) cannot well represent volumetric strain (in incompressible problems) or shear strain (in structural elements such as beams, plates, shells, etc.). The stress element uses independent interpolation for stress, and the deformation mode can better reflect the above strain characteristics.
	
	However, in DEM, the displacement field is approximated by the neural networks, so it is extremely "high-order" to avoid the shear and locking problem. As a result, DCEM does not have the absolute advantage of handling volumetric and shear problems theoretically.  
	
	\section{Conclusion}\label{sec:Conclusion}
	We propose the deep complementary energy method (DCEM) based on the minimum complementary energy principle, a novel deep energy framework for solving elasticity problems in solid mechanics. Our comprehensive comparison between DEM, DCEM, and PINNs strong forms has demonstrated the superior accuracy and efficiency of DCEM, particularly in stress predictions.
	The construction of admissible function fields plays a crucial role in DEM and DCEM.
	From the perspective of constructing admissible function fields, DEM is better suited for problems dominated by force boundaries, whereas DCEM excels in handling problems dominated by displacement boundaries.
	We have extended the DCEM framework to DCEM-Plus (DCEM-P), introducing terms that naturally satisfy PDEs, and proposed the Deep Complementary Energy Operator Method (DCEM-O), leveraging operator learning in conjunction with physical equations.
	The result shows that DCEM-P and DCEM-O can further improve the accuracy and efficiency of DCEM, opening up exciting avenues for future research. 
	
	While DCEM has been applied successfully to the two most common stress functions, we recognize the need for testing on three-dimensional problems, involving Morera and Maxwell stress functions, which we intend to explore in our future work.
	 This work proposes an important supplementary energy form of the deep energy method. In linear elasticity, the relationship between stress and strain is linear, and the relationship between displacement and strain is also linear. Thus, there is no problem for computation based on both energy principles in linear elasticity, i.e. potential and complementary energy. However, there are lots of challenges in nonlinear problems.
	Complementary energy exists in solid mechanics and other physical fields, such as Helmholtz free energy, enthalpy, Gibbs free energy, and internal energy in thermodynamics. The relationship between energy conforms to the Legendre transformation in mathematics. Therefore, DCEM can also apply to the energy principles of Legendre transformation in other physics.
	Although the computational efficiency is different according to the different descriptions of PDEs, such as strong form, weak form, and energy form, the different descriptions are the same mathematically because the different descriptions solve the same PDEs problem.  As a result, it is interesting to research the weak form of DCEM  based on Variational PINNs (V-PINNs \cite{kharazmi2019variational}) and even the weak form of DCEM with subdomain (hp-VPINNs \cite{hp-VPINN}).
	  Standard approaches like FEM can easily and efficiently solve linear problems, and ROD (Reduced Order Models) can be used to drastically reduce the simulation time of such problems. However, deep learning-based solutions for linear models convert the well-posed linear problem into a nonlinear non-convex optimization problem, which cannot guarantee a unique solution. Thus, methods based on deep learning for solving PDEs should focus on nonlinear forward problems and inverse analysis problems. Additionally, the combination of force and displacement loads on the same boundary deserves further investigation by DCEM.
	
	The work in this paper reflects the broad prospect of combining deep learning with computational mechanics. It is not easy for a purely physics-based energy method to surpass traditional finite elements, and a purely data-driven model will bring about the problem of excessive data requirement. With the in-depth research on neural networks and the advancement of computing power, combined with more and more calculation data results, there are strong reasons to believe that operator learning based on the energy method can bring a balance between data and physical equations.  If more high-precision calculation results are stored in the future, the operator learning based on the energy method has the potential to exceed the traditional algorithms, which will give computational mechanics new and broad research prospects.
	
    Finally, let's remember what Frank Wilczek (2004 Nobel Prize in Physics) said: "Complementarity, in its most basic form, is the concept that one single thing, when considered from different perspectives, can seem to have very different or even contradictory properties. When the description of a system using one kind of model gets too complicated to work with, we sometimes can find a complementary model, based on different concepts, to answer important questions. For that reason, a complete understanding of the fundamental laws, if we ever achieve it, would be neither “the Theory of Everything" nor "the End of Science." We would still need complementary descriptions of reality. There would still be plenty of great questions left unanswered, and plenty of great scientific work left to do."
    This is the philosophy of our DCEM. 

	\bmsection*{Author contributions}
	
	\textbf{Yizheng Wang}: Conceptualisation, Methodology, Coding, Formal analysis, Writing-Original draft. \textbf{Jia Sun}: Advising, Scientific Discussions, Writing-Reviewing and Editing.  \textbf{Timon Rabczuk}: Discussions, Writing-Reviewing and Editing. \textbf{Yinghua Liu}: Conceptualisation, Discussions, Writing-Reviewing and Editing, Supervision.
	
	\bmsection*{Acknowledgement}
	The study was supported by the Key Project of the National Natural Science Foundation of China (12332005). The authors would like to thank Jingyun Bai, Zaiyuan Lu, Yuqing Du, Hugo Santos, and Lu Lu for their helpful discussions.

	\bmsection*{Financial disclosure}
	
	None reported.
	
	\bmsection*{Conflict of interest}
	
	The authors declare no potential conflict of interests.
	\bmsection*{Data availability statement}
	The code of this work is available at \url{https://github.com/yizheng-wang/Research-on-Solving-Partial-Differential-Equations-of-Solid-Mechanics-Based-on-PINN}.

	\bibliography{bibtex_DCEM}
	
	\appendix
\bmsection{The construction of the basis function} \label{sec:Appendix-C.basis function}

In this section, we discuss the construction of the basis function. The distance function is a special form of the basis function, as explained in \Cref{sec: DCEM_method}. If the deep energy method is based on the principle of the minimum potential energy, the basis function is zero only if the position is on the Dirichlet boundary condition. Thus, we can construct different basis functions, as shown in \Cref{fig: distance function}. The idea of the contour line is adopted in the construction of the basis function.   

\begin{figure}
	\begin{centering}
		\includegraphics[scale=1.1]{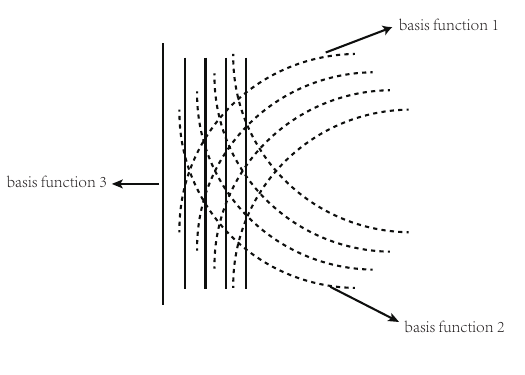}
		\par\end{centering}
	\caption{The schematic of the different basis functions. \label{fig: distance function}}
\end{figure}

For the sake of simplicity, we use three different basis functions $NN_{b}^{1}$, $NN_{b}^{2}$, and $NN_{b}^{3}$ to illustrate the idea.
We assume that  the direction of the derivative of the $NN_{b}^{1}$, $NN_{b}^{2}$, and $NN_{b}^{3}$ are zero, i.e. 

\begin{equation}
	\begin{aligned}\frac{\partial NN_{b}^{(1)}}{\partial \boldsymbol{l}_{1}} & =0\\
		\frac{\partial NN_{b}^{(2)}}{\partial \boldsymbol{l}_{2}} & =0\\
		\frac{\partial NN_{b}^{(3)}}{\partial \boldsymbol{l}_{3}} & =0,
	\end{aligned}
	\label{eq:=0065B9=005411=005BFC=0065700}
\end{equation}
where $\boldsymbol{l}_{1}$, $\boldsymbol{l}_{2}$, and $\boldsymbol{l}_{3}$ are the corresponding direction of the derivative, i.e. tangent to contours. It is hard for the general net to learn because the $NN_{b}=0$ and $\partial NN_{b}/ \boldsymbol{l}=0$ when the direction is parallel to the contours near the boundary. The direction of the derivative can be written as follows:
\begin{equation}
	\begin{aligned}\frac{\partial\phi}{\partial \boldsymbol{l}_{1}} & =\frac{\partial NN_{p}}{\partial \boldsymbol{l}_{1}}+\sum_{i=1}^{3}\frac{\partial NN_{g}^{(i)}}{\partial \boldsymbol{l}_{1}}*NN_{b}^{(i)}+\sum_{i=1}^{3}NN_{g}^{(i)}*\frac{\partial NN_{b}^{(i)}}{\partial \boldsymbol{l}_{1}}\\
		\frac{\partial\phi}{\partial \boldsymbol{l}_{2}} & =\frac{\partial NN_{p}}{\partial \boldsymbol{l}_{2}}+\sum_{i=1}^{3}\frac{\partial NN_{g}^{(i)}}{\partial \boldsymbol{l}_{2}}*NN_{b}^{(i)}+\sum_{i=1}^{3}NN_{g}^{(i)}*\frac{\partial NN_{b}^{(i)}}{\partial \boldsymbol{l}_{2}}\\
		\frac{\partial\phi}{\partial \boldsymbol{l}_{3}} & =\frac{\partial NN_{p}}{\partial \boldsymbol{l}_{3}}+\sum_{i=1}^{3}\frac{\partial NN_{g}^{(i)}}{\partial \boldsymbol{l}_{3}}*NN_{b}^{(i)}+\sum_{i=1}^{3}NN_{g}^{(i)}*\frac{\partial NN_{b}^{(i)}}{\partial \boldsymbol{l}_{3}}.
	\end{aligned}
	\label{eq:the dirivative of the basis function}
\end{equation}
Because $\partial NN_{b}^{1}/ \boldsymbol{l}_{1}=0$,  $\partial NN_{b}^{2}/ \boldsymbol{l}_{2}=0$, and  $\partial NN_{b}^{3}/ \boldsymbol{l}_{3}=0$, 
\Cref{eq:the dirivative of the basis function} can be further written as
\begin{equation}
	\begin{aligned}\frac{\partial\phi}{\partial \boldsymbol{l}_{1}} & =\frac{\partial NN_{p}}{\partial \boldsymbol{l}_{1}}+NN_{g}^{(2)}*\frac{\partial NN_{b}^{(2)}}{\partial \boldsymbol{l}_{1}}+NN_{g}^{(3)}*\frac{\partial NN_{b}^{(3)}}{\partial \boldsymbol{l}_{1}}\\
		\frac{\partial\phi}{\partial \boldsymbol{l}_{2}} & =\frac{\partial NN_{p}}{\partial \boldsymbol{l}_{2}}+NN_{g}^{(1)}*\frac{\partial NN_{b}^{(1)}}{\partial \boldsymbol{l}_{2}}+NN_{g}^{(3)}*\frac{\partial NN_{b}^{(3)}}{\partial \boldsymbol{l}_{2}}\\
		\frac{\partial\phi}{\partial \boldsymbol{l}_{3}} & =\frac{\partial NN_{p}}{\partial \boldsymbol{l}_{3}}+NN_{g}^{(1)}*\frac{\partial NN_{b}^{(1)}}{\partial \boldsymbol{l}_{3}}+NN_{g}^{(2)}*\frac{\partial NN_{b}^{(2)}}{\partial \boldsymbol{l}_{3}}
	\end{aligned}
	\label{eq:=005316=007B80=00540E=007684=0053EF=0080FD=005E94=00529B=0051FD=006570=007684=0065B9=005411=005BFC=006570}
\end{equation}

In the energy principle training, the basis function network and the particular net are frozen and not trained, and only the general network can be trained. The \Cref{eq:=005316=007B80=00540E=007684=0053EF=0080FD=005E94=00529B=0051FD=006570=007684=0065B9=005411=005BFC=006570} shows that although one of the basis function networks makes the corresponding general network impossible training, the other two generalized networks are still trainable. As a result, constructing different basis functions is beneficial to training general networks based on the energy principle and improving the accuracy of DEM near the boundary.

\bmsection{The basis function in Airy stress function } \label{sec:Appendix-C.the basis function in Airy stress function}

If we use the idea of the distance function (basis function), we show
that the basis function must satisfy $\{\boldsymbol{x}_{i}\in\Gamma^{t},\phi|_{\boldsymbol{x}_{i}}=0\}_{i=1}^{n};\{\boldsymbol{x}_{i}\in\Gamma^{t},\partial\phi/\partial\boldsymbol{n}|_{\boldsymbol{x}_{i}}=0\}_{i=1}^{n}$.
The construction of the admissible stress function can be written
as:
\[
\phi=\phi_{p}+\phi_{g}*\phi_{b},
\]
where the particular solution $\phi_{p}$ satisfy the natural boundary
condition. We analyze 
\[
\frac{\partial\phi}{\partial\boldsymbol{n}}=\frac{\partial\phi_{p}}{\partial\boldsymbol{n}}+\frac{\partial\phi_{g}}{\partial\boldsymbol{n}}*\phi_{b}+\phi_{g}*\frac{\partial\phi_{b}}{\partial\boldsymbol{n}}.
\]

If we set $\{\boldsymbol{x}_{i}\in\Gamma^{t},\phi_{b}|_{\boldsymbol{x}_{i}}=0\}_{i=1}^{n};\{\boldsymbol{x}_{i}\in\Gamma^{t},\partial\phi_{b}/\partial\boldsymbol{n}|_{\boldsymbol{x}_{i}}=0\}_{i=1}^{n}$,
the stress function can satisfy the admissible stress function.

	\bmsection{The application of the property of the Airy stress function }\label{sec:Appendix-B.-proof of the particular solution network}
	
	The proof is based on free body force. 
	
	In this appendix, we show 
	\begin{equation}
		\begin{aligned}n_{x}\frac{\partial^{2}\phi}{\partial y^{2}}-n_{y}\frac{\partial^{2}\phi}{\partial x\partial y} & =\bar{t}_{x}\\
			-n_{x}\frac{\partial^{2}\phi}{\partial x\partial y}+n_{y}\frac{\partial^{2}\phi}{\partial^{2}x} & =\bar{t}_{y}
		\end{aligned}
		\label{eq:the external condition}
	\end{equation}
	is equal to 
	\begin{equation}
		\begin{aligned}\phi & =M\\
			\frac{\partial\phi}{\partial\boldsymbol{n}} & =-R_{s}=-s_{x}R_{x}-s_{y}R_{y},
		\end{aligned}
		\label{eq:Airy property}
	\end{equation}
	where $M$ and $R_{s}$ are explained in \Cref{fig:The-value-of stress function}a.
	We consider the moment $M$ of an external force about a point $B$
	from the initial point $A$
	\begin{equation}
		M|_{B}=\int_{A}^{B}[\bar{t}_{x}(y_{B}-y)+\bar{t}_{y}(x-x_{B})]ds.\label{eq:momont of point A in Airy stress function}
	\end{equation}
	
	We substitute \Cref{eq:the external condition} into \Cref{eq:momont of point A in Airy stress function},
	and we can obtain
	\[
	\begin{aligned}M|_{B} & =\int_{A}^{B}[(n_{x}\frac{\partial^{2}\phi}{\partial y^{2}}-n_{y}\frac{\partial^{2}\phi}{\partial x\partial y})(y_{B}-y)+(-n_{x}\frac{\partial^{2}\phi}{\partial x\partial y}+n_{y}\frac{\partial^{2}\phi}{\partial^{2}x})(x-x_{B})]ds\\
		& =\int_{A}^{B}[(\frac{dy}{ds}\frac{\partial^{2}\phi}{\partial y^{2}}+\frac{dx}{ds}\frac{\partial^{2}\phi}{\partial x\partial y})(y_{B}-y)+(-\frac{dy}{ds}\frac{\partial^{2}\phi}{\partial x\partial y}-\frac{dx}{ds}\frac{\partial^{2}\phi}{\partial^{2}x})(x-x_{B})]ds\\
		& =\int_{A}^{B}[\frac{d(\frac{\partial\phi}{\partial y})}{ds}(y_{B}-y)+\frac{d(\frac{\partial\phi}{\partial x})}{ds}(x_{B}-x)]ds\\
		& =\int_{A}^{B}d(\frac{\partial\phi}{\partial y})(y_{B}-y)+d(\frac{\partial\phi}{\partial x})(x_{B}-x)\\
		& =[\frac{\partial\phi}{\partial y}(y_{B}-y)+\frac{\partial\phi}{\partial x}(x_{B}-x)]|_{A}^{B}-\int_{A}^{B}[\frac{\partial\phi}{\partial y}(-\frac{dy}{ds})-\frac{\partial\phi}{\partial x}(\frac{dx}{ds})]ds\\
		& =-[\frac{\partial\phi}{\partial y}|_{A}(y_{B}-y_{A})+\frac{\partial\phi}{\partial x}|_{A}(x_{B}-x_{A})]+\int_{A}^{B}d\phi,
	\end{aligned}
	\]
	where $n_{x}$ and $n_{y}$ are the normal direction of boundary,
	and $s_{x}$ and $s_{y}$ are the tangent direction of the boundary,
	
	\[
	\begin{aligned}n_{x} & =\frac{dy}{ds}=s_{y}\\
		n_{y} & =-\frac{dx}{ds}=-s_{x}.
	\end{aligned}
	\]
	
	We set 
	\[
	\phi|_{A}=\frac{\partial\phi}{\partial x}|_{A}=\frac{\partial\phi}{\partial y}|_{A}=0,
	\]
	because point $A$ is the initial point.
	
	Thus 
	\[
	M=\phi|_{B}.
	\]
	
	Now we show the second part of the \Cref{eq:Airy property}, and we also
	substitute \Cref{eq:the external condition} into the second part of
	\Cref{eq:Airy property}, and we can obtain
	\begin{align*}
		-R_{s}|_{B} & =-s_{x}R_{x}-s_{y}R_{y}\\
		& =-s_{x}\int_{A}^{B}\bar{t}_{x}ds-s_{y}\int_{A}^{B}\bar{t}_{y}ds\\
		& =-\frac{dx}{ds}\int_{A}^{B}(\frac{dy}{ds}\frac{\partial^{2}\phi}{\partial y^{2}}+\frac{dx}{ds}\frac{\partial^{2}\phi}{\partial x\partial y})ds-\frac{dy}{ds}\int_{A}^{B}(-\frac{dy}{ds}\frac{\partial^{2}\phi}{\partial x\partial y}-\frac{dx}{ds}\frac{\partial^{2}\phi}{\partial^{2}x})ds\\
		& =-\frac{dx}{ds}\int_{A}^{B}(\frac{d(\frac{\partial\phi}{\partial y})}{ds})ds+\frac{dy}{ds}\int_{A}^{B}(\frac{d(\frac{\partial\phi}{\partial x})}{ds})ds\\
		& =-\frac{dx}{ds}(\frac{\partial\phi}{\partial y}|_{A}^{B})+\frac{dy}{ds}(\frac{\partial\phi}{\partial x}|_{A}^{B})\\
		& =n_{y}(\frac{\partial\phi}{\partial y}|_{B})+n_{x}(\frac{\partial\phi}{\partial x}|_{B}).
	\end{align*}
	As a result, 
	\[
	\frac{\partial\phi}{\partial\boldsymbol{n}}=-R_{s}.
	\]
	
	The proof from \Cref{eq:Airy property} to \Cref{eq:the external condition}
	is the same, so we will not explain the details here.

\end{document}